\documentclass[letterpaper]{article} 
\usepackage{aaai25} 
\usepackage{times}  
\usepackage{helvet}  
\usepackage{courier}  
\usepackage[hyphens]{url}  
\usepackage{graphicx} 
\usepackage{natbib}  
\usepackage{caption} 
\usepackage{algorithm}
\usepackage{algorithmic}

\usepackage{arydshln}

\usepackage{newfloat}
\usepackage{listings}
\DeclareCaptionStyle{ruled}{labelfont=normalfont,labelsep=colon,strut=off} 
\floatstyle{ruled}
\newfloat{listing}{tb}{lst}{}
\floatname{listing}{Listing}
\title{Realistic Evaluation of Test-Time Adaptation: \\ Unsupervised Hyperparameter Selection}

\title{Realistic Evaluation of Test-Time Adaptation Algorithms: \\ Unsupervised Hyperparameter Selection}
\author {
    S. Cygert\textsuperscript{\rm 1,2},
    D. Sójka\textsuperscript{\rm 1,3},
    T. Trzciński\textsuperscript{\rm 1,4,6},
    B. Twardowski\textsuperscript{\rm 1,5}
}
\affiliations {
    \textsuperscript{\rm 1}IDEAS NCBR, 
    \textsuperscript{\rm 2}Gdańsk University of Technology\\
    \textsuperscript{\rm 3}Poznań University of Technology,
    \textsuperscript{\rm 4}Warsaw University of Technology\\
    \textsuperscript{\rm 5}Tooploox,
     \textsuperscript{\rm 6}CVC Barcelona\\
    sebastian.cygert@ideas-ncbr.pl
}

\usepackage{bibentry}

\usepackage{multirow}
\usepackage{xcolor}
\usepackage{subcaption}
\usepackage{amsmath}
\usepackage{booktabs}

\definecolor{dontkillmyeyesgreen}{rgb}{0.09, 0.45, 0.27}
\definecolor{debianred}{rgb}{0.84, 0.04, 0.33}

\begin{document}

\maketitle

\begin{abstract}
Test-Time Adaptation (TTA) has recently emerged as a promising strategy for tackling the problem of machine learning model robustness under distribution shifts by adapting the model during inference without access to any labels.
Because of task difficulty,  hyperparameters strongly influence the effectiveness of adaptation. However, the literature has provided little exploration into optimal hyperparameter selection.
In this work, we tackle this problem by evaluating existing TTA methods using surrogate-based hp-selection strategies (which do not assume access to the test labels) to obtain a more realistic evaluation of their performance. 
We show that some of the recent state-of-the-art methods exhibit inferior performance compared to the previous algorithms when using our more realistic evaluation setup. Further, we show that forgetting is still a problem in TTA as the only method that is robust to hp-selection resets the model to the initial state at every step.
We analyze different types of unsupervised selection strategies, and while they work reasonably well in most scenarios, the only strategies that work consistently well use some kind of supervision (either by a limited number of annotated test samples or by using pretraining data). 
Our findings underscore the need for further research
with more rigorous benchmarking by explicitly stating model selection strategies, to facilitate which we open-source our code\footnote{The code will be released upon acceptance.}.
\end{abstract}

%

\section{Introduction}

\label{sec:intro}

Machine learning models typically assume that training and testing data originate from similar distribution. 
However, in real-world applications, one observes a distribution shift between
the training (source) and testing (target) data domains,
which, in turn, hurts the performance of the model during inference~\cite{texturebias,hendrycksD19,wilds}. 
Unsupervised domain adaptation methods explore adapting a model trained on the source domain to the target one, having access only to unlabelled target data~\cite{pmlr-v28-gong13}. Test-time adaptation (TTA) extends this scenario to the \textit{online} setting, where the model needs to adapt on the fly. 

Recently, various approaches for test-time adaptation were developed. 
An example strategy involves entropy minimization on test data using various pseudo-labeling approaches~\cite{TENT,eata,SAR}. 
Yet, without any supervision at test time, errors in pseudo-labels are unavoidable, which can result in error accumulation~\cite{erroraccumulation} and pose a great challenge for TTA methods. 
As a result, it was shown
that existing TTA methods are very vulnerable to the choice of hyperparameters (such as learning rate or momentum of the optimizer)~\cite{boudiafMAB22_parameterfree,zhao_2023_pitfallsTTA}. 

Apart from the inherent difficulty of TTA one subtle but important challenge arises: how to select the hyperparameters for a given method and testing scenario, in a practical way, that is without access to test labels? This question has received little attention in the literature, with most of the methods, not specifying model selection strategy ~\cite{TENT, adacontrast, COTTA,  eata,note,SAR,rmt,sojka,rotta,dart,aaaiSu0J24}, and only listing the final hyperparameters.
To provide a fair comparison, recent benchmark~\cite{zhao_2023_pitfallsTTA} compared accuracies of various TTA methods using \textsc{oracle} selection strategy, that is assuming access to the test data labels. 
However, this can only give an overoptimistic measure of their potential performance, as in practice, we won't have access to test labels to perform optimal model selection. Also, this can give an advantage to methods with more hyperparameters.

To provide a realistic evaluation of TTA methods, we propose and evaluate different surrogate measures for model selection in TTA that do not use test labels. 
While the problem of unsupervised model selection is also known from a domain adaptation perspective ~\cite{benchmarkingUDA,reproducible}, in this work, we argue that TTA presents a unique challenge due to its dynamic nature, e.g., by showing that simply increasing the length of the adaptation sequence may result in degenerated solutions.
Some of our findings include:
\begin{itemize}
    \item The ranking of TTA methods varies across different selection strategies, i.e., some methods that perform well using the ORACLE selection strategy underperform when more realistic (unsupervised) hyperparameter selection. 
    \item We show that forgetting is an important problem in TTA as the only method that is robust to hp-selection resets the model to the initial state at every step (which comes at a high computational cost).
    \item None of the existing model selection strategies is superior to others and able to match \textsc{oracle's} strategy consistently. 
    \item When any kind of supervision is prohibited, we show that existing TTA methods can be grouped by their optimization objective and accompanied with an unsupervised hp-selection that works close to optimal. However, those findings do not hold after we increase the adaptation difficulty and can result in degenerated solutions.
    \item Across all evaluated metrics, using some source of supervision (using a very limited number of annotated samples or source data) is the only strategy that works universally well across all evaluated scenarios.
    \item We show that the main problem for unsupervised hyperparameter selection strategies is a small margin between viable and suboptimal solutions.

\end{itemize}

In conclusion, the hyperparameter tuning approach employed in the existing evaluation protocols is impractical and fails to adequately evaluate the capabilities of each TTA method. 
Our work emphasizes the importance of conducting thorough assessments using clearly defined model selection strategies, for which we create a testbed.

\section{Related Work}

\noindent\textbf{Model selection in TTA}
Model selection is a known problem in literature~\cite{raschkaModelSelection}. The situation complicates for TTA where we cannot assume access to the validation set.
This raises concerns for TTA, as existing methods are vulnerable to changes in hyperparameters~\cite{,boudiafMAB22_parameterfree,unravel}.
Recently,~\citet{zhao_2023_pitfallsTTA} evaluated TTA methods assuming \textsc{oracle} selection strategy, that is, by accessing test labels. However, this may not give a fair comparison between different methods as it may favor methods with
more hyperparameters.
To the best of our knowledge, only two papers have used model selection strategies that do not access access to target labels. \citet{rusakSPEGBBB22}, have used cross-dataset validation to select hyperparameters. In particular, they used the optimal parameters selected from the ImageNet-C dataset for testing on different benchmarks. 
Analogous strategy for parameter selection was used in~\citet{boudiafMAB22_parameterfree}. 
The authors found that existing TTA methods lack hyperparameter transferability, and they proposed a parameter-free method (\textsc{lame}). 

\noindent\textbf{Unsupervised model selection} For domain adaptation, 
~\citet{gulrajaniL21LostDomain} stressed the importance of providing model selection strategies. They tested different strategies for unsupervised model selection: i) source dataset accuracy ii) cross-dataset accuracy and iii) limited access to the target labels and have shown that under such fair comparison, no domain adaptation method works better than standard empirical risk minimization training. 
Recently, unsupervised model selection strategies were evaluated for domain~\cite{benchmarkingUDA} and partial domain adaptation~\cite{reproducible} and in time-series anomaly detection~\cite{goswamiCCMK23}.
We are inspired by those studies and examine various strategies for TTA. 
\section{Test-time adaptation preliminaries}

Given the source model $f_{\theta_S}$, trained on the labeled data from the source domain $\mathcal{D}_S=\{\mathcal{X}_S, \mathcal{Y}_S\}$, the objective is to adapt it to the test data $\mathcal{D}_T=\{\mathcal{X}_T, \mathcal{Y}_T\}$.
In the source data $\mathcal{D}_S$, sample-label pairs $(x_i\in\mathcal{X}_S, y_i\in\mathcal{Y}_S)$ are distributed according to the probability distribution $\mathcal{P}_S(\boldsymbol{x},y)$. The test samples $x_i\in\mathcal{X}_T$ are modeled by a different distribution $\mathcal{P}_T(\boldsymbol{x}, y) \neq \mathcal{P}_S(\boldsymbol{x},y)$. 
At each testing time step $t$, the TTA process aims to adapt the model $f_{\theta_S}$ to better align with the distribution $\mathcal{P}_T$. 
Importantly, unlabeled data samples $\boldsymbol{x}^t \sim \mathcal{P}_T$ are the only information available regarding the unseen distribution $\mathcal{P}_T$.

Hyperparameter selection is a crucial part of all machine learning pipelines. 
In a standard supervised learning setting, a validation set can be used to estimate the model’s accuracy. However, in TTA such an approach is not possible as there is no access to labeled target samples. Below, we discuss several approaches for model selection in TTA, which we use in this work.

\subsection{Improved hyperparameter selection in TTA}
There is a lack of rigorous and realistic hyperparameter selection in TTA, as it typically assumes the availability of test labels. This assumption, we argue, may give an overoptimistic measure of the TTA method's performance and does not compare them fairly. Therefore, we design a protocol of more realistic hyperparameter selection, and to achieve that, we examine the feasibility of the following selection strategies, which do not assume access to test labels. 

\noindent\textbf{Source accuracy} \textsc{(s-acc)} The most simple strategy uses a validation set from the source domain to estimate the model's performance in the target domain, e.g., in domain adaptation~\cite{Ganin15}. 
Interestingly, \citet{TaoriDSCRS20} and~\citet{accontheline} show that there is a linear correlation between accuracy on i.i.d. and o.o.d. data which may favor such a strategy. However, there are also some contradictory results that show that this may not be always true~\cite{YouWLJ19,TeneyLOA23}. Hence, it is important to validate this strategy in TTA settings. Finally, it is important to note that some TTA methods assume there is no access to the source data, e.g., because of privacy concerns~\cite{sfda}. In such applications, this strategy may not be valid.

\noindent\textbf{Cross-validation accuracy} \textsc{(c-acc)} Another viable approach is to select hyperparameters on another dataset. This is a potentially valid strategy since plenty of benchmarks exist for TTA that could be utilized for this purpose. Here, we follow~\citet{rusakSPEGBBB22} and perform hyperparameter selection on ImageNet-C dataset, which they show to work well. When the target dataset of the TTA procedure is ImageNet-C we use CIFAR-100-C for fair comparison. 

\noindent\textbf{Entropy} \textsc{(ent)} Minimizing the Shannon entropy $H(\hat{y})=-\sum_{c}p(\hat{y}_c)\log p(\hat{y}_c)$, where $\hat{y}_c$ is a probability prediction of class $c$, of predictions for target samples $\hat{y}=f_\theta(\boldsymbol{x}^t)$ was introduced for test-time adaptation in~\citet{TENT} because i) it is related to error, as more confident predictions are all-in-all more correct and ii) entropy is related to shifts due to corruption, as more corruption results in more entropy. This strategy was adopted by several works showing its efficiency ~\cite{TENT,eata,SAR}. However, as shown by~\citet{boudiafMAB22_parameterfree} in some scenarios the entropy does not correlate with the accuracy. 
One possible failure mode is a model that is incorrectly confident. For example, the model might incorrectly classify all samples in the dataset as belonging to the same class, which results in low entropy but also small model accuracy.

\noindent\textbf{Model consistency} \textsc{(con)} Consistency regularization is an important component of many self-supervised learning algorithms~\cite{fixmatch}.
It enforces the model $f_\theta$ to have similar predictions when fed perturbed versions $\boldsymbol{\Tilde{x}}_i$ of the same image $\boldsymbol{x}_i$ ($f_\theta(\boldsymbol{x}_i)\simeq f_\theta(\boldsymbol{\Tilde{x}}_i)$). It was also used to drive training of the model during the TTA phase~\cite{Nguyen_2023_CVPR}. As the choice of perturbations, we use the augmentation pipeline introduced in CoTTA~\cite{COTTA}, as it is quite commonly used in TTA and was shown to work well, which include, e.g., color jittering, Gaussian noise, and blur. Further information can be found in the appendix. 

\noindent\textbf{Soft neighborhood density} \textsc{(snd)} SND metric was proposed for unsupervised model selection in domain adaptation~\cite{saitoKTSDS21}. It is based on the intuition that a well-trained model embeds target samples of the same class nearby, forming dense neighborhoods in feature space. A high \textsc{snd} score means that each feature is close to many other features with similar pseudo-labels, which can indicate good clustering. A failure mode for this metric is predicting all features into a single cluster.

\noindent\textbf{Using test labels} \textsc{(oracle}) We also incorporate \textsc{oracle} hyperparameter selection strategy to measure upper-bound for evaluated methods. Additionally, we use one more additional selection strategy, \textsc{100-rnd}, that assumes limited access to the test data: following ~\citet{reproducible}, we use labels from 100 randomly selected images from test data.

\noindent\textbf{Median} (\textsc{med}) Finally, we show results when selecting a median model. It shows whether a given selection strategy works better than chosing an average model and the sensitivity of TTA methods to hyperparameters, i.e., a highly robust model will have a median close to the \textsc{oracle} results.

All methods are compared against the \textsc{source} model, which assumes no adaptation at all.

\section{Experimental Setup}
\label{sec:exp}

\begin{table*}[t!]
    \centering
    \resizebox{0.8\textwidth}{!}{
    \begin{tabular}{@{\hskip2pt}l@{\hskip2pt}l@{\hskip2pt}}
        TTA Methods  & TENT,
        SAR, LAME, 
        EATA, RMT, AdaContrast, MEMO\\
        \midrule
        Model Selection 
        & \textsc{s-acc}, \textsc{ent}, 
        \textsc{c-acc}, 
        \textsc{con}, \textsc{snd}
        \textsc{100-rnd}, \textsc{oracle},  \\
        \midrule
        Datasets & CIFAR100-C, ImageNet-C, DomainNet-126, ImageNet-R, ImageNetV2 \\
        \midrule
        Tuned parameters & learning rate, momentum and method-specific parameters 
        \\
        \midrule
        Experimental protocol & 5 datasets, varying levels of temporal correlation and adaptation sequence length\\
    \end{tabular}
    } 
    \caption{Summary of experimental setup considered in this work.}
    \label{table:our_summary_exp}
\end{table*}
\raggedbottom

\subsection{Datasets}
For experiments, we utilize two widely used datasets for TTA evaluation with artificial image corruptions: \mbox{CIFAR100-C} and \mbox{ImageNet-C}~\cite{corrupted_datasets}. Moreover, we use two datasets consisting of images without corruptions, but portraying objects in different domains: \mbox{DomainNet-126}~\cite{domainnet126} and \mbox{ImageNet-R}~\cite{imagenetr}. Finally, we incorporate an ImageNetV2~\cite{RechtRSS19} dataset, which allows us to measure model adaptation performance under real-world distribution shift.
More details in the appendix.

Additionally, we test the influence of temporally correlated class distribution within the \mbox{CIFAR100-C} test sequence on the performance of TTA methods. Temporal correlation of classes breaks the assumption of independence and identically distributed (i.i.d.) data and results in performance loss. For this purpose, we utilize Dirichlet distribution with different concentration parameters $\delta$ to sample the class instances, similar to~\citet{note}. The lower the value of $\delta$ is, the more severe the temporal correlation becomes. Moreover, we test the scenario of a long TTA operation by utilizing testing sequences ten times.

\subsection{Methods} 

In our testbed, we include the seven different methods, which differ by the optimized objective, number of hyperparameters, fine-tuned layers, etc. 
\begin{itemize}
    \item TENT~\cite{TENT} is a popular TTA method that includes minimizing the entropy of the predictions while optimizing only batch norm statistics of the model.
    \item SAR~\cite{SAR} additionally applies filtering to remove noisy samples, uses a sharpness-aware optimizer that encourages flat minima, and finally applies a model reset scheme. 
    \item EATA~\cite{eata} uses a filtering scheme that removes not only noisy samples but also redundant ones and additionally introduces a Fisher regularizer~\cite{kirkpatrick2017overcoming} to constrain important model parameters from drastic changes.  
\item  
AdaContrast~\cite{adacontrast} incorporates self-supervised contrastive objective with consistency regularization
 and fine-tuning all weights of the model. 
\item RMT method~\cite{rmt} also uses contrastive objective but additionally uses it to align the learned feature space with the source model combined with novel symmetrical cross-entropy loss. 
\item MEMO \cite{ZhangLF22} performs adaptation on each image independently by minimizing the entropy of the model’s
output distribution across a set of augmented images. This method alleviates the forgetting issue by resetting the model at each timestep, however, this comes at a high computational cost.
\item 
LAME~\cite{boudiafMAB22_parameterfree} is a parameter-free approach, that modifies only the model's output and solves the objective with a concave-convex procedure (i.e., without the need to tune the learning rate).
As a result, hyperparameter selection is not required.
\end{itemize}

\noindent\textbf{Access to the source data}
\textsc{RMT} method additionally assumes a replay mechanism of the source data, which may give this method an advantage over other methods, which were developed with the motivation of not using a memory buffer during training. Thus, we experiment with \textsc{RMT} source-free variant which we denote as \textsc{RMT-SF}.

\subsection{Evaluation Details}

\noindent\textbf{Hyperparameters}
For each of the algorithms, we perform a search over the two most important hyperparameters, that is the learning rate and momentum, similar to~\cite{boudiafMAB22_parameterfree}, with 8 possible configurations. For SAR and EATA we additionally search over the method-specific parameters using random search, which gives 8 additional configurations for those methods. More details are provided in the appendix. 
Each experiment is repeated 3 times. 

\noindent\textbf{Hyperparameter selection} To select hyperparameters, we compute appropriate statistics (entropy loss, consistency loss, etc.) during the adaptation stage. Then, the final hyperparameters are selected based on the accumulated value of the given metric. In case of \textsc{s-acc}, computing accuracy on the source data is performed only after the whole adaptation stage to reduce the computational cost.  

\noindent \textbf{Implementation details} 
For CIFAR100-C, ResNeXt-29~\cite{resnext} architecture is used following the RobustBench benchmark~\cite{croce2021robustbench}.
For remaining datasets, a source pre-trained ResNet-50 is used as int~\cite{rmt}.
We use SGD for all methods (except for SAR, which uses a custom optimizer). We are interested in a batch size equal to one, which allows for fast model adaptation. However, some methods do not work well in the single-image adaptation setting. Therefore, we use batch size equal to ten, which was shown in other works to still work well~\cite{SAR}, and is commonly used in the continual learning for \textit{online} scenarios~\cite{online_continual}. Results with different batch sizes are reported in the appendix (Table~\ref{table:bs_cifar} and ~\ref{table:bs_imagenetr}).

\noindent \textbf{Augmentations}
 Some of the methods use data augmentation at test-time, including AdaContrast, RMT, and MEMO, which are used in their contrastive loss objective. Those augmentations include some of those that were used to generate ImageNet-C and CIFAR100-C benchmarks. 
 Therefore, it is important to note that those augmentations can give an advantage to those methods on those benchmarks. 

\section{Experiments}


Firstly, we analyze our main results in the standard scenario on all datasets. Secondly, we perform experiments on less-common scenarios (e.g., when adaptation is performed on long sequences or images with temporal correlation). Finally, we take a detailed look at the stability of unsupervised selection strategies by also looking at their dynamic.

\begin{table*}[h!] 
\centering
\resizebox{0.72\textwidth}{!}{

\begin{tabular}{l@{\hskip 3pt}l@{\hskip 3pt}c@{\hskip 3pt}c@{\hskip 3pt}c@{\hskip 3pt}c@{\hskip 3pt}c@{\hskip 3pt}c@{\hskip 3pt}c@{\hskip 3pt}c}
\textbf{Dataset} & \textbf{Method} & \textbf{\textsc{s-acc}} & \textbf{\textsc{c-acc}} & \textbf{\textsc{ent}} 
& \textbf{\textsc{con}} 
& \textbf{\textsc{snd}} & \textbf{\textsc{100-rnd}} & \textbf{\textsc{med}} & \textbf{\textsc{oracle}}\\
\midrule
\multirow{7}{*}{\textsc{CIFAR100-C}} & \textsc{source} & \multicolumn{8}{c}{53.55}\\
 & \textsc{tent} 
 &  \textcolor{dontkillmyeyesgreen}{59.93} 
 & 59.56 
 & \textcolor{debianred}{5.32} 
 & 59.56 
 & 40.78
 & 59.17 
 & 54.46 
 & 60.16\\ 

 & \textsc{eata}  
 & \textcolor{dontkillmyeyesgreen}{61.78} 
 & \textcolor{dontkillmyeyesgreen}{61.86} 
 & \textcolor{debianred}{21.01} 
 & 58.60 
 & 39.40
 & 61.95 
 & 60.51 
 & 62.44\\ 

 & \textsc{sar} 
 & \textcolor{dontkillmyeyesgreen}{59.47} 
 & 45.65 
 & \textcolor{debianred}{11.68} 
 & 48.75 
 & 58.26
 & 59.26 
 & 57.03 
 & 60.01\\ 
 
   & \textsc{adacontrast} 
   & \textcolor{dontkillmyeyesgreen}{\textbf{64.43}} 
 & 63.70 
 & 63.69 
 & \textcolor{debianred}{37.97}
 &  \textcolor{dontkillmyeyesgreen}{\textbf{64.43}}
 & 64.43 
 & 48.48 
 & 64.43\\ 

 & \textsc{memo} 
   & 60.66 
 & \textcolor{dontkillmyeyesgreen}{61.08} 
 & \textcolor{dontkillmyeyesgreen}{61.03} 
 & \textcolor{debianred}{59.32}
& 60.66
 & 60.87 
 & 60.82 
 & 61.28\\ 
 
 & \textsc{lame} & \multicolumn{8}{c}{53.01} \\ 

  & \textsc{rmt-sf} & \textcolor{debianred}{59.22} 
 & 60.04 
 & \textcolor{dontkillmyeyesgreen}{60.17} 
 & \textcolor{dontkillmyeyesgreen}{60.17} 
 & \textcolor{debianred}{59.22}
 & 59.93 
 & 60.06 
 & 60.33\\ 
 
\midrule
\multirow{7}{*}{\textsc{ImageNet-C}} & \textsc{source} & \multicolumn{8}{c}{17.97}\\
  & \textsc{tent} 
  & 27.98 
 & \textcolor{dontkillmyeyesgreen}{31.08} 
 & \textcolor{debianred}{1.52} 
 & 29.79 
  & 27.97
 & 30.29 
 & 28.83 
 & 31.37\\ 

 & \textsc{eata} 
 & \textcolor{dontkillmyeyesgreen}{28.87} 
 & 14.60 
 & \textcolor{debianred}{1.96} 
 & 23.39 
  & 19.34
 & 32.25 
 & 18.74 
 & 35.98\\ 

  & \textsc{sar} 
  & 27.14 
 & 28.72 
 & \textcolor{debianred}{8.41} 
 & \textcolor{dontkillmyeyesgreen}{31.19} 
  & 27.14
 & 30.93 
 & 29.30 
 & 31.30\\ 

  & \textsc{adacontrast} 
  & \textcolor{dontkillmyeyesgreen}{33.29} 
 & 31.81 
 & \textcolor{dontkillmyeyesgreen}{33.29} 
 & \textcolor{debianred}{3.90} 
  & 31.79
 & 31.81 
 & 18.94 
 & 33.29\\ 

   & \textsc{memo} 
  & 24.17 
 & 23.93 
 & \textcolor{debianred}{22.55} 
 & \textcolor{dontkillmyeyesgreen}{24.37} 
  & 23.37
 & 23.59 
 & 24.17 
 & 25.01 \\ 

 & \textsc{lame} & \multicolumn{8}{c}{17.72} \\ 

  & \textsc{rmt-sf} 
  & \textcolor{dontkillmyeyesgreen}{28.81} 
 & 27.54 
 & \textcolor{debianred}{23.76} 
 & \textcolor{debianred}{23.76} 
  & \textcolor{dontkillmyeyesgreen}{28.81}
 & 30.51 
 & 29.16 
 & 31.54
 \\ 

\midrule
\multirow{7}{*}{\textsc{DomainNet-126}} & \textsc{source} & \multicolumn{8}{c}{54.71}\\
  & \textsc{tent} 
 & 49.92 
 & \textcolor{dontkillmyeyesgreen}{52.13} 
 & \textcolor{debianred}{8.21} 
 & \textcolor{dontkillmyeyesgreen}{52.13} 
  & 49.91
 & 52.04 
 & 50.32 
 & 52.24\\ 

 & \textsc{eata} 
 & 50.17 
 & \textcolor{dontkillmyeyesgreen}{53.37} 
 & \textcolor{debianred}{19.76} 
 & \textcolor{dontkillmyeyesgreen}{53.59} 
  & 50.17
 & 53.30 
 & 51.88 
 & 54.44\\ 

  & \textsc{sar} 
  & 49.73 
 & 50.91 
 & \textcolor{debianred}{28.48} 
 & \textcolor{dontkillmyeyesgreen}{51.24} 
  & 49.82
 & 51.45 
 & 50.55 
 & 52.75\\ 

  & \textsc{adacontrast} 
  & \textcolor{dontkillmyeyesgreen}{\textbf{56.51}} 
 & 55.69 
 & 46.50 
 & \textcolor{debianred}{19.61} 
  & \textcolor{dontkillmyeyesgreen}{\textbf{56.51}}
 & 54.32 
 & 51.00 
 & 56.51 \\ 

   & \textsc{memo} 
  & \textcolor{dontkillmyeyesgreen}{\textbf{53.98}} 
 & 53.71 
 & \textcolor{debianred}{51.30} 
 & 53.64 
 & \textcolor{dontkillmyeyesgreen}{\textbf{53.98}}
 & 53.908 
 & 53.68 
 & 53.98 \\ 

 & \textsc{lame} & \multicolumn{8}{c}{54.34} \\ 
  
  & \textsc{rmt-sf} 
  & 51.02 
 & \textcolor{dontkillmyeyesgreen}{52.67} 
 & \textcolor{debianred}{46.36} 
 & \textcolor{debianred}{46.36} 
  & 51.01
 & 52.40 
 & 52.42 
 & 52.70\\ 

\midrule 

\multirow{7}{*}{\textsc{ImageNet-R}} & \textsc{source} & \multicolumn{8}{c}{36.17}\\
   & \textsc{tent} 
   & 37.43 
  & \textcolor{dontkillmyeyesgreen}{\textbf{38.92}} 
  & \textcolor{debianred}{11.01} 
  & 38.30 
   & 36.51
  & 38.40 
  & 37.55 
  & 38.92\\ 

 & \textsc{eata} 
 & 37.17 
 & \textcolor{dontkillmyeyesgreen}{\textbf{43.09}} 
 & \textcolor{debianred}{2.35} 
 & 42.27 
  & 37.17
 & 42.03 
 & 39.70 
 & 43.09\\ 

  & \textsc{sar} 
  & 37.09 
 & \textcolor{dontkillmyeyesgreen}{\textbf{41.94}} 
 & \textcolor{debianred}{24.28} 
 & \textcolor{dontkillmyeyesgreen}{\textbf{41.94}} 
  & 36.45
 & 41.31 
 & 37.84 
 & 41.94\\ 

  & \textsc{adacontrast} 
  & 39.23 
 & \textcolor{dontkillmyeyesgreen}{\textbf{39.53}} 
 & 37.94 
 & \textcolor{debianred}{15.00} 
  & 39.23
 & 38.86 
 & 35.80 
 & 39.53\\ 

 & \textsc{memo} 
 & 40.40 
 & \textcolor{dontkillmyeyesgreen}{40.79}
 & \textcolor{debianred}{39.35} 
 & \textcolor{debianred}{39.36} 
  &  40.40
 & 40.49 
 & 40.12 
 & 40.83\\ 

  & \textsc{lame} & \multicolumn{8}{c}{35.95} \\ 

  & \textsc{rmt-sf} 
  & \textcolor{debianred}{36.81} 
 & \textcolor{dontkillmyeyesgreen}{\textbf{40.97}} 
 & \textcolor{dontkillmyeyesgreen}{\textbf{40.97}} 
 & \textcolor{dontkillmyeyesgreen}{\textbf{40.97}} 
  & \textcolor{debianred}{36.81}
 & 40.64 
 & 38.77 
 & 40.97\\ 

\midrule

\multirow{7}{*}{\textsc{ImageNet-V2}} & \textsc{source} & \multicolumn{8}{c}{58.72}\\
  & \textsc{tent} 
 & \textcolor{dontkillmyeyesgreen}{58.69} 
 & \textcolor{dontkillmyeyesgreen}{58.73} 
 & \textcolor{debianred}{55.82 } 
 & 55.99 
  & 57.41 
 & 57.69 
 & 58.70 
 & 58.89\\ 

 & \textsc{eata} 
 & \textcolor{dontkillmyeyesgreen}{58.72} 
 & \textcolor{dontkillmyeyesgreen}{58.72} 
 & \textcolor{debianred}{21.52} 
 & 56.83 
  & 58.67 
 & 58.18 
 & 58.62 
 & 58.93\\ 

  & \textsc{sar} 
  & \textcolor{dontkillmyeyesgreen}{58.70} 
 & \textcolor{dontkillmyeyesgreen}{58.65} 
 & \textcolor{debianred}{57.11} 
 & 57.26 
  & \textcolor{dontkillmyeyesgreen}{58.67}
 & 58.33 
 & 58.71 
 & 58.90\\ 

  & \textsc{adacontrast} 
  & \textcolor{dontkillmyeyesgreen}{\textbf{62.28}} 
 & 62.07 
 & \textcolor{dontkillmyeyesgreen}{\textbf{62.28}} 
 & \textcolor{debianred}{26.68} 
  & \textcolor{dontkillmyeyesgreen}{\textbf{62.28}}
 & 62.17 
 & 58.05 
 & 62.28 \\ 

   & \textsc{memo} 
  & 62.91 
 & \textcolor{dontkillmyeyesgreen}{63.12} 
 & 62.19 
 & \textcolor{debianred}{62.02}
 &   62.91
 & 62.32 
 & 62.66 
 & 63.20 \\ 

 & \textsc{lame} & \multicolumn{8}{c}{1.07} \\ 
  
  & \textsc{rmt-sf} 
  & \textcolor{dontkillmyeyesgreen}{\textbf{58.70}}
 & 58.66 
 & 58.14 
 & \textcolor{debianred}{42.28} 
  & \textcolor{dontkillmyeyesgreen}{\textbf{58.70}}
 &  58.40 
 & 48.36 
 & 58.75\\ 

\midrule
 
\end{tabular}
}

\caption{Final accuracy of evaluated test-time adaptation methods under different model selection strategies. \textcolor{dontkillmyeyesgreen}{Green} color marks the best surrogate strategy (five first columns) for each of the adaptation methods (\textbf{row-wise}), and the \textcolor{debianred}{red} color marks the worst one. \textcolor{dontkillmyeyesgreen}{\textbf{Green bolded text}} marks unsupervised results that match oracle selection strategy. 
Standard deviations in the appendix. 
}
\label{table:main_results_nostd}
\end{table*}

\subsection{Main Results}\label{sec:main}

\begin{figure}[t]
    \centering
    \includegraphics[width=0.99\linewidth]{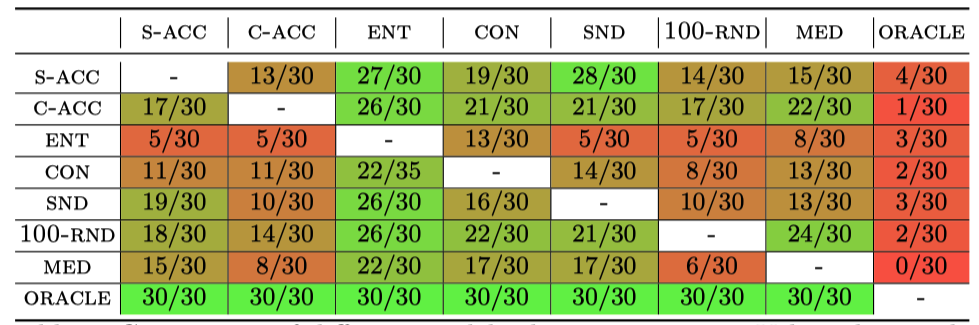}
    \caption{Comparison of different model selection strategies. Values denote the number of experiments (this includes different
adaptation methods and datasets) on which the row model selection method \textbf{outperforms or matches} the column method. No method does consistently better than any other method across all setups or match the ORACLE model selection strategy}
    \label{fig:teaser}
\end{figure}

\begin{figure}[t!]
    \centering
    \includegraphics[width=0.7\linewidth]{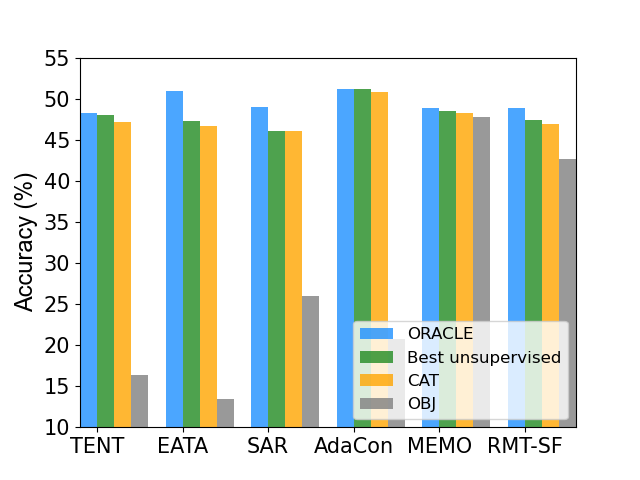}
    \caption{
    TTA results using different hyperparameter selection strategies aggregated over 5 datasets. There is a varying gap between \textsc{Oracle} and unsupervised selection strategies. 
    Using \textsc{cat} strategy to guide selection works fairly well, whereas using the \textsc{obj} strategy performs poorly.
    }
    \label{fig:rec}
\end{figure}

\noindent \textbf{Categorization of selection strategies}
As shown in Fig.~\ref{fig:teaser} and Table~\ref{table:main_results_nostd}, there is no single best-unsupervised metric that scores consistently well. 
An interesting observation can be made when grouping TTA methods by their optimization objective. The CON strategy works fairly well for methods that use entropy-minimization (TENT, SAR, EATA), and the SND strategy is a good choice for methods that use consistency objectives (RMT, AdaContrast, MEMO). We call this strategy \textit{Category} (\textsc{cat}). The opposite strategy of using the optimization objective to guide parameter selection is called \textit{Objective} (\textsc{obj}).
Using \textsc{cat} strategy to guide hyperparameter selection works fairly well, usually a few percents below the \textsc{oracle} results (Fig.~\ref{fig:rec}).
On the contrary, using the \textsc{obj} strategy performs poorly, which seems intuitive as, e.g., optimizing for the entropy using pseudo-labels might lead to degenerate solutions: consistently predicting only one class with high confidence. 

\begin{figure*}[t]
 \hfill
  \begin{minipage}[l]{.32\linewidth}
    \centering
    \includegraphics[width=0.99\linewidth]{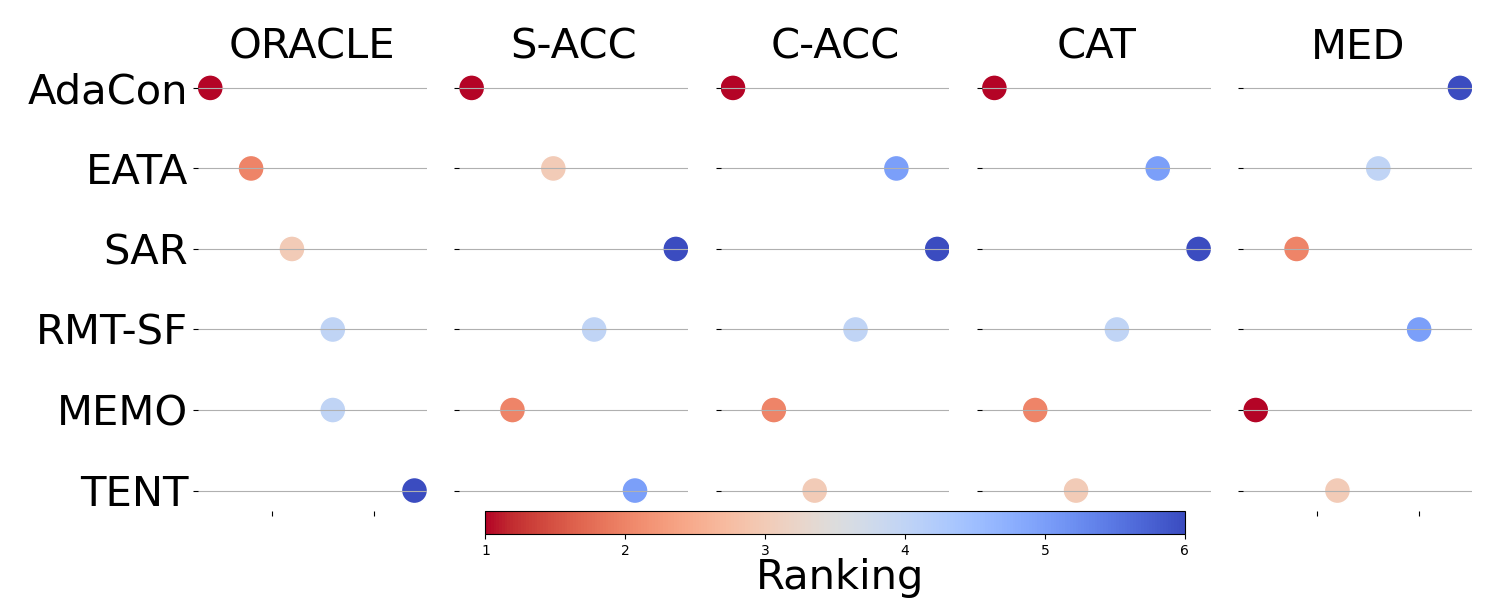}
    \caption{Average ranking of TTA methods under different model selection strategies. Under the proposed hyperparameter selection procedure, simple methods (TENT) or those that are robust to changes in hyperparameters (MEMO) score more favorably compared to the \textsc{oracle} selection. Methods are ordered by a decreasing performance of \textsc{oracle} selection.}
    \label{fig:main_ranking}
  \end{minipage}
  \hfill
  \begin{minipage}[r]{.32\linewidth}
    \centering
    \includegraphics[width=0.99\linewidth]{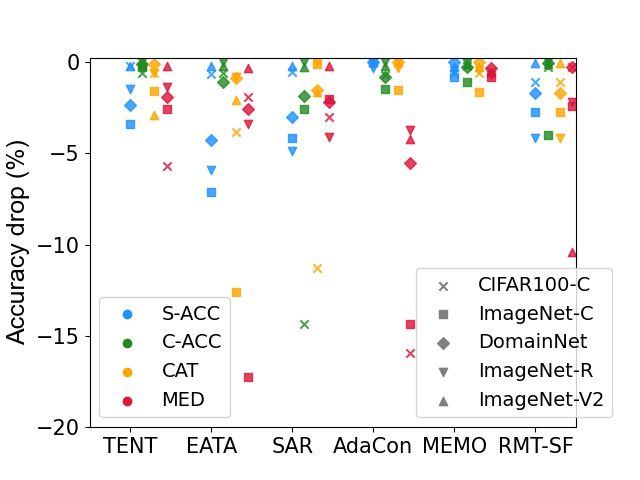}
    \caption{The varying gap between surrogate-based selection strategies and the \textsc{oracle} selection. 
    The MEMO method is robust to changes in hyperparameters, and therefore, it simplifies hyperparameter selection.
    }
    \label{fig:variability}
  \end{minipage}
  \hfill
  \begin{minipage}[r]{.32\linewidth}
    \centering
    \includegraphics[width=0.99\linewidth]{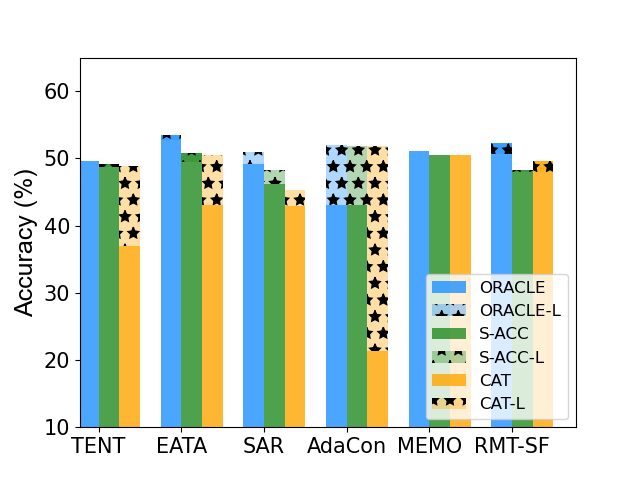}
    \caption{Accuracy gap when testing the methods on longer adaptation scenarios (results with "-L" suffix). \textsc{s-acc} seems to be the most stable measure metric. AdaContrast 
    is severely affected on longer scenarios. }
    \label{fig:long}

  \end{minipage} \hfill
\end{figure*}

\noindent \textbf{TTA methods' performance highly depends on model selection strategy}
While EATA is the 2nd best under the \textsc{oracle} selection strategy, it is outperformed, for example, by TENT (6th method using \textsc{oracle} selection) when using 4 out of 5 surrogate-based metrics (Fig.~\ref{fig:main_ranking}). Similarly, MEMO, which is the 4th method using \textsc{oracle} strategy, scores consistently well when using surrogate-based selection strategies, always being ranked either as the 1st or 2nd best method. AdaContrast is clearly the best method when using surrogate-based metrics on all metrics except for consistency loss. Detailed results in the appendix (Fig.~\ref{fig:hp_average}).

\noindent \textbf{Varying gap between surrogate-based and \textsc{oracle} hyperparameter selection} The gap varies between different methods and datasets (Fig.~\ref{fig:variability}). For example, for the EATA method when using a \textsc{s-acc} strategy the gap varies from 0.21\% on ImageNetV2 to 7.11\% on ImageNet-C (which accounts for 24.62\% of the relative gap). 
One method that allows for robust hyperparameter selection using all unsupervised selection strategies is MEMO. However, this comes with a high computational cost (the method is an order of magnitude slower than other methods, see Table~\ref{table:runtime} in appendix) as the method performs independent adaptation on each data point, helping to alleviate the problem of forgetting~\cite{french1999catastrophic}.

On the contrary, the AdaContrast method using source accuracy for model selection allowed to almost perfectly match the \textsc{oracle} selection strategy on all of the datasets.

\noindent \textbf{Limited access to the target labels helps a lot} As shown in the results in Table~\ref{table:main_results_nostd}, having access to just one hundred target labels (\textsc{100-rnd}) helps to consistently improve over the unsupervised model selection strategies. In most cases, this strategy results in selecting hyperparameters that are within 1\% of accuracy compared to the \textsc{oracle} selection.

\subsection{On stability of surrogate-metrics}\label{sec:stability}

\noindent \textbf{Adaptation difficulty complicates (unsupervised) model selection} 
Further, we increase the difficulty of the testing conditions by incorporating less common, more difficult adaptation scenarios. 
Fig.~\ref{fig:long} shows the performance gap for adaptation when the adaptation is performed on longer scenarios (by repeating the testing data 10 times for CIFAR100-C and ImageNetR datasets, detailed results in the appendix, Table~\ref{table:long_results}).
In general, we can see that the gap between supervised and unsupervised model selection strategies increases. For example, a \textsc{c-acc} metric, which was previously shown to provide stable results, now can easily generate degenerated solutions, e.g., when using TENT (28.68\% accuracy vs. 60.36\% accuracy of \textsc{oracle}, on CIFAR100-C). 
Notably, a \textsc{snd} metric which previously matched \textsc{oracle} results for AdaContrast now performs very poorly on both benchmarks (e.g., only 7.43\% accuracy on ImageNetR). This is an important result that shows that even if we have obtained optimal parameters on some datasets, simply increasing the adaptation length might result in a degenerated model.

Further, we have controlled for the temporal class correlation (Fig.~\ref{fig:dirich} in the appendix). As presented in the figures, the difference between surrogate-based model selection and \textsc{oracle} one increases for some of the selection strategies as we increase task difficulty, which is especially visible for the \textsc{CON} strategy.

\noindent\textbf{Understanding decision margin}
First, we plot the correlation between surrogate metrics and target accuracy (Fig.~\ref{fig:corr_plots_main}, more plots in the appendix). We have previously found that incorporating a metric for model selection that was a part of the adaptation objective leads to degenerate solutions.
In the figure, we can see that optimizing entropy for EATA method initially leads to better selection results (1st column). Nevertheless, once a certain threshold is reached, the model's generated solutions gradually deteriorate in quality. 
What's more, we can see that the transition between optimal and suboptimal solutions is rather smooth. When using the consistency to guide hyperparameter selection solutions close to the optimal can be achieved, but we can see that the margin between viable and suboptimal solutions is rather small.
This small margin explains the varying performance of unsupervised model selection strategies. \textsc{s-acc} is the only unsupervised metric that clearly isolates viable solutions from the degenerated ones. However, it usually does not select the optimal hyperparameters.
Similar observations can be made for other methods (Fig.~\ref{fig:corr_plots_appendix} in the appendix).

\begin{figure*}[ht]
    \centering
    \begin{subfigure}[t]{0.32\textwidth}
        \centering
        \includegraphics[width=\textwidth]{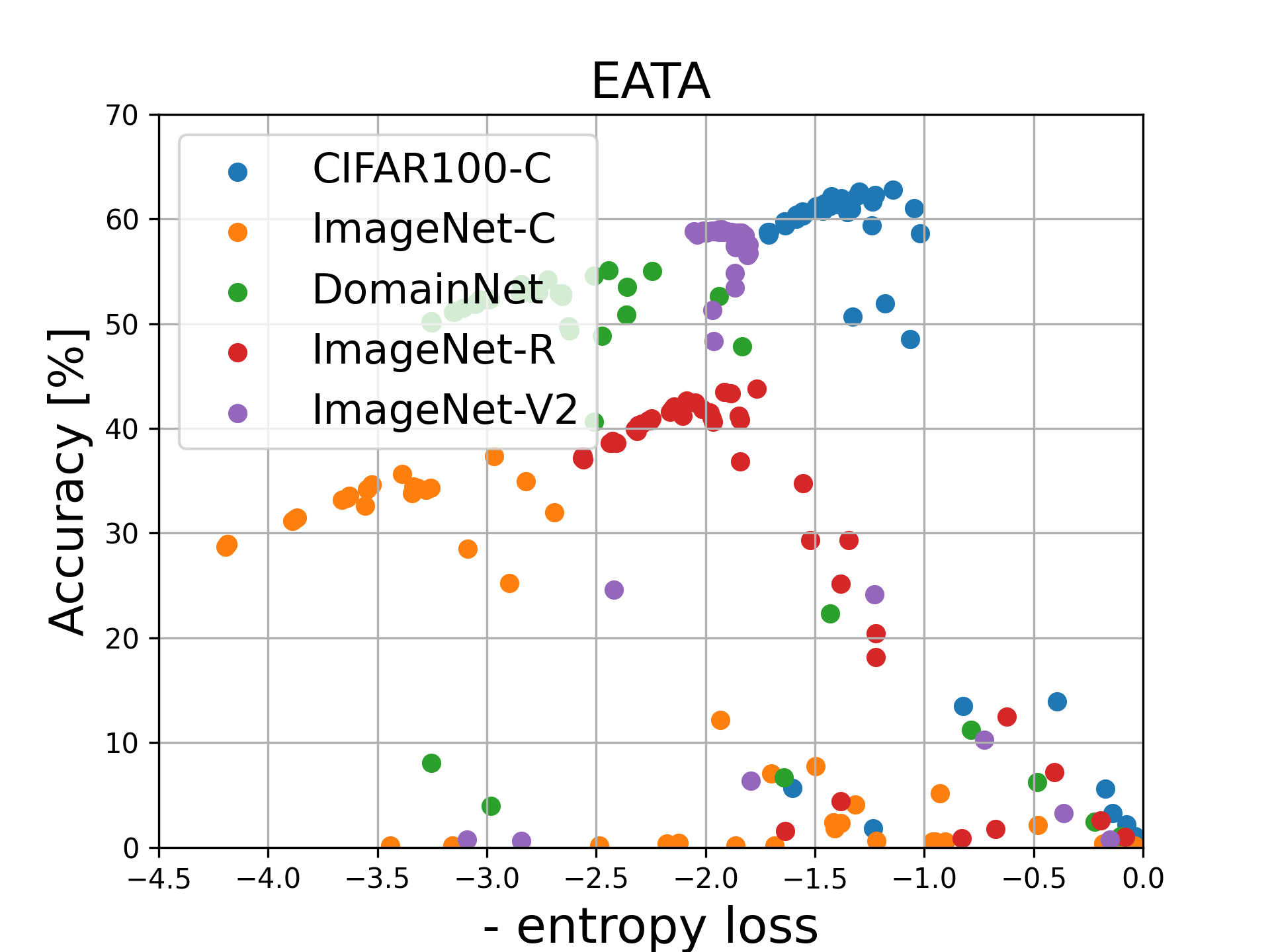}
    \end{subfigure}
    \begin{subfigure}[t]{0.32\textwidth}
        \centering
        \includegraphics[width=\textwidth]{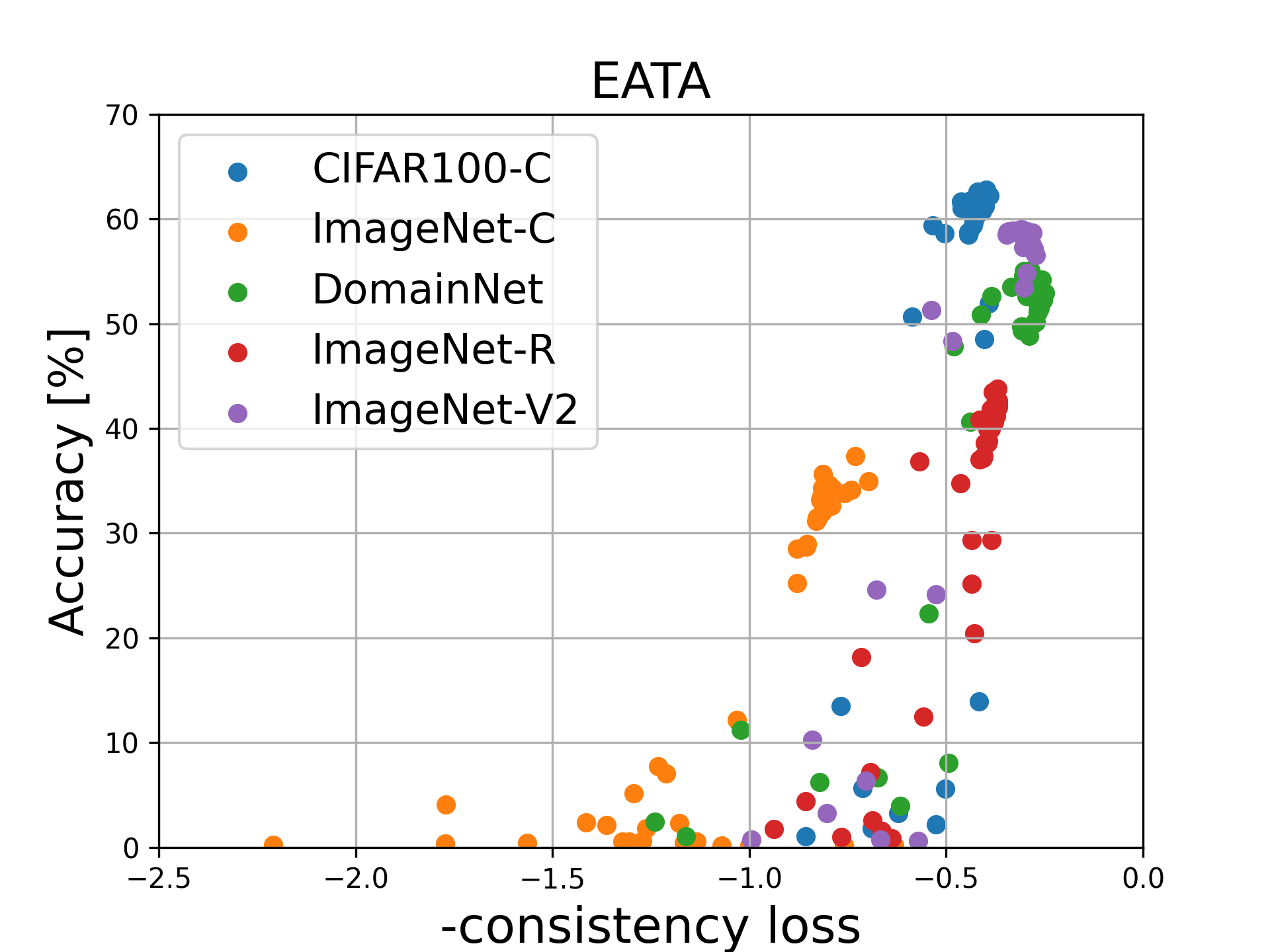}
    \end{subfigure}
     \begin{subfigure}[t]{0.32\textwidth}
        \centering
        \includegraphics[width=\textwidth]{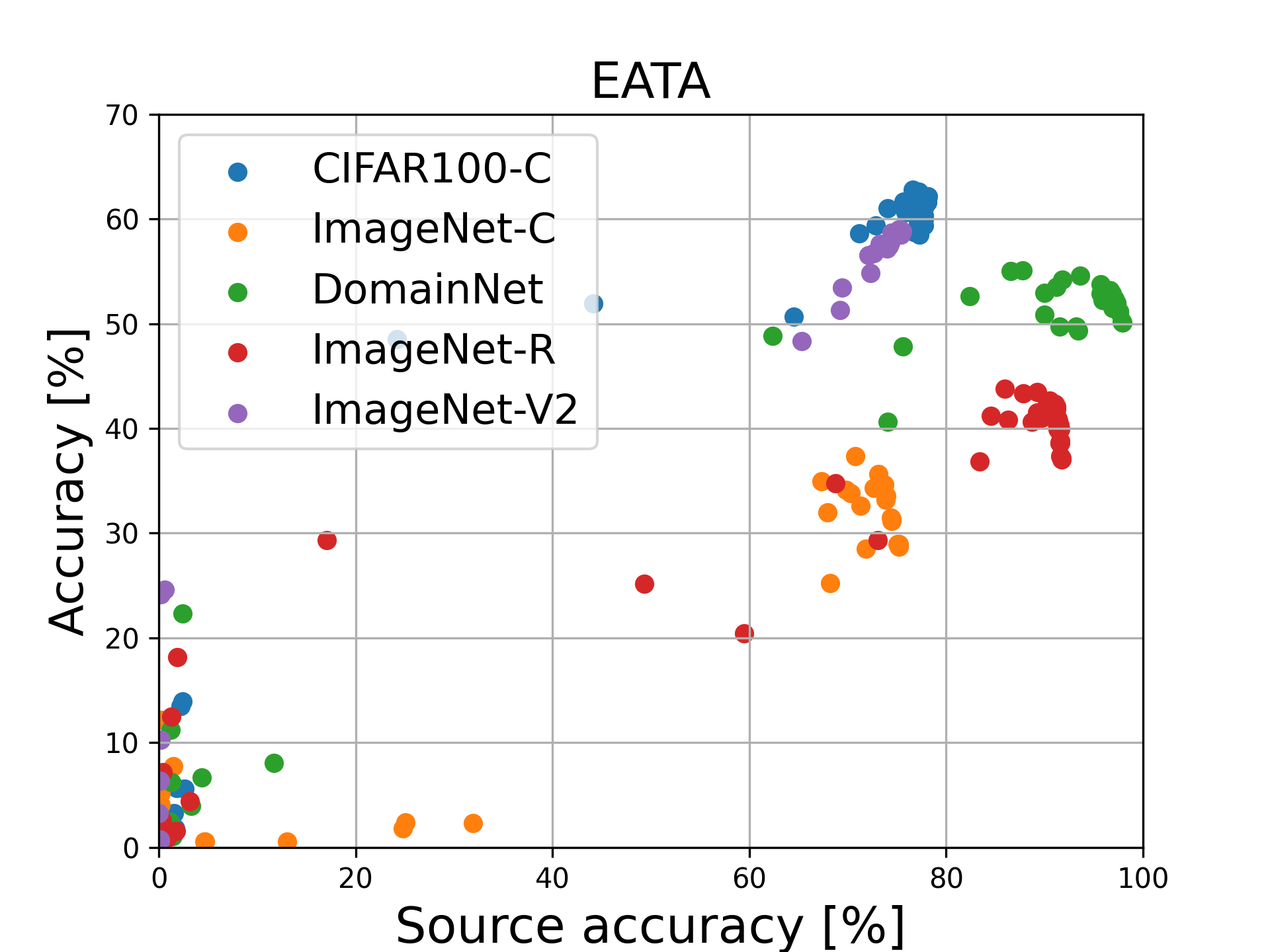}
    \end{subfigure}

    \caption{Correlation plots for the EATA method between target accuracy (y-axis) and selected surrogate measures (x-axes). 
    Selecting the model based on the entropy loss works up to some point, after which minimizing the entropy leads to degenerate solutions with small accuracy. However, there is a smooth transition between favorable to unfavorable solutions exists .
    (Center) Selecting the model based on the consistency loss leads to solutions close to the optimal, however, \textbf{the margin between viable and suboptimal solutions is rather small}. (Right) Selecting the model based on the source accuracy leads to selecting fairly good (but non-optimal) solutions and it clearly isolates degenerated models. 
    }
    \label{fig:corr_plots_main}
\end{figure*}

\begin{figure}[ht]
    \centering
    \includegraphics[width=0.9\linewidth]{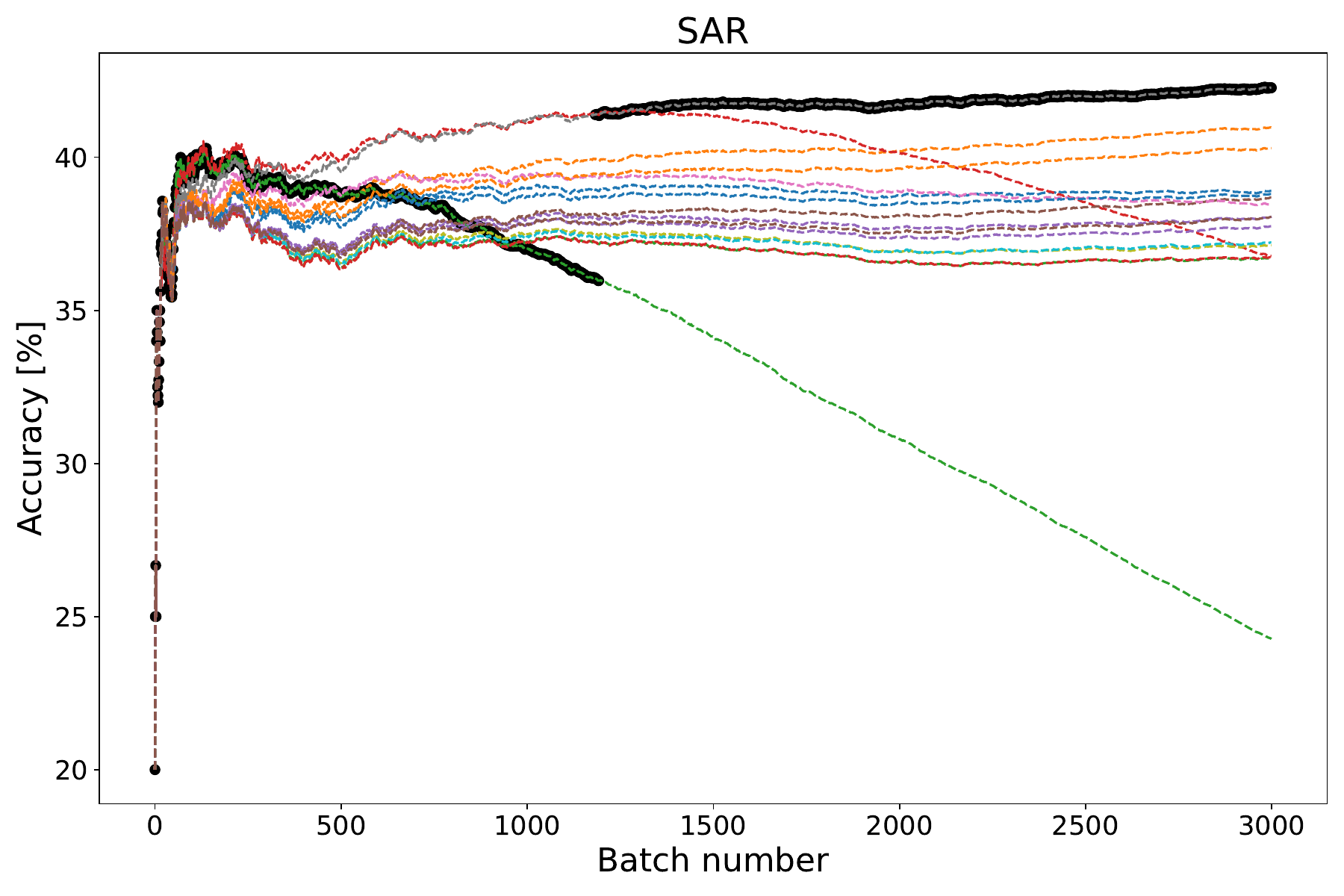}
   
    \caption{Accumulated accuracy on ImageNetR dataset for SAR method with \textsc{CON}-based hyperparameter selection. Different lines relate to runs with different hyperparameters. A black dot ($\bullet$) marks the optimal model according to the selected unsupervised strategy at each timestep.  }
    \label{fig:online}

\end{figure}

\noindent\textbf{Stability in the time-domain}
In Fig.~\ref{fig:online} we plot the accumulated accuracy during adaptation for all hyperparameter runs and mark the best model, according to the chosen unsupervised metric, after each batch (SAR method with \textsc{CON} hyperparameter selection).  
We can see that it leads to optimal model selection in the very early stages of the training, however the performance steadily decreases afterwards. Finally, the metric chooses a different (optimal) model around the 1100th batch. More plots can be found in the appendix Fig.~\ref{fig:online_imagenetr} and ~\ref{fig:online_cifar} (i.e., which shows how using the \textsc{OBJ} selection leads to degenerated solutions).

\section{Discussion}
\noindent\textbf{On the TTA difficulty} As it was shown previously, it is hard to design TTA methods that are robust to changes in hyperparameters~\cite{boudiafMAB22_parameterfree,rmt} or that work well across different sets of adaptation scenarios~\cite{zhao_2023_pitfallsTTA,RDUMB}. We extend those works by showing that it is not only hard to design a method that works well across different adaptation scenarios but also because choosing optimal hyperparameters is a challenging problem. It is not entirely new, given existing research work on unsupervised model selection for domain adaptation~\cite{YouWLJ19,saitoKTSDS21,gulrajaniL21LostDomain}. However, we show here that the time domain, which is inherent to TTA, increases the problem complexity even further:  e.g., as we have shown, parameters that work reasonably well on a given dataset might not work well when we have increased adaptation length. Also, we have shown that the margin between viable and suboptimal solutions is often quite small when using unsupervised hyperparameter selection.  

\newpage

\noindent\textbf{Recommendations} The ultimate goal of TTA is to propose a method that works well in a vast number of scenarios and allows for robust hyperparameter selection procedures. 
Our recommendations
in this area are as follows:
\begin{itemize}
    \item 
    \textbf{Evaluation}
i) Report the results of TTA methods by stating the model selection strategy. This would allow to receive more reliable estimates of performance gains from TTA and could lead to the development of methods that simplify the model selection procedure. ii) Include testing across different scenarios (increased adaptation length or temporal correlation). As shown in our paper, it can lead to not only adaptation performance degradation (i.e., AdaContrast results for longer adaptation scenarios) but also more complicated hyperparameter selection.

\item 
\textbf{Design choices} i) Use information from source-data (e.g., using source prototypes as in RMT or source-data weights importance as in EATA) combined with multi-task loss objective to prevent forgetting and model collapse.
ii) Compare against methods with periodical reset (i.e., MEMO) as they alleviate the problem with forgetting.
\item \textbf{Model selection} i) Notably, using source-data supervision also allows for stable model selection without choosing degenerated solutions (however, usually, a few \% worse than \textsc{oracle} selection). 
One promising approach found in this work was matching the AdaConstrast method with the source-data selection strategy, which in many cases allowed to match the performance of the \textsc{oracle} selection in base scenarios. 
ii) Annotating even a small number of data points from the target domain greatly improves model selection when possible. iii) If no access to any kind of supervised data is allowed, we recommend using the \textsc{cat} selection strategy.
\end{itemize}

\section{Conclusions}
In this work, we evaluated existing test-time adaptation methods on a diverse array of benchmarks using practical model selection strategies, that is, without accessing target labels. Throughout the experiments, we found numerous findings.  We showed that the accuracy of model selection strategies strongly varies across datasets and adaptation methods and have reported a failure mode for each of the selection strategies. 
Out of all evaluated approaches, only the AdaContrast method allows for surrogate-based model selection that matches \textsc{oracle} selection performance, using source data, however, the method itself was inferior in adaptation over long scenarios. Finally, we showed the limitations of the current surrogate-based metrics, which are all inferior to a strategy of using a tiny set of labeled test samples. The study 
shows the importance of the model selection problem in TTA, which we believe did not receive enough attention from the community.

\noindent\textbf{Limitations}
For our hyperparameter selection, we have chosen only the most important parameters (similar to others, i.e., ~\citet{boudiafMAB22_parameterfree}), and we may have omitted some other important hyperparameters.

\bibliography{aaai25}

\newpage

\appendix
{
\noindent
\Huge \textbf{Appendix}
}

\section{Experimental setup}

\label{sec:exp_appendix}
\noindent \textbf{Hyper-parameters}
We use the following hyper-parameter search:
\begin{itemize}
    \item Learning rate $\in$ \{0.001, 0.00025, 0.0000625, 0.0000156\}
    \item Momentum $\in$ \{0.0, 0.9\}
    \item For SAR method: \begin{itemize}
        \item The parameter $E_0$ $\in$ \{0.1, 0.4, 0.8\}
        \item Reset threshold $e_0$ $\in$ \{0.01, 0.05, 0.2, 0.5\}
    \end{itemize}
    \item For EATA method: \begin{itemize}
        \item The parameter $E_0$ $\in$ \{0.1, 0.4, 0.8\}
        \item $\epsilon$ hyperparameter for similarity thresholding $\in$ \{0.05, 0.1, 0.4, 0.8\}
        \item $\beta$ parameter for Fisher regularization $\in$ \{1, 1000, 2000, 5000\}
    \end{itemize}
    \item Access to source data: \begin{itemize}
        \item EATA - we use 2000 source samples to compute Fisher regularization term.
        \item RMT - here we use 10\% of the original source data to compute source prototypes.
    \end{itemize}
\end{itemize}

\noindent \textbf{Data Augmentations} Our calculation of model consistency includes the same augmentations as in CoTTA paper~\cite{COTTA}, which is a combination of color jittering, gaussian blur, gaussian noise, affine transformations, and image cropping. The same augmentations are performed by the RMT method during the adaptation phase.

AdaContrast method use during adaptation augmentations from the MoCo method~\cite{he2020momentum} used in representation learning. The augmentations include image resizing, color jittering, grayscale augmentation, gaussian blur, and horizontal flip. For details about CoTTA and MoCo augmentations please refer to the repository that we have used~\cite{rmt}.

\noindent\textbf{Long adaptation scenarios} Because of the heavy computational requirements of the \textsc{memo} method (see Table~\ref{table:runtime}) in the adaptation over long scenarios (sec.~5.2) we copy the Results of the method on the standard length of adaptation. This is possible because \textsc{memo} performs adaptation on each instance independently. 



\section{Datasets}\label{sec:app_datasets}
CIFAR100-C and ImageNet-C datasets are created by applying 15 different corruption types with 5 severity levels to the images from the test split of the clean CIFAR100
and ImageNet
datasets, creating multiple domains. Following the state-of-the-art TTA works~\cite{COTTA, eata, rmt}, for testing TTA methods we utilize the standard sequence of corruption types with the 5th level of severity sequentially, without mixing the domains. The source models are trained on the clean CIFAR100 or ImageNet dataset.

\mbox{DomainNet-126} is a cleaned from noisy labels subset of the DomainNet dataset. It consists of images of 126 different objects from 4 different domains: real-world photos, clipart images, sketches, and paintings. The source model is trained on actual photos and adapted during test time to the images from the sequence of the remaining domains.

\mbox{ImageNet-R} is composed of various renditions of 200 objects from the original ImageNet dataset, spanning various domains such as art, cartoons, graffiti, embroidery, origami, tattoos, and more. 
This dataset comprises a total of 30000 images. Source data come from the original ImageNet and TTA is tested on all of the rendition types at once.

ImageNetV2 contains images collected by closely following the original, ImageNet labelling protocol. Therefore, it measures the adaptation performance on natural distribution shift.  

\section{Methods runtime}\label{app:runtime}
Efficient TTA method will add a small computational overhead so that the model can perform adaptation in real-time. 
A fair comparison between different methods will take adaptation time into account, e.g., a slower method will perform adaptation on a smaller number of data points, as was already shown in online continual learning~\cite{GhunaimBAAHPTG23} and TTA~\cite{alfarra2024evaluation}. 
As our main goal is to provide a more realistic hyperparameter selection setup to compare  TTA methods, we simply report runtime results in Table~\ref{table:runtime}. 
In general, the fastest methods are those that simply minimize entropy (TENT, SAR, and EATA).  RMT is up to 1.5 times slower, and AdaContrast is 2-3 times slower than TENT. Finally, the MEMO method is \textbf{order of magnitude slower} than competitors since it heavily uses data augmentations and performs independent adaptation to each data sample.
In our experiments, we used NVIDIA A100 graphic cards with 40GB of memory.

\begin{table}[h]
\centering
\setlength{\tabcolsep}{.5em}
\begin{tabular}{lll}

\textbf{Method} & \textbf{CIFAR-100-C}  & \textbf{ImageNet-C} \\ \hline

TENT & 481s & 1256s \\ 
EATA & 651s & 1198s   \\ 
SAR & 686s  & 1061s  \\ 
AdaContrast & 1233s & 3179s   \\ 
MEMO & 22474s & 47344s   \\ 
RMT-SF & 1093s  & 1615s  \\ \hline

\end{tabular}
\caption{TTA runtime for evaluated methods. }
\label{table:runtime}
\end{table}

\section{Results with varying batch size}\label{app:bs}

\begin{table*}[h] 

\resizebox{\textwidth}{!}{

\begin{tabular}{l@{\hskip 3pt}l@{\hskip 3pt}c@{\hskip 3pt}c@{\hskip 3pt}c@{\hskip 3pt}c@{\hskip 3pt}c@{\hskip 3pt}c@{\hskip 3pt}c@{\hskip 3pt}c}
\textbf{Batch Size} & \textbf{Method} & \textbf{\textsc{s-acc}} 
& 
\textbf{\textsc{c-acc}} & \textbf{\textsc{ent}} & \textbf{\textsc{con}} & \textbf{\textsc{snd}} & \textbf{\textsc{100-rnd}} & \textbf{\textsc{med}} & \textbf{\textsc{oracle}}\\
\midrule

- & \textsc{source} & \multicolumn{8}{c}{53.55}\\

- & \textsc{memo} 
   & 60.66$\pm$0.02 
 & \textcolor{dontkillmyeyesgreen}{61.08$\pm$0.14} 
 & \textcolor{dontkillmyeyesgreen}{61.03$\pm$0.17} 
 & \textcolor{debianred}{59.32$\pm$0.01}
& 60.66$\pm$0.02
 & 60.87$\pm$0.20 
 & 60.82$\pm$0.08 
 & 61.28$\pm$0.02\\ 

\midrule
\multirow{6}{*}{\textsc{10}} 

 & \textsc{tent} 
 &  \textcolor{dontkillmyeyesgreen}{59.93$\pm$0.15} 
 & 59.56$\pm$0.2 
 & \textcolor{debianred}{5.32$\pm$0.05} 
 & 59.56$\pm$0.20 
 & 40.78$\pm$25.04
 & 59.17$\pm$0.41 
 & 54.46$\pm$0.51 
 & 60.16$\pm$0.14\\ 

 & \textsc{eata}  
 & \textcolor{dontkillmyeyesgreen}{61.78$\pm$0.40} 
 & \textcolor{dontkillmyeyesgreen}{61.86$\pm$0.47} 
 & \textcolor{debianred}{21.01$\pm$20.12} 
 & 58.60$\pm$4.71 
 & 39.40$\pm$27.15
 & 61.95$\pm$0.57 
 & 60.51$\pm$0.36 
 & 62.44$\pm$0.35\\ 

 & \textsc{sar} 
 & \textcolor{dontkillmyeyesgreen}{59.47$\pm$0.11} 
 & 45.65$\pm$5.83 
 & \textcolor{debianred}{11.68$\pm$5.92} 
 & 48.75$\pm$0.21 
 & 58.26$\pm$0.13
 & 59.26$\pm$0.2 
 & 57.03$\pm$0.27 
 & 60.01$\pm$0.12\\ 
 
   & \textsc{adacontrast} 
   & \textcolor{dontkillmyeyesgreen}   
{\textbf{64.43$\pm$0.01}}
 & 63.70$\pm$0.11 
 & 63.69$\pm$0.11 
 & \textcolor{debianred}{37.97$\pm$0.43}
 &  \textcolor{dontkillmyeyesgreen}{\textbf{64.43$\pm$0.01}}
 & 64.43$\pm$0.01 
 & 48.48$\pm$0.56 
 & 64.43$\pm$0.01\\ 
 
 & \textsc{lame} & \multicolumn{8}{c}{53.014$\pm$0.03} \\ 
  
  & \textsc{rmt-sf} & \textcolor{debianred}{59.22$\pm$0.13} 
 & 60.04$\pm$0.1 
 & \textcolor{dontkillmyeyesgreen}{60.17$\pm$0.07} 
 & \textcolor{dontkillmyeyesgreen}{60.17$\pm$0.07} 
 & \textcolor{debianred}{59.22$\pm$0.13}
 & 59.93$\pm$0.13 
 & 60.06$\pm$0.12 
 & 60.33$\pm$0.11\\ 
 
\midrule
\multirow{6}{*}{\textsc{32}} 

  & \textsc{tent} 
  & 64.90$\pm$0.22 
 & {65.22$\pm$0.03} 
 & \textcolor{debianred}{25.94$\pm$0.10} 
 & \textcolor{dontkillmyeyesgreen}{
 65.43$\pm$0.04} 
  & \textcolor{debianred}{25.94$\pm$0.10}
 & 64.47$\pm$1.62 
 & 64.21$\pm$0.02 
 & 65.62$\pm$0.01\\ 

 & \textsc{eata} 
 & {64.97$\pm$0.14} 
 & \textcolor{dontkillmyeyesgreen}{66.02$\pm$0.26} 
 & {42.75$\pm$28.88} 
 & 65.80$\pm$1.27 
  & \textcolor{debianred}{23.08$\pm$28.30}
 & 65.89$\pm$0.40 
 & 65.04$\pm$0.54 
 & 66.84$\pm$0.24\\ 

 & \textsc{sar} 
 & {64.48$\pm$0.03} 
 & \textcolor{dontkillmyeyesgreen}{ 65.19$\pm$0.21} 
 & \textcolor{debianred}{37.32$\pm$5.32} 
 & \textcolor{dontkillmyeyesgreen}{
 65.14$\pm$0.17} 
  & 53.07$\pm$13.55
 & 65.30$\pm$0.34 
 & 64.28$\pm$0.01 
 & 65.82$\pm$0.05\\ 

  & \textsc{adacontrast} 
  & \textcolor{dontkillmyeyesgreen}{65.82$\pm$0.07} 
 & \textcolor{dontkillmyeyesgreen}{65.82$\pm$0.07} 
 & {61.80$\pm$0.17} 
 & \textcolor{debianred}{ 
 49.21$\pm$0.24} 
  & 64.70$\pm$0.04
 &  62.88$\pm$4.01 
 & 64.14$\pm$0.06 
 & 65.91$\pm$0.02\\ 

 & \textsc{lame} & \multicolumn{8}{c}{17.72$\pm$0.02} \\ 

  & \textsc{rmt-sf} 
  & {65.06$\pm$0.01} 
 & 65.41$\pm$0.03 
 & \textcolor{dontkillmyeyesgreen}{\textbf{67.42$\pm$0.19}} 
 & \textcolor{dontkillmyeyesgreen}{\textbf{67.42$\pm$0.19}} 
  & \textcolor{debianred}{63.20$\pm$0.03}
 & 65.65$\pm$0.97 
 & 65.67$\pm$0.05 
 & 67.42$\pm$0.19
 \\ 

\midrule

\multirow{6}{*}{\textsc{64}} 

  & \textsc{tent} 
 & 66.04$\pm$0.49 
 & \textcolor{dontkillmyeyesgreen}{65.70$\pm$0.04} 
 & \textcolor{debianred}{57.94$\pm$0.30} 
 & {66.54$\pm$0.07} 
  & 63.83$\pm$0.02
 & 66.66$\pm$0.18 
 & 65.39$\pm$0.04 
 & 66.91$\pm$0.06\\ 

 & \textsc{eata} 
 & 66.79$\pm$0.23 
 & \textcolor{dontkillmyeyesgreen}{\textbf{66.70$\pm$0.0.20}} 
 & {44.58$\pm$28.52} 
 & {67.51$\pm$0.11} 
  & \textcolor{debianred} 
  {4.00$\pm$0.18}
 & 53.30$\pm$0.54 
 & 65.88$\pm$0.22 
 & 67.73$\pm$0.18\\ 

  & \textsc{sar} 
  & 66.38$\pm$0.24 
 & \textcolor{dontkillmyeyesgreen}{66.97$\pm$0.34} 
 & \textcolor{debianred}{59.38$\pm$0.29} 
 & {64.77$\pm$0.17} 
  & 63.78$\pm$0.01
 & 65.79$\pm$0.89 
 & 65.00$\pm$0.03 
 & 67.15$\pm$0.11\\ 

  & \textsc{adacontrast} 
  & \textcolor{dontkillmyeyesgreen}{66.20$\pm$0.04} 
 & 65.52$\pm$0.02 
 & \textcolor{debianred}{59.29$\pm$0.06} 
 & \textcolor{debianred}{59.29$\pm$0.06} 
  & 64.37$\pm$0.04
 & 65.73$\pm$0.29 
 & 65.62$\pm$0.01 
 & 66.40$\pm$0.02 \\ 

 & \textsc{lame} & \multicolumn{8}{c}{54.34$\pm$0.01} \\ 
  
  & \textsc{rmt-sf} 
  & 65.93$\pm$0.02 
 & {65.93$\pm$0.02} 
 & \textcolor{dontkillmyeyesgreen}{\textbf{68.49$\pm$0.03}} 
 & \textcolor{dontkillmyeyesgreen}{\textbf{68.49$\pm$0.03}} 
  & \textcolor{debianred}{ 64.02$\pm$0.02}
 & 67.00$\pm$1.10 
 & 66.26$\pm$0.02 
 & 68.49$\pm$0.03\\ 

\midrule
\multirow{6}{*}
{\textsc{128}} 

   
   & \textsc{tent} 
   & \textcolor{dontkillmyeyesgreen}{\textbf{66.99$\pm$0.27}} 
  & {65.52$\pm$0.04} 
  & {66.10$\pm$0.01} 
  & 66.10$\pm$0.01 
   & \textcolor{debianred}{64.40$\pm$0.03}
  & 67.06$\pm$0.68 
  & 65.81$\pm$0.02 
  & 67.59$\pm$0.11\\ 

 & \textsc{eata} 
 & \textcolor{dontkillmyeyesgreen}{67.04$\pm$0.50} 
 & {66.48$\pm$0.22} 
 & {44.53$\pm$25.84} 
 & \textcolor{dontkillmyeyesgreen}{ 67.22$\pm$1.17} 
  & \textcolor{debianred}{25.36$\pm$23.72}
 & 67.07$\pm$0.82 
 & 66.35$\pm$0.31 
 & 68.22$\pm$0.40\\ 

  & \textsc{sar} 
  & 66.70$\pm$0.15 
 & {67.19$\pm$0.50} 
 & {66.13$\pm$0.04} 
 & \textcolor{dontkillmyeyesgreen} {67.47$\pm$0.04} 
  & \textcolor{debianred}{64.38$\pm$0.03}
 & 67.40$\pm$0.05 
 & 65.72$\pm$0.05 
 & 67.62$\pm$0.06\\ 

  & \textsc{adacontrast} 
  & \textcolor{dontkillmyeyesgreen}{66.13$\pm$0.01} 
 & 64.80$\pm$0.03 
 & 63.98$\pm$0.15 
 & {63.98$\pm$0.15} 
  & \textcolor{debianred}{63.87$\pm$0.03}
 & 65.56$\pm$1.09 
 &  65.86$\pm$0.01 
 & 66.66$\pm$0.01\\ 

  & \textsc{lame} & \multicolumn{8}{c}{35.95$\pm$0.01} \\ 

  & \textsc{rmt-sf} 
  & {67.02$\pm$0.36} 
 & {65.93$\pm$0.03} 
 & \textcolor{dontkillmyeyesgreen}{\textbf{68.55$\pm$0.07}} 
 & \textcolor{dontkillmyeyesgreen}{\textbf{68.55$\pm$0.07}} 
  &  \textcolor{debianred}{64.50$\pm$0.02}
 & 67.39$\pm$1.72 
 & 66.26$\pm$0.03 
 & 68.55$\pm$0.07\\ 

\midrule
 
\end{tabular}
}

\caption{CIFAR-100-C: final accuracy of evaluated test-time adaptation methods under different model selection strategies. \textcolor{dontkillmyeyesgreen}{Green} color marks the best surrogate strategy for each of the methods, and the \textcolor{debianred}{red} color marks the worst one. 
\textcolor{dontkillmyeyesgreen}
{\textbf{Green bolded text}} marks unsupervised results that match oracle selection strategy.
}
\label{table:bs_cifar}
\end{table*}

\begin{table*}[h] 

\resizebox{\textwidth}{!}{

\begin{tabular}{l@{\hskip 3pt}l@{\hskip 3pt}c@{\hskip 3pt}c@{\hskip 3pt}c@{\hskip 3pt}c@{\hskip 3pt}c@{\hskip 3pt}c@{\hskip 3pt}c@{\hskip 3pt}c}
\textbf{Batch Size} & \textbf{Method} & \textbf{\textsc{s-acc}} & \textbf{\textsc{c-acc}} & \textbf{\textsc{ent}} & \textbf{\textsc{con}} & \textbf{\textsc{snd}} & \textbf{\textsc{100-rnd}} & \textbf{\textsc{med}} & \textbf{\textsc{oracle}}\\
\midrule

- & \textsc{source} & \multicolumn{8}{c}{36.17}\\

- & \textsc{memo} 
 & 40.41$\pm$0.01 
 & \textcolor{dontkillmyeyesgreen}{\textbf{40.79$\pm$0.04}} 
 & \textcolor{debianred}{39.35$\pm$1.19} 
 & \textcolor{debianred}{39.36$\pm$0.01} 
  & 40.40$\pm$0.01
 & 40.49$\pm$0.12 
 & 40.12$\pm$0.24 
 & 40.83$\pm$0.02\\ 

\midrule

\multirow{6}{*}{\textsc{10}} 

 & \textsc{tent} 
   & 37.43$\pm$0.28 
  & \textcolor{dontkillmyeyesgreen}{\textbf{38.92$\pm$0.07}} 
  & \textcolor{debianred}{11.01$\pm$0.59} 
  & 38.30$\pm$0.17 
   & 36.51$\pm$0.18
  & 38.40$\pm$0.13 
  & 37.55$\pm$0.14 
  & 38.92$\pm$0.07\\ 

 & \textsc{eata} 
 & 37.17$\pm$0.13 
 & \textcolor{dontkillmyeyesgreen}
 {\textbf
 {43.09$\pm$0.75}} 
 & \textcolor{debianred}{2.35$\pm$1.46} 
 & 42.27$\pm$0.18 
  & 37.17$\pm$0.13
 & 42.03$\pm$0.43 
 & 39.70$\pm$0.83 
 & 43.09$\pm$0.76\\ 

  & \textsc{sar} 
  & 37.09$\pm$0.73 
 & \textcolor{dontkillmyeyesgreen}
 {\textbf
 {41.94$\pm$0.24}} 
 & \textcolor{debianred}{24.28$\pm$0.71} 
 & 41.94$\pm$0.23 
  & 36.45$\pm$0.20
 & 41.31$\pm$1.06 
 & 37.84$\pm$0.17 
 & 41.94$\pm$0.24\\ 

  & \textsc{adacontrast} 
  & {39.23$\pm$0.12} 
 & \textcolor{dontkillmyeyesgreen}{\textbf{39.53$\pm$0.14}} 
 & 37.94$\pm$0.26 
 & \textcolor{debianred}{15.00$\pm$0.39} 
  & 39.23$\pm$0.12
 & 38.86$\pm$0.78 
 & 35.80$\pm$0.18 
 & 39.53$\pm$0.14\\ 

  & \textsc{lame} & \multicolumn{8}{c}{35.95$\pm$0.01} \\ 

  & \textsc{rmt-sf} 
  & \textcolor{debianred}{36.81$\pm$0.16} 
 & \textcolor{dontkillmyeyesgreen}
 {\textbf{40.97$\pm$0.32}} 
 & \textcolor{dontkillmyeyesgreen}{\textbf{40.97$\pm$0.32}} 
 & \textcolor{dontkillmyeyesgreen}{\textbf{40.97$\pm$0.32}} 
  & 36.81$\pm$0.16
 & 40.64$\pm$0.78 
 & 38.77$\pm$0.10 
 & 40.97$\pm$0.32\\ 

\midrule

\multirow{6}{*}{\textsc{32}} 

 & \textsc{tent} 
 &  {39.66$\pm$0.43} 
 & 40.15$\pm$0.09 
 & \textcolor{debianred}{37.58$\pm$0.69} 
 & \textcolor{dontkillmyeyesgreen} 
 {42.12$\pm$0.39} 
 & 39.08$\pm$0.10
 & 41.92$\pm$0.67 
 & 39.96$\pm$0.10 
 & 42.34$\pm$0.09\\ 

 & \textsc{eata}  
 & {39.55$\pm$0.47} 
 & {43.70$\pm$0.21} 
 & \textcolor{debianred}{4.67$\pm$3.43} 
 & \textcolor{dontkillmyeyesgreen}{46.05$\pm$0.59} 
 & 39.29$\pm$0.10
 & 45.53$\pm$0.42 
 & 42.30$\pm$0.46 
 & 46.58$\pm$0.45\\ 

 & \textsc{sar} 
 & {39.42$\pm$0.46} 
 & 42.44$\pm$1.25 
 & {42.24$\pm$0.12} 
 & \textcolor{dontkillmyeyesgreen}{45.45$\pm$0.01} 
 & \textcolor{debianred}{39.06$\pm$0.09}
 & 44.77$\pm$0.96 
 & 40.33$\pm$0.09 
 & 45.45$\pm$0.01\\ 
 
   & \textsc{adacontrast} 
   & \textcolor{dontkillmyeyesgreen}{40.02 $\pm$0.45} 
 & \textcolor{dontkillmyeyesgreen}{40.06$\pm$0.11} 
 & 39.90$\pm$0.11 
 & \textcolor{debianred}{30.47$\pm$0.30}
 &  39.42$\pm$0.10
 & 40.20$\pm$0.27 
 & 39.98$\pm$0.11 
 & 40.71$\pm$0.17\\ 
 
 & \textsc{lame} & \multicolumn{8}{c}{53.014$\pm$0.03} \\ 
  
  & \textsc{rmt-sf} & {39.69$\pm$0.39} 
 & 40.53$\pm$0.08 
 & \textcolor{dontkillmyeyesgreen}{\textbf{42.62$\pm$0.14}} 
 & \textcolor{dontkillmyeyesgreen}{\textbf{42.62$\pm$0.14}} 
 & \textcolor{debianred}
 {39.20$\pm$0.10}
 & 42.62$\pm$0.14 
 & 40.82$\pm$0.05 
 & 42.62$\pm$0.14\\ 
 
\midrule
\multirow{6}{*}{\textsc{64}} 

  & \textsc{tent} 
  & 40.03$\pm$0.28 
 & {40.18$\pm$0.11} 
 & \textcolor{dontkillmyeyesgreen}{\textbf{42.79$\pm$0.18}} 
 & 42.61$\pm$0.32 
  & \textcolor{debianred}{39.58$\pm$0.10}
 & 42.79$\pm$0.18 
 & 40.51$\pm$0.09 
 & 42.79$\pm$0.18\\ 

 & \textsc{eata} 
 & {41.12$\pm$0.34} 
 & 42.78$\pm$0.07 
 & \textcolor{debianred}{19.00$\pm$19.64} 
 & \textcolor{dontkillmyeyesgreen} 
 {46.32$\pm$1.46} 
  & 39.69$\pm$0.08
 & 46.59$\pm$1.43 
 & 42.86$\pm$0.49 
 & 47.01$\pm$0.96\\ 

 & \textsc{sar} 
 & {39.96$\pm$0.12} 
 & 41.89$\pm$1.04 
 & {43.16$\pm$0.03} 
 & \textcolor{dontkillmyeyesgreen}{
 \textbf{45.04$\pm$0.06}} 
  & \textcolor{debianred}{39.56$\pm$0.09}
 & 45.04$\pm$0.06 
 & 40.31$\pm$0.08 
 & 45.04$\pm$0.06\\ 

  & \textsc{adacontrast} 
  & {40.02$\pm$0.34} 
 & 39.78$\pm$0.06 
 & \textcolor{dontkillmyeyesgreen}{40.78$\pm$0.27} 
 & \textcolor{debianred}{36.04$\pm$0.35} 
  & 39.32$\pm$0.07
 &  41.32$\pm$0.27 
 & 40.52$\pm$0.07 
 & 41.56$\pm$0.10\\ 

 & \textsc{lame} & \multicolumn{8}{c}{17.72$\pm$0.02} \\ 

  & \textsc{rmt-sf} 
  & {40.19$\pm$0.67} 
 & 40.58$\pm$0.07 
 & \textcolor{dontkillmyeyesgreen}{\textbf{42.36$\pm$0.02}} 
 & \textcolor{dontkillmyeyesgreen}{\textbf{42.36$\pm$0.02}} 
  & \textcolor{debianred}{39.63$\pm$0.09}
 & 41.74$\pm$0.49 
 & 40.85$\pm$0.05 
 & 42.36$\pm$0.02
 \\ 

\midrule

\multirow{6}{*}
{\textsc{128}} 

   
   & \textsc{tent} 
   & 40.60$\pm$0.39 
  & {40.19$\pm$0.10} 
  & \textcolor{dontkillmyeyesgreen}{\textbf{43.29$\pm$0.04}} 
  & \textcolor{dontkillmyeyesgreen}{\textbf{43.29$\pm$0.04}} 
   & \textcolor{debianred}{39.81$\pm$0.08}
  & 42.79$\pm$0.73 
  & 40.39$\pm$0.08 
  & 43.29$\pm$0.04\\ 

 & \textsc{eata} 
 & 40.31 $\pm$0.35 
 & {41.93$\pm$0.08} 
 & \textcolor{debianred}{18.88$\pm$12.04} 
 & \textcolor{dontkillmyeyesgreen}{46.55$\pm$0.83} 
  & 39.88$\pm$0.06
 & 47.43$\pm$0.44 
 & 42.27$\pm$0.14 
 & 47.43$\pm$0.44\\ 

  & \textsc{sar} 
  & 40.64$\pm$0.20 
 & {41.30$\pm$0.62} 
 & {43.03$\pm$0.05} 
 & \textcolor{dontkillmyeyesgreen}{\textbf{44.12$\pm$0.03}} 
  & \textcolor{debianred}{39.80$\pm$0.09}
 & 43.77$\pm$0.53 
 & 40.26$\pm$0.06 
 & 44.12$\pm$0.03\\ 

  & \textsc{adacontrast} 
  & \textcolor{dontkillmyeyesgreen}{40.25$\pm$0.16} 
 & 39.55$\pm$0.03 
 & 39.53$\pm$0.09 
 & {39.53$\pm$0.09} 
  & \textcolor{debianred}{39.29$\pm$0.02}
 & 40.89$\pm$1.05 
 &  40.20$\pm$0.04 
 & 41.80$\pm$0.06\\ 

  & \textsc{lame} & \multicolumn{8}{c}{35.95$\pm$0.01} \\ 

  & \textsc{rmt-sf} 
  & {40.42$\pm$0.05} 
 & {40.49$\pm$0.04} 
 & \textcolor{dontkillmyeyesgreen}{\textbf{42.30$\pm$0.06}} 
 & \textcolor{dontkillmyeyesgreen}{\textbf{42.30$\pm$0.06}} 
  &  \textcolor{debianred}{39.82$\pm$0.08}
 & 41.38$\pm$0.59 
 & 40.73$\pm$0.03 
 & 42.30$\pm$0.06\\ 

\midrule
 
\end{tabular}
}

\caption{ImageNet-R: final accuracy of evaluated test-time adaptation methods under different model selection strategies. \textcolor{dontkillmyeyesgreen}{Green} color marks the best surrogate strategy for each of the methods, and the \textcolor{debianred}{red} color marks the worst one. 
\textcolor{dontkillmyeyesgreen}
{\textbf{Green bolded text}} marks unsupervised results that match oracle selection strategy.
}
\label{table:bs_imagenetr}
\end{table*}

Table~\ref{table:bs_cifar} and Table~\ref{table:bs_imagenetr} shows results on CIFAR-100-C and ImageNet-R datasets when using larger batch sizes. In general, we can see the increase in performance of TTA methods and \textsc{AdaContrast} is no longer the top-performing method. Using \textsc{s-acc} for this method still allows for reliable model selection, however, the distance to using \textsc{oracle} strategy has increased. In general, we can see that \textsc{s-acc} and \textsc{c-acc} allow from quite relable model selection, however, they are quite far from the \textsc{oracle} selection strategy. Once again, the methods ranking under an unsupervised selection strategy can be different from the \textsc{oracle} selection one.
Under large batch size the \textsc{rmt-sf} seems to consistently work very well using \textsc{ent} and \textsc{con} selection strategies, however, this method assumes access to source data (to compute prototypes), which may be limiting in many scenarios.

\section{Additional plots}

\begin{figure*}
    \centering
     \begin{subfigure}[t]{0.32\linewidth}
        \centering
        \includegraphics[width=\linewidth]{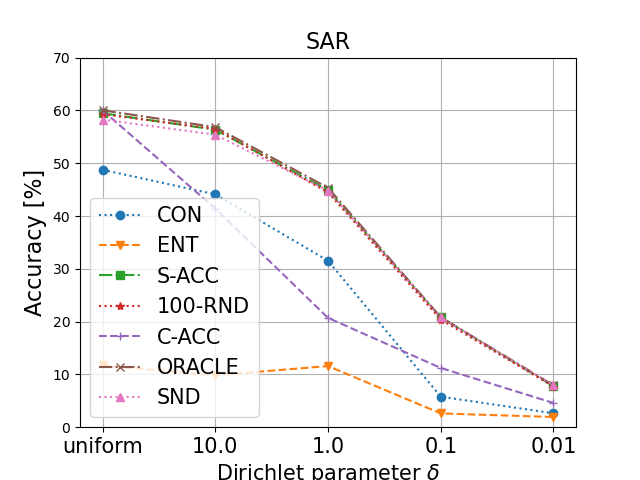}
    \end{subfigure}
     \begin{subfigure}[t]{0.32\linewidth}
        \centering
        \includegraphics[width=\linewidth]{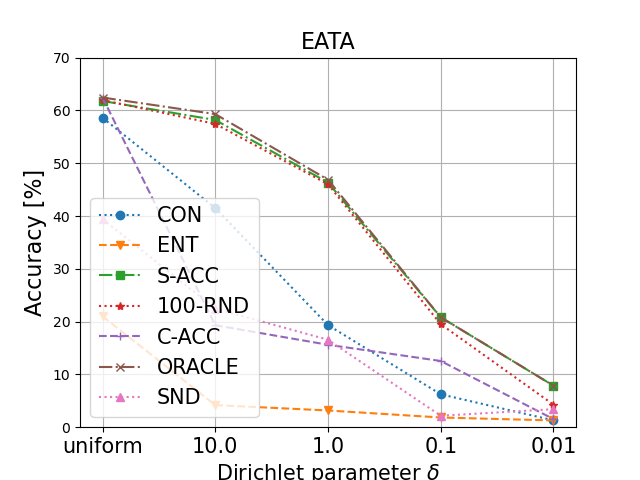}
    \end{subfigure}
    \begin{subfigure}[t]{0.32\linewidth}
        \centering
        \includegraphics[width=\linewidth]{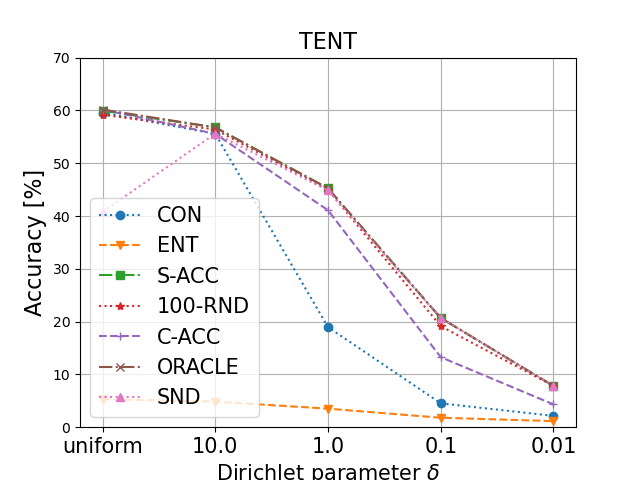}
    \end{subfigure}
    \begin{subfigure}[t]{0.32\linewidth}
        \centering
        \includegraphics[width=\linewidth]{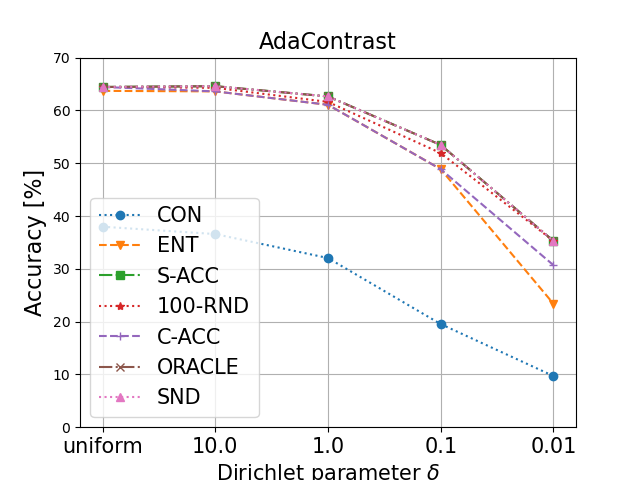}
    \end{subfigure}
     \begin{subfigure}[t]{0.32\linewidth}
        \centering
        \includegraphics[width=\linewidth]{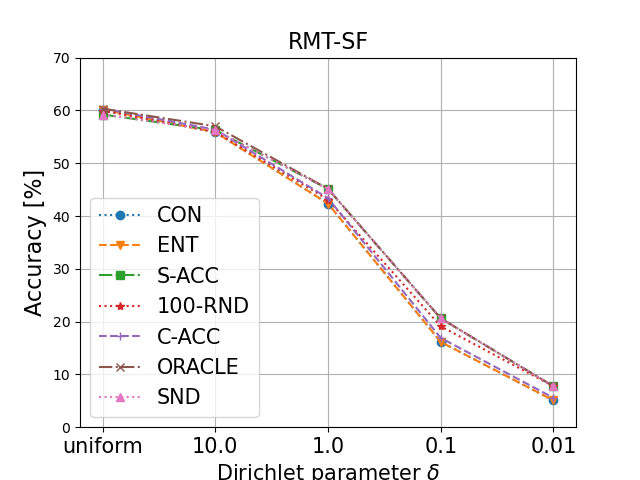}
    \end{subfigure}
    \caption{CIFAR100-C results: the effect of Dirichlet concentration parameter $\delta$ on the model performance for different methods and selection strategies. Even if the surrogate-based metrics match \textsc{oracle} accuracy in the base scenario (uniform class distribution), they may fall short in the non-i.i.d scenario.}
    \label{fig:dirich}
\end{figure*}

\begin{figure*}[t]
    \centering
    \includegraphics[width=1.0\linewidth]{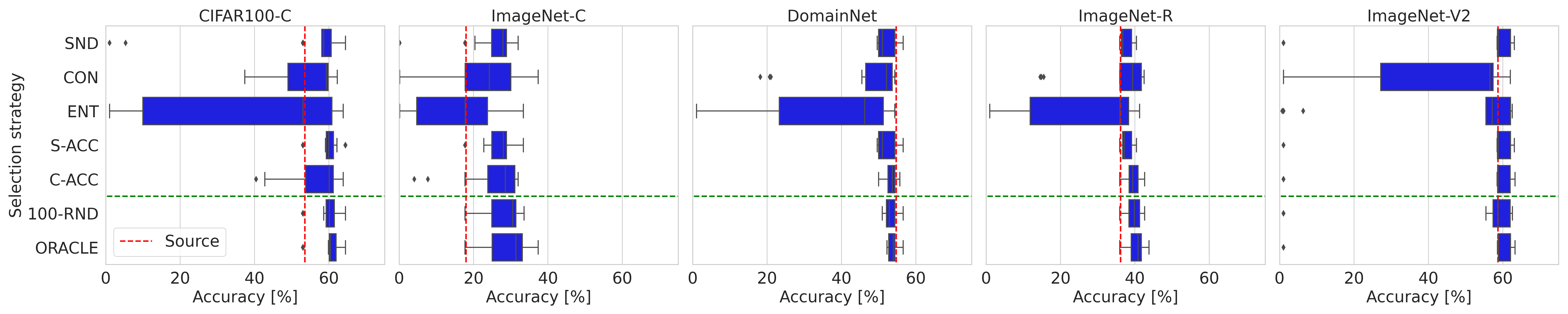}
    \caption{Target accuracy split by selection strategy, aggregated for all evaluated TTA methods. 
    The \textcolor{dontkillmyeyesgreen}{horizontal} line separates selection strategies that use target labels and the \textcolor{debianred}{vertical} one marks \textsc{source} model accuracy.  }
    \label{fig:selection_strategy}
\end{figure*}

\begin{figure*}
    \centering
    \includegraphics[width=1.02\linewidth]{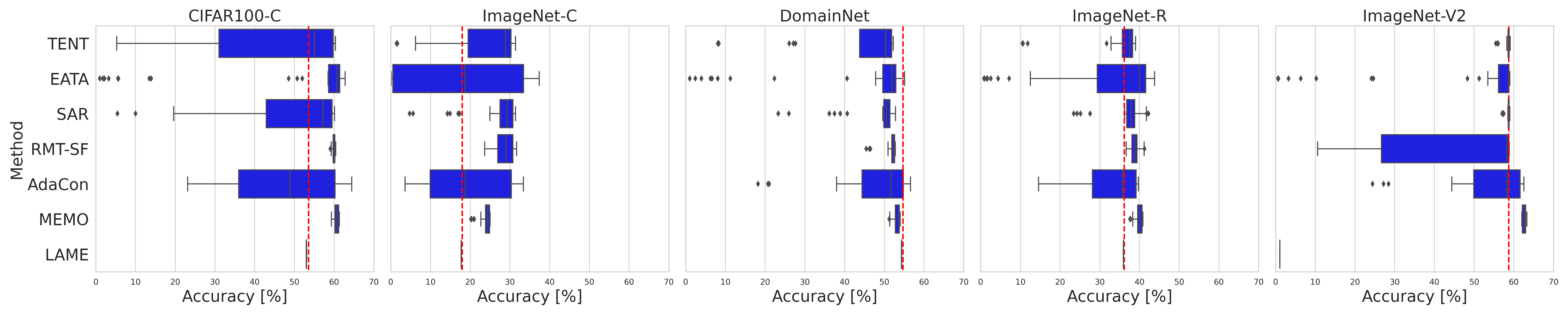}
    \caption{Target accuracy per TTA methods for different datasets, results aggregate all evaluated hyper-parameter configurations. \textcolor{debianred}{Red} vertical line marks the source model accuracy. }
    \label{fig:methods}
\end{figure*}

\begin{figure*}[ht]
    \centering
    \includegraphics[width=0.99\linewidth]{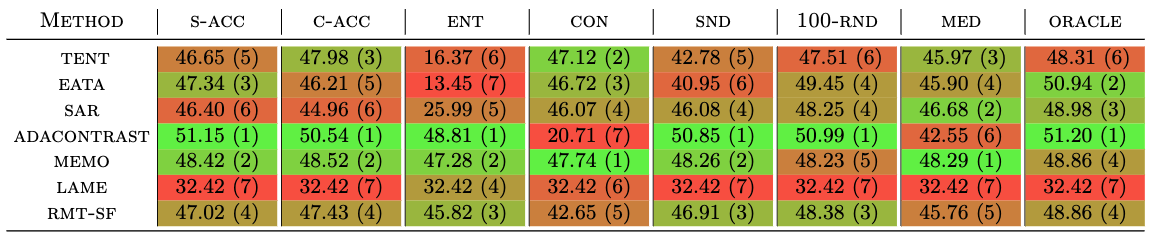}
    \caption{Average accuracy of evaluated test-time adaptation methods under different model selection strategies. Each method
is ranked (value in parenthesis) according to the selection strategies (\textbf{within columns}). The color gradient indicates ranking}
    \label{fig:hp_average}
\end{figure*}

\begin{itemize}
     \item Fig.~\ref{fig:dirich} shows the effect of increased class correlation on the model performance for different methods and selection strategies.
     \item Fig.~\ref{fig:selection_strategy} plots the adaptation accuracy split by selection strategy. Alternatively,  Fig.~\ref{fig:methods} splits the results by TTA method.
     \item Fig.~\ref{fig:hp_average} extends the Fig.~\ref{fig:main_ranking} by showing exact numerical results for all selection strategies.   
     \item Table~\ref{table:long_results} extends the Fig.~\ref{fig:long} by showing exact numerical results.
    \item Fig.~\ref{fig:corr_plots_appendix} show correlation plots for different adaptation methods, not included in the main paper (Fig.~\ref{fig:corr_plots_main}). 
    \item Fig.~\ref{fig:online_imagenetr} and Fig.~\ref{fig:online_cifar} show stability of the unsupervised metrics in the time-domain not included in the main paper (Fig.~\ref{fig:online}).
\end{itemize}

\begin{table*}[t] 
\resizebox{\textwidth}{!}{
\begin{tabular}{l@{\hskip 3pt}l@{\hskip 3pt}c@{\hskip 3pt}c@{\hskip 3pt}c@{\hskip 3pt}
c@{\hskip 3pt}
c@{\hskip 3pt}c@{\hskip 3pt}c}
\textbf{Dataset} 
& \textbf{Method}
& \textbf{\textsc{s-acc}} 
& \textbf{\textsc{c-acc}} 
& \textbf{\textsc{ent}} 
& \textbf{\textsc{con}} 
& \textbf{\textsc{snd}} 
& 
\textbf{\textsc{median}} 
& \textbf{\textsc{oracle}}\\
\midrule

   \multirow{7}{*}{\textsc{CIFAR100-C long}} & \textsc{source} & \multicolumn{7}{c}{53.55}\\
   
   & \textsc{tent}
   & \textcolor{dontkillmyeyesgreen}{\textbf{60.36$\pm$0.16}} 
  & 28.68$\pm$0.7 
  & \textcolor{debianred}{1.90$\pm$0.21} 
  & 36.09$\pm$23.99 
  & 1.91$\pm$0.21 
  & 13.90$\pm$0.41 
  & 60.36$\pm$0.16 \\ 

 & \textsc{eata}
 & 61.59$\pm$0.91
 & \textcolor{dontkillmyeyesgreen}{\textbf{62.17$\pm$0.94}} 
 & \textcolor{debianred}{3.41$\pm$2.45} 
 & 42.64$\pm$25.36 
 & 22.61$\pm$26.58 
 & 52.11$\pm$12.37 
 & 62.47$\pm$0.41 \\ 

  & \textsc{sar} 
  & \textcolor{dontkillmyeyesgreen}{\textbf{54.59$\pm$7.92}} 
 & 46.98$\pm$1.13 
 & \textcolor{debianred}{8.17$\pm$2.82} 
 & 42.22$\pm$1.14 
 & 53.72$\pm$9.14 
 & 40.85$\pm$3.25 
 & 54.59$\pm$7.92 \\ 

 & \textsc{adacontrast} 
 & \textcolor{dontkillmyeyesgreen}{\textbf{49.45$\pm$9.73}} 
 & \textcolor{dontkillmyeyesgreen}{\textbf{49.45$\pm$9.73}} 
 & \textcolor{debianred}{42.92$\pm$18.98} 
 & 15.63$\pm$0.78 
 & 35.26$\pm$0.75 
 & 19.88$\pm$1.36 
 & 49.45$\pm$9.73 \\ 

    & \textsc{memo*} 
   & 60.66$\pm$0.02 
 & \textcolor{dontkillmyeyesgreen}{61.08$\pm$0.14} 
 & \textcolor{dontkillmyeyesgreen}{61.03$\pm$0.17} 
 & \textcolor{debianred}{59.32$\pm$0.01}
& 60.66$\pm$0.02
 & 60.82$\pm$0.08 
 & 61.28$\pm$0.02\\ 

  & \textsc{rmt-sf} 
  & 58.02$\pm$3.84 
 & \textcolor{debianred}{31.53$\pm$0.15} 
 & \textcolor{dontkillmyeyesgreen}{\textbf{60.69$\pm$0.12}} 
 & \textcolor{dontkillmyeyesgreen}{\textbf{60.69$\pm$0.12}} 
 & \textcolor{dontkillmyeyesgreen}{\textbf{60.69$\pm$0.12}} 
 & 56.02$\pm$0.92 
 & 60.81$\pm$0.06\\ 

\midrule

\multirow{7}{*}{\textsc{ImageNet-R Long}} & \textsc{source} & \multicolumn{7}{c}{36.17}\\
   
   & \textsc{tent}
   & 37.92$\pm$0.15 
  & 20.20$\pm$0.34 
  & \textcolor{debianred}{1.84$\pm$0.22} 
  & \textcolor{dontkillmyeyesgreen}{37.78$\pm$0.85} 
  & 37.92$\pm$0.15 
  & 27.12$\pm$0.25 
  & 38.94$\pm$0.11 \\ 

 & \textsc{eata}
 & 40.03$\pm$0.12
 & {41.94$\pm$2.80} 
 & \textcolor{debianred}{3.41$\pm$2.65} 
 & \textcolor{dontkillmyeyesgreen}{43.40$\pm$1.95} 
 & 26.98$\pm$18.48 
 & 34.17$\pm$8.60 
 & 44.49$\pm$1.28 \\ 

  & \textsc{sar} 
  & {37.78$\pm$0.18} 
 & 41.37$\pm$1.60 
 & \textcolor{debianred}{7.31$\pm$4.21} 
 & \textcolor{dontkillmyeyesgreen}{\textbf{43.62$\pm$0.11}}
 & 37.7$\pm$0.18 
 & 38.58$\pm$0.13
 & 43.62$\pm$0.11 \\ 

 & \textsc{adacontrast} 
 & \textcolor{dontkillmyeyesgreen}{\textbf{36.81$\pm$0.16}}
 & 26.97$\pm$0.19 
 & \textcolor{dontkillmyeyesgreen}{\textbf{36.81$\pm$0.16}} 
 & \textcolor{dontkillmyeyesgreen}{\textbf{36.81$\pm$0.16}} 
 & \textcolor{debianred}{7.43$\pm$0.49} 
 & 11.82$\pm$0.05 
 & 36.81$\pm$0.16 \\ 

 & \textsc{memo} 
 & 40.41$\pm$0.01 
 & \textcolor{dontkillmyeyesgreen}{\textbf{40.79$\pm$0.04 }}
 & \textcolor{debianred}{39.35$\pm$1.19} 
 & \textcolor{debianred}{39.36$\pm$0.01} 
  &  40.40$\pm$0.01
 & 40.12$\pm$0.24 
 & 40.83$\pm$0.02\\ 

  & \textsc{rmt-sf} 
  & \textcolor{debianred}{38.57$\pm$0.05} 
 & \textcolor{dontkillmyeyesgreen}{\textbf{43.87$\pm$0.19}}
 & \textcolor{dontkillmyeyesgreen}{\textbf{43.87$\pm$0.19}} 
 & \textcolor{dontkillmyeyesgreen}{\textbf{43.87$\pm$0.19}} 
 & \textcolor{debianred}{38.57$\pm$0.05} 
 & 38.30$\pm$0.16 
 & 43.87$\pm$0.19\\ 

\midrule
 
\end{tabular}%
}

\caption{Target accuracy for different hyper-parameter selection strategies when adaptation length is multiplied 10 times. Using oracle-chosen parameters from standard ImageNet-C benchmark (\textsc{c-acc} strategy) performs poorly for SAR and EATA. Also, greater variance in results can be observed. }
\label{table:long_results}
\vspace{-0.5cm}
\end{table*}

\newcommand{\corrwidth}{0.31}

\begin{figure*}[h]
    \centering
    \begin{subfigure}{\corrwidth\textwidth}
        \centering
        \includegraphics[width=\textwidth]{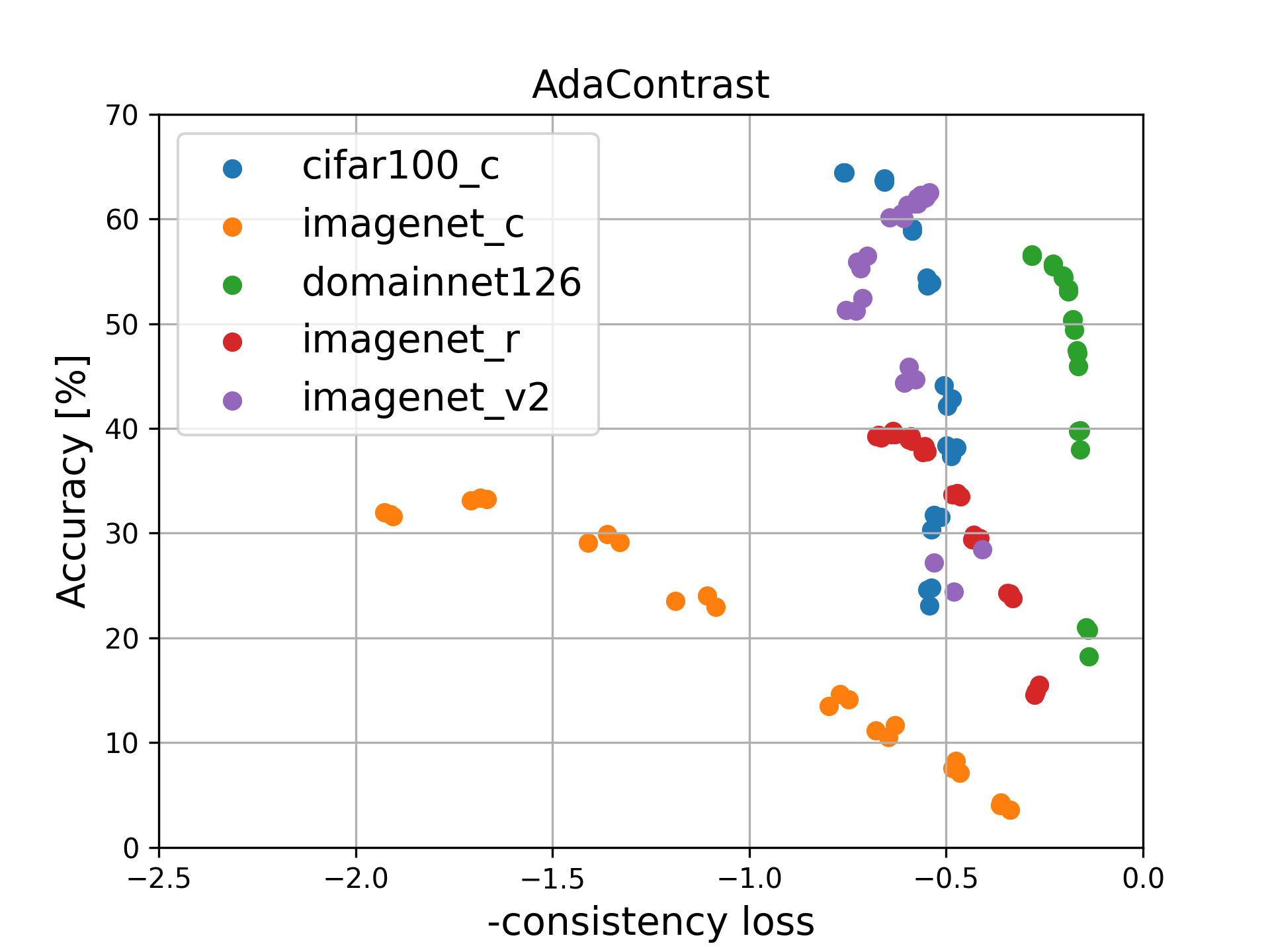}
    \end{subfigure}
    \begin{subfigure}{\corrwidth\textwidth}
        \centering
        \includegraphics[width=\textwidth]{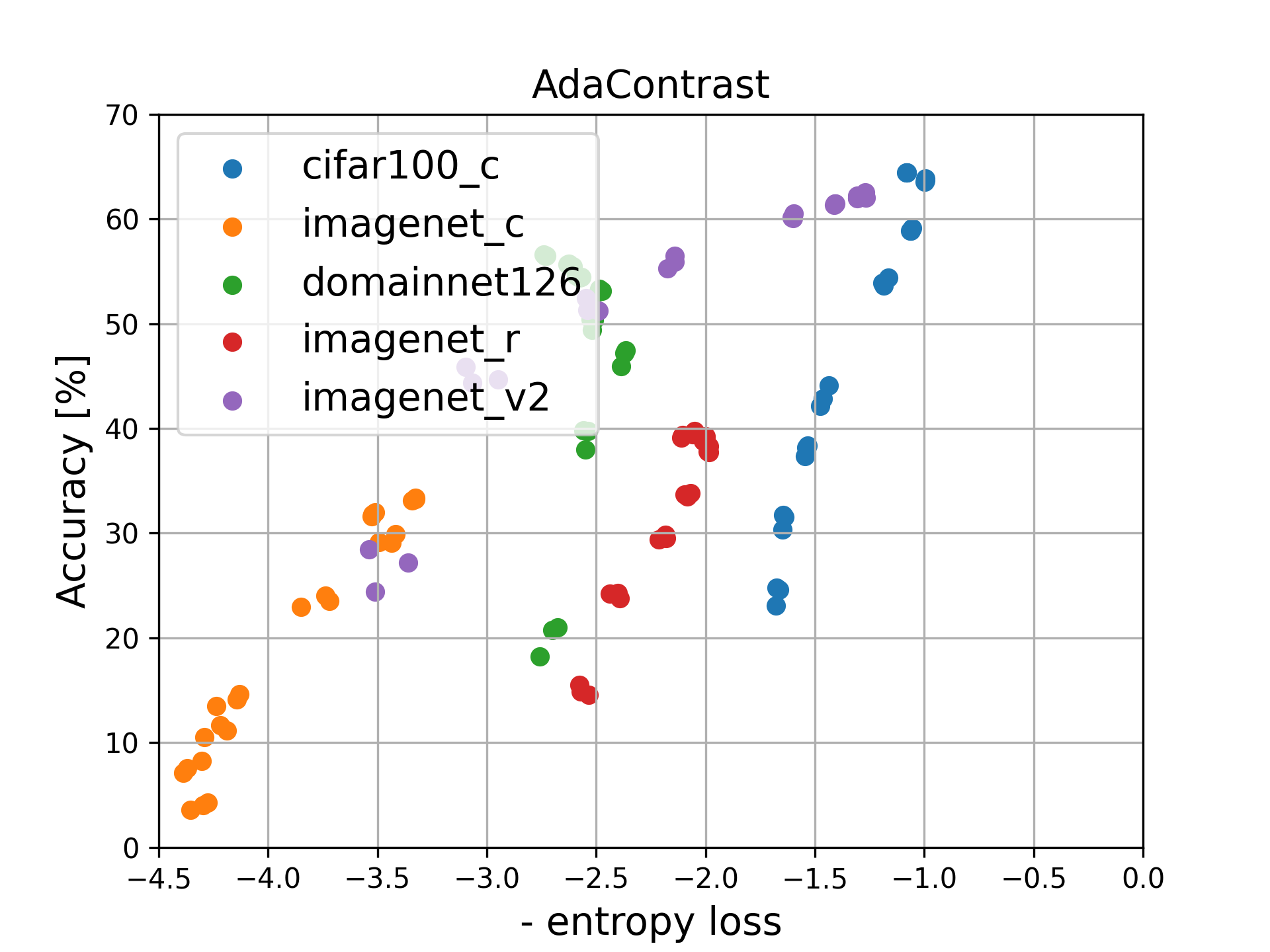}
    \end{subfigure}
     \begin{subfigure}{\corrwidth\textwidth}
        \centering
        \includegraphics[width=\textwidth]{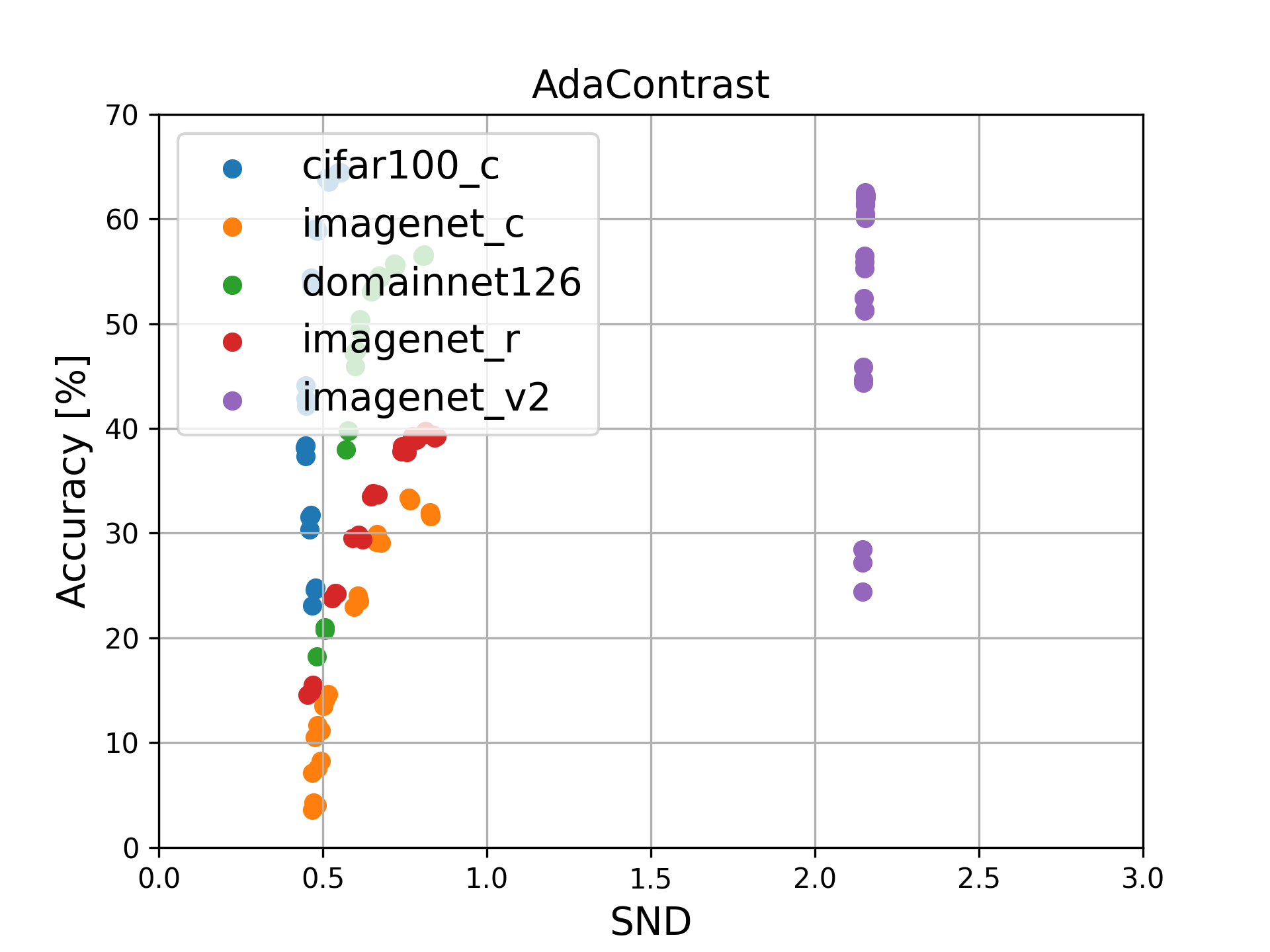}
    \end{subfigure}
    \begin{subfigure}{\corrwidth\textwidth}
        \centering
        \includegraphics[width=\textwidth]{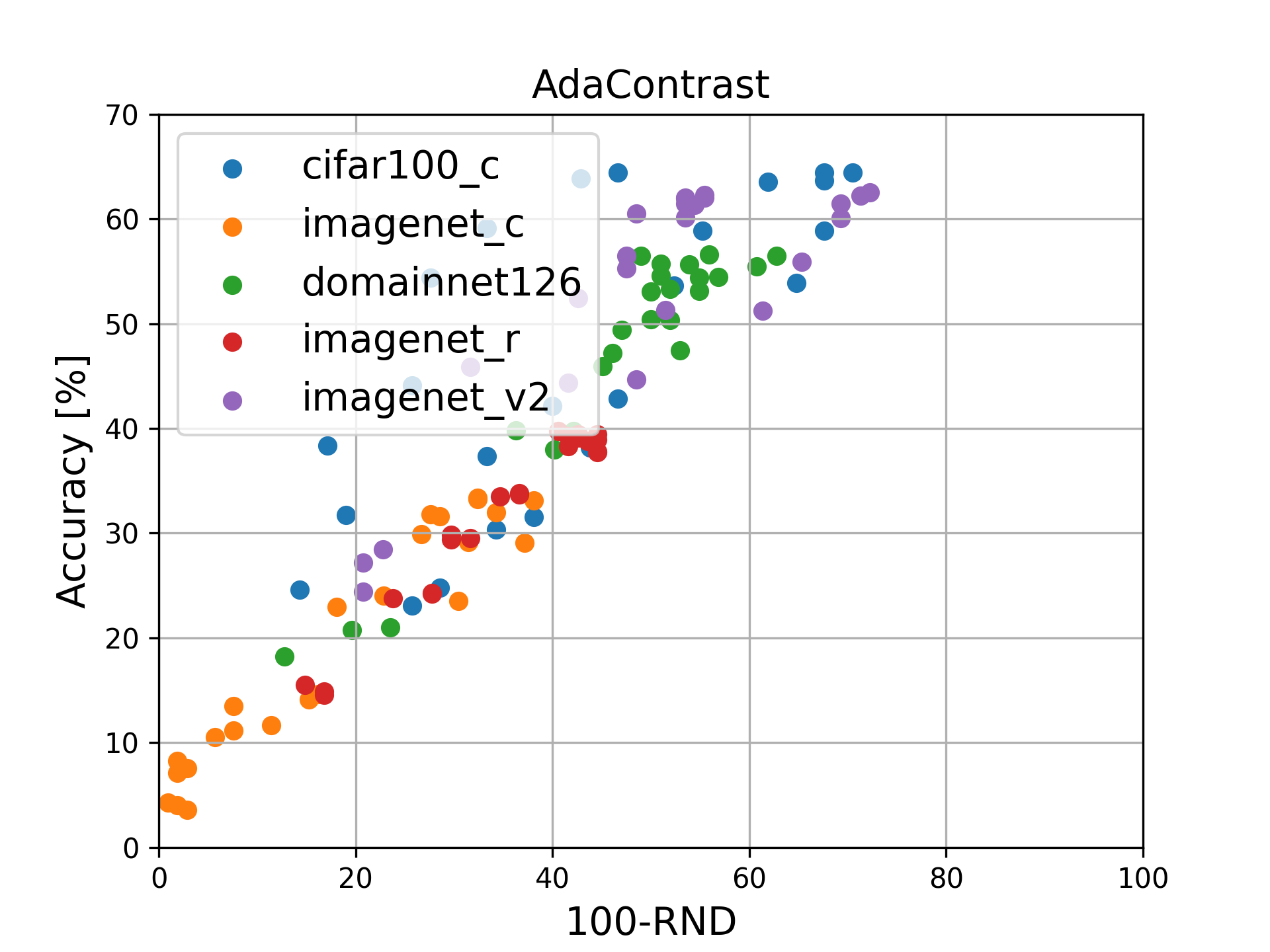}
    \end{subfigure}
    \begin{subfigure}{\corrwidth\textwidth}
        \centering
        \includegraphics[width=\textwidth]{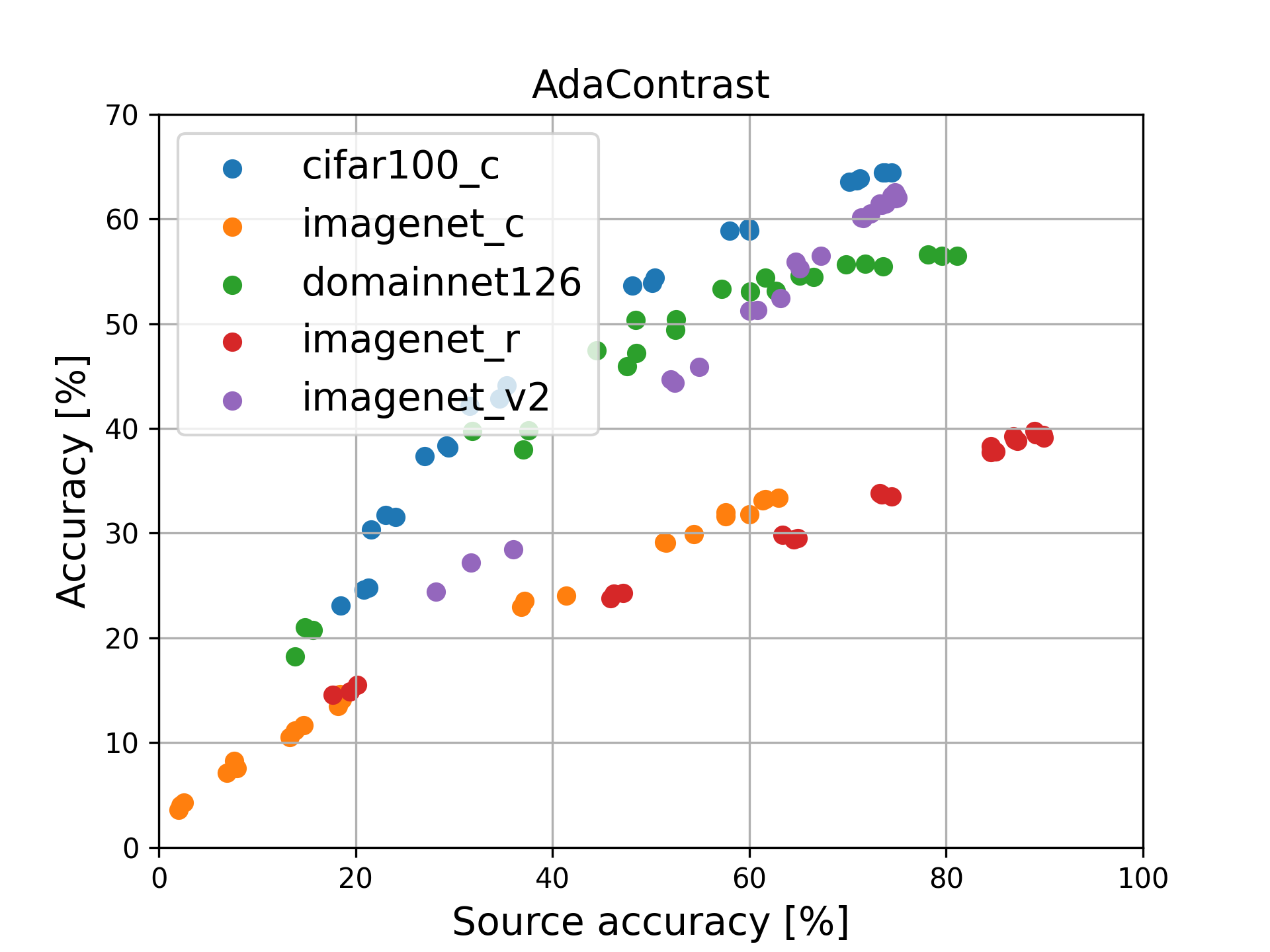}
    \end{subfigure} \\
    
    \begin{subfigure}{\corrwidth\textwidth}
        \centering
        \includegraphics[width=\textwidth]{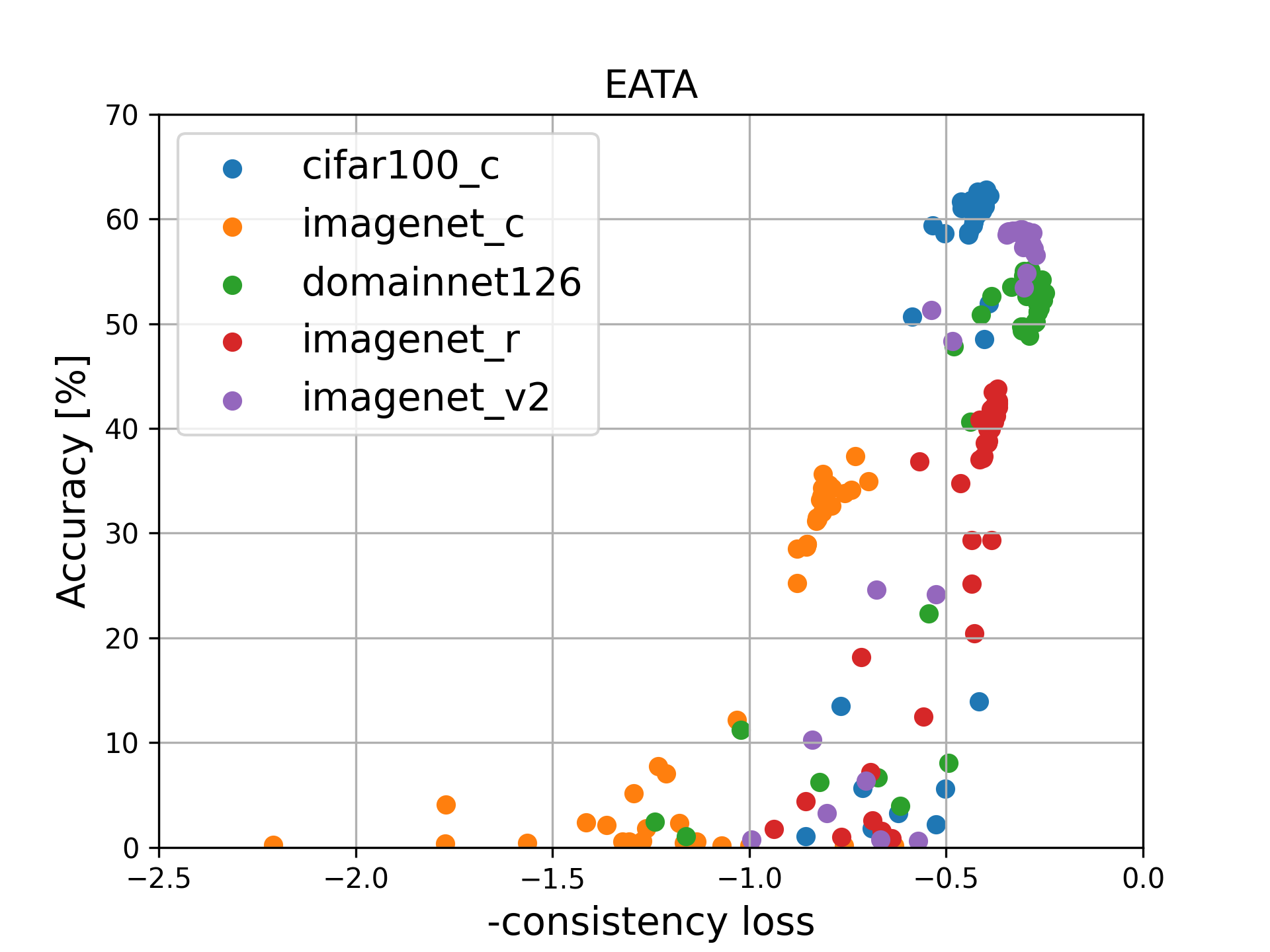}
    \end{subfigure}
    \begin{subfigure}{\corrwidth\textwidth}
        \centering
        \includegraphics[width=\textwidth]{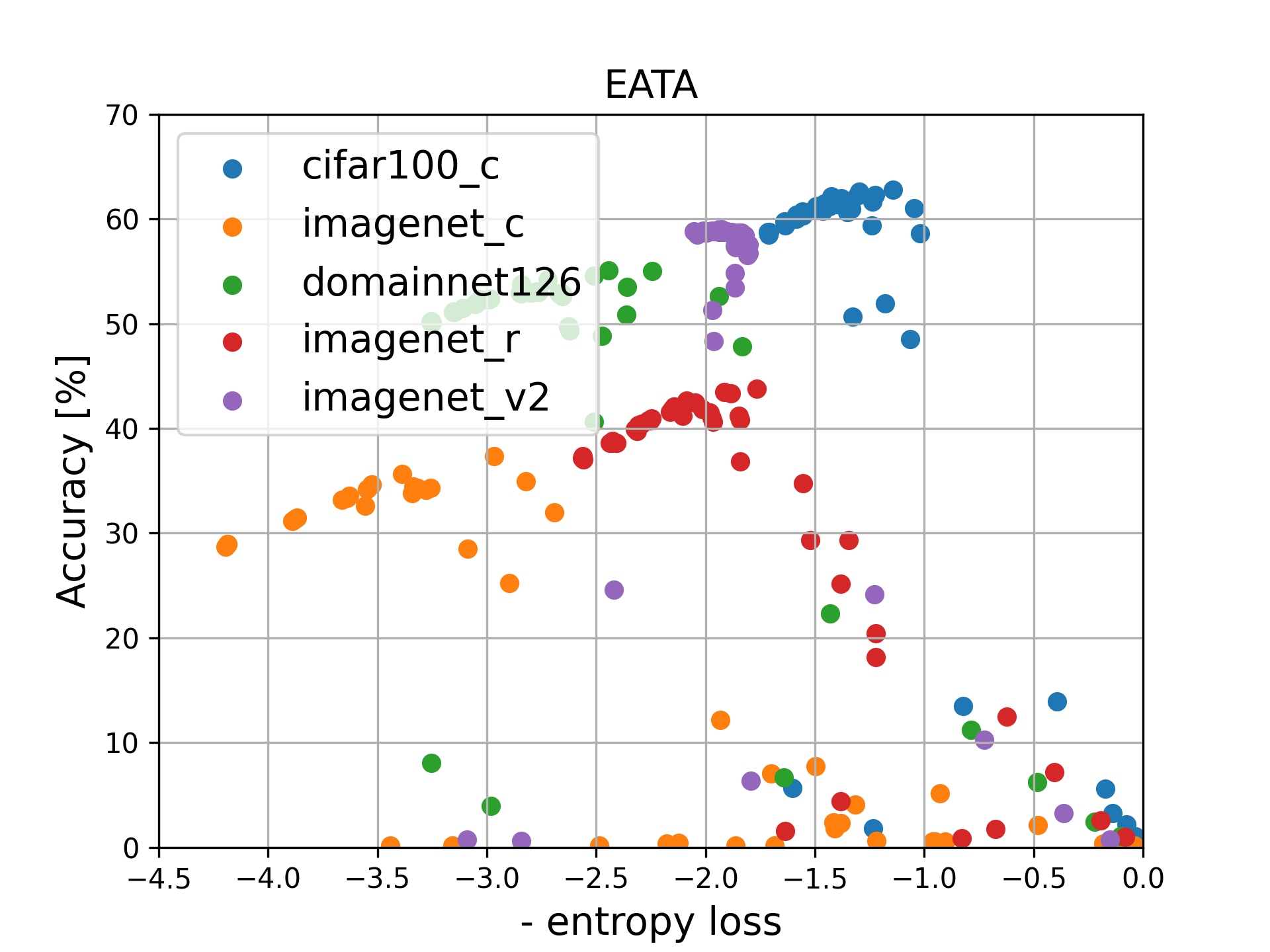}
    \end{subfigure}
    \begin{subfigure}[t]{\corrwidth\textwidth}
        \centering
        \includegraphics[width=\textwidth]{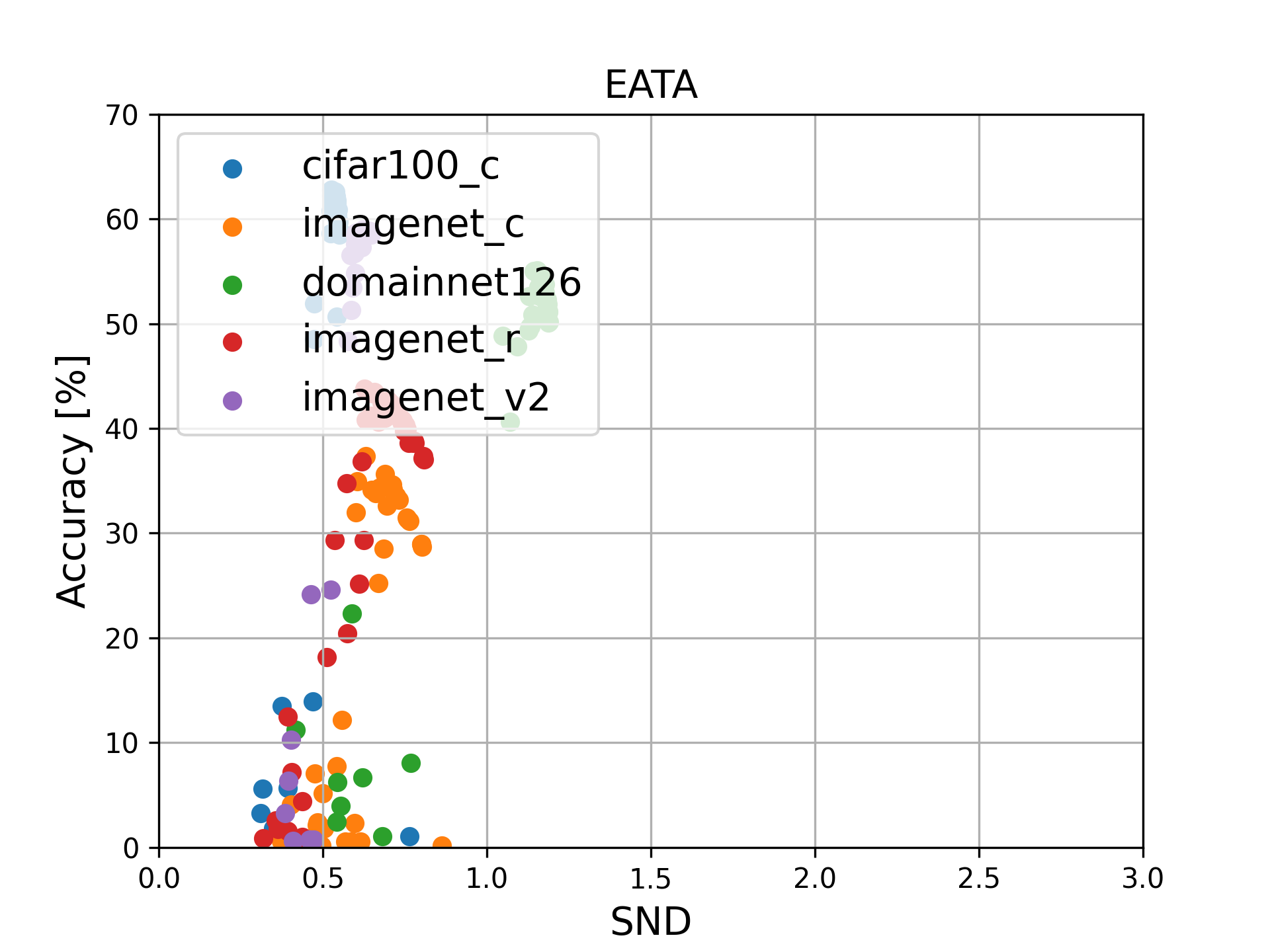}
    \end{subfigure}
    \begin{subfigure}{\corrwidth\textwidth}
        \centering
        \includegraphics[width=\textwidth]{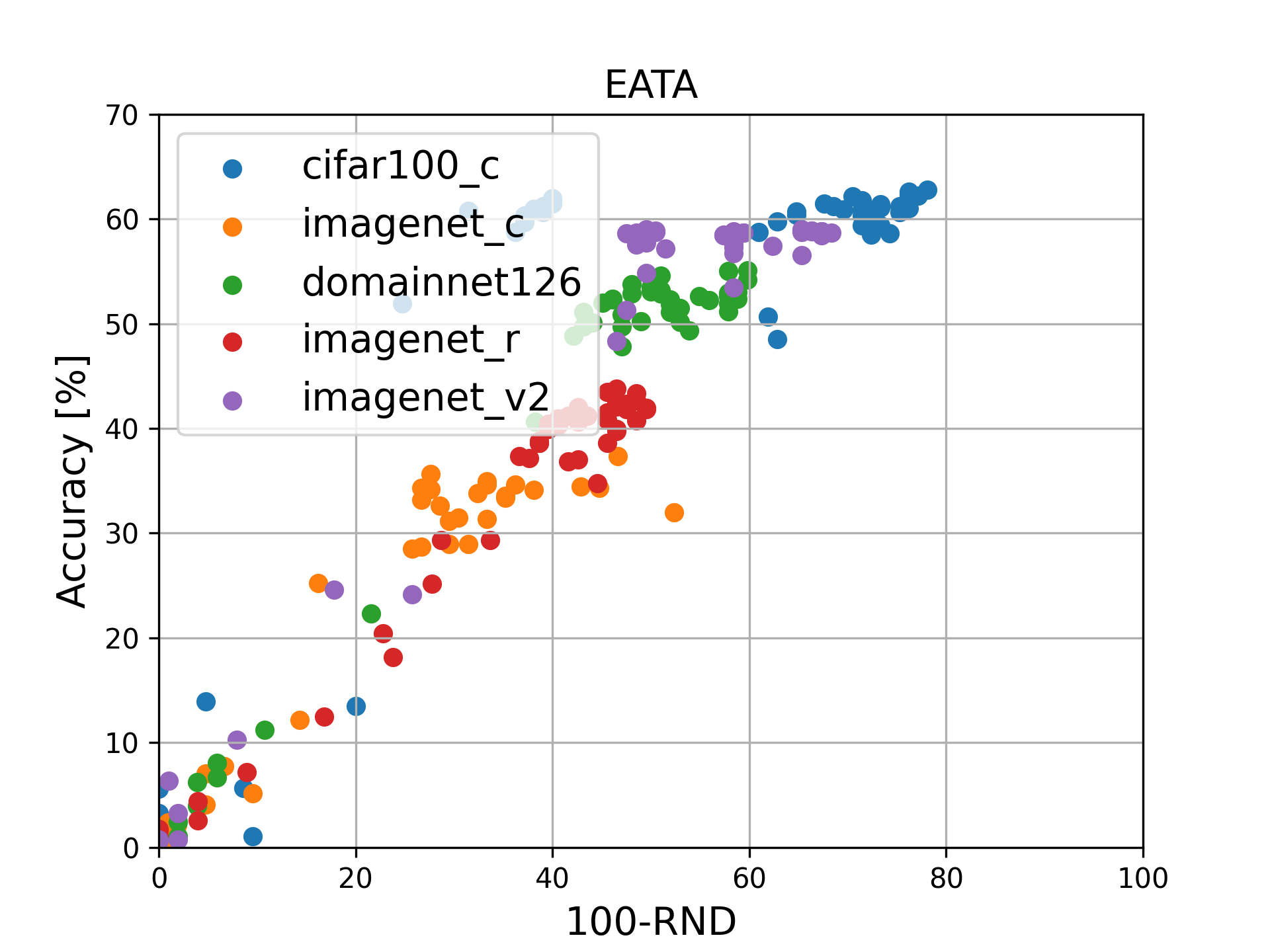}
    \end{subfigure}
    \begin{subfigure}{\corrwidth\textwidth}
        \centering
        \includegraphics[width=\textwidth]{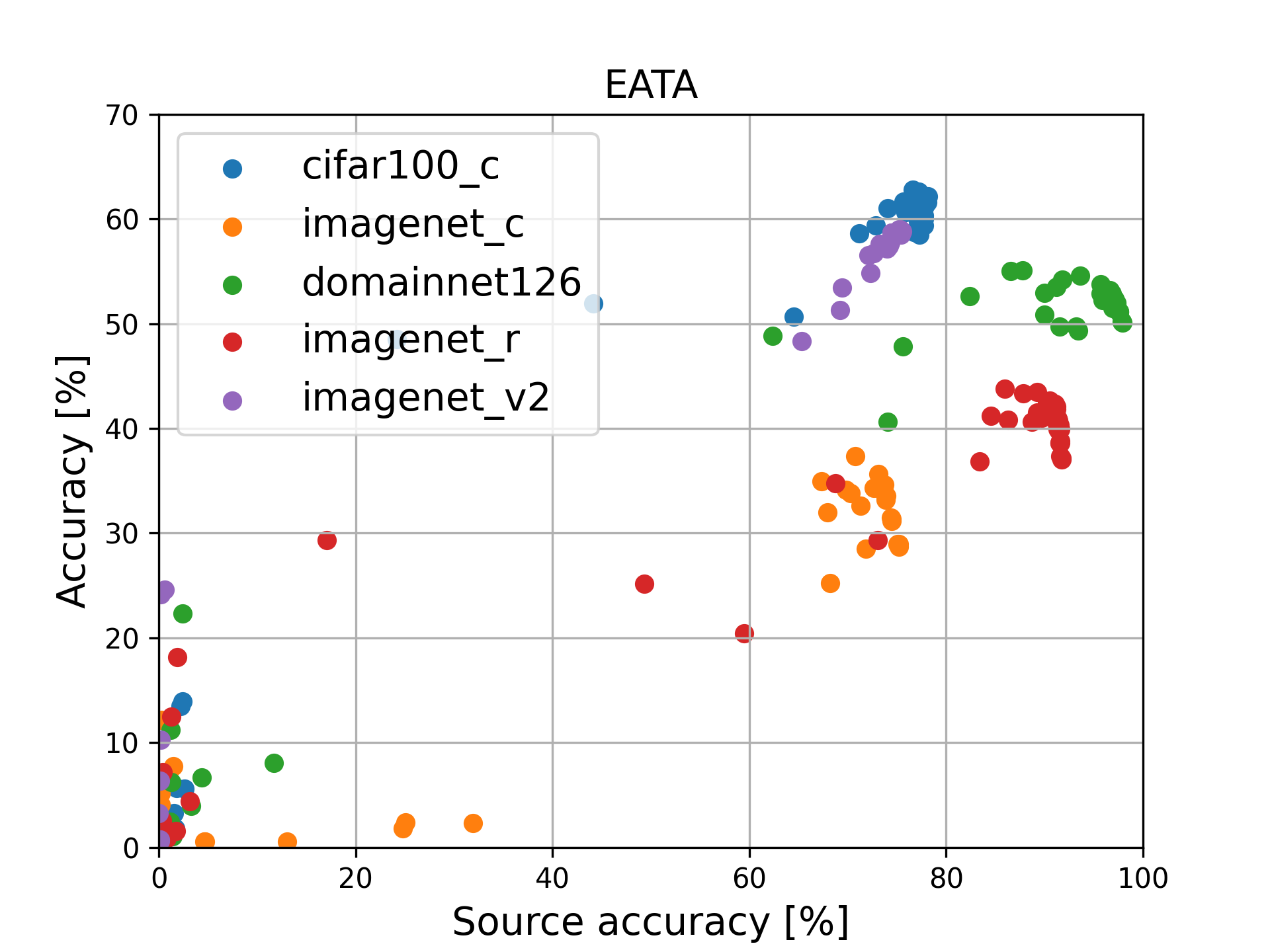}
    \end{subfigure}
    
     \begin{subfigure}{\corrwidth\textwidth}
        \centering
        \includegraphics[width=\textwidth]{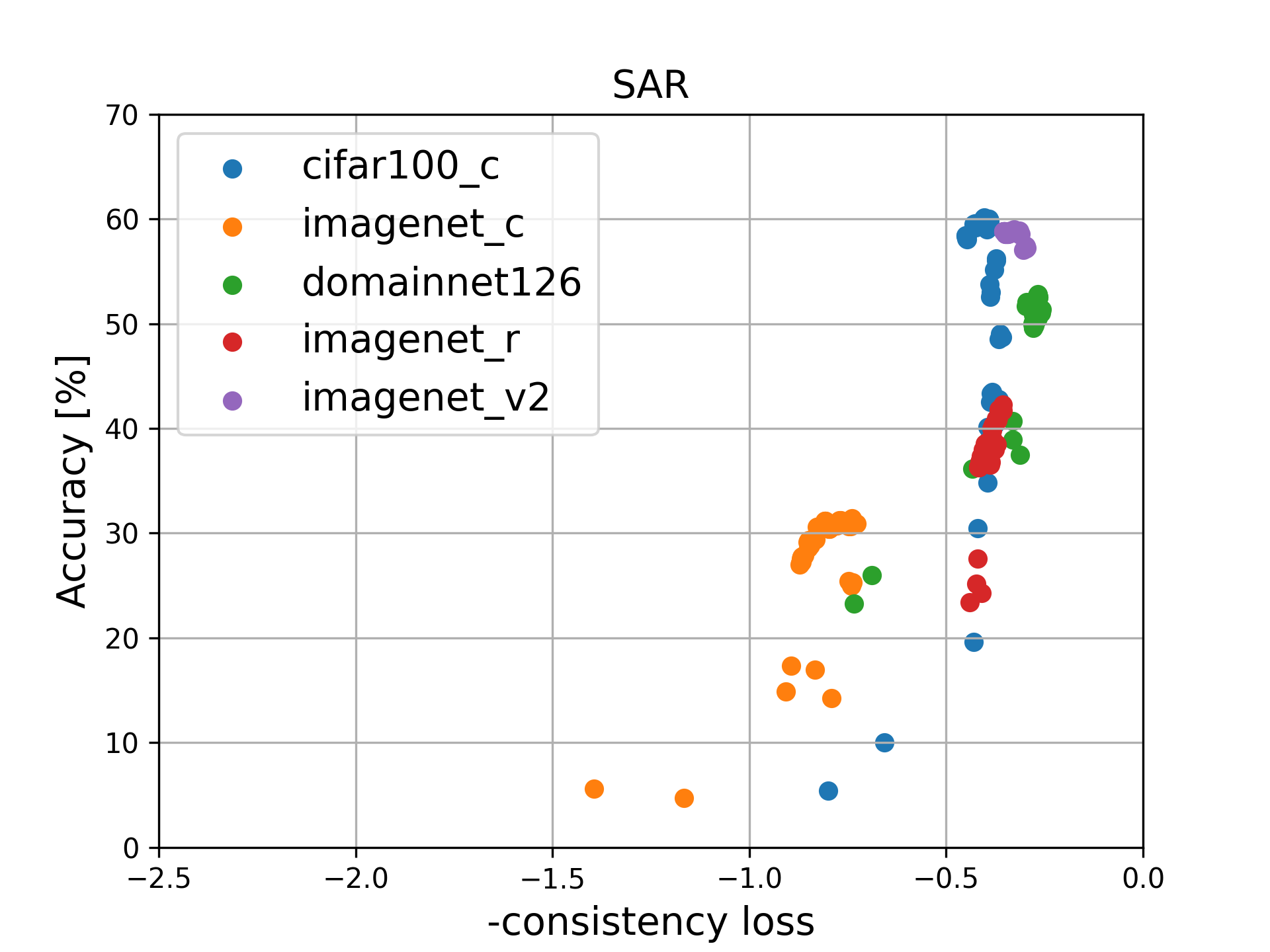}
    \end{subfigure}
     \begin{subfigure}{\corrwidth\textwidth}
        \centering
        \includegraphics[width=\textwidth]{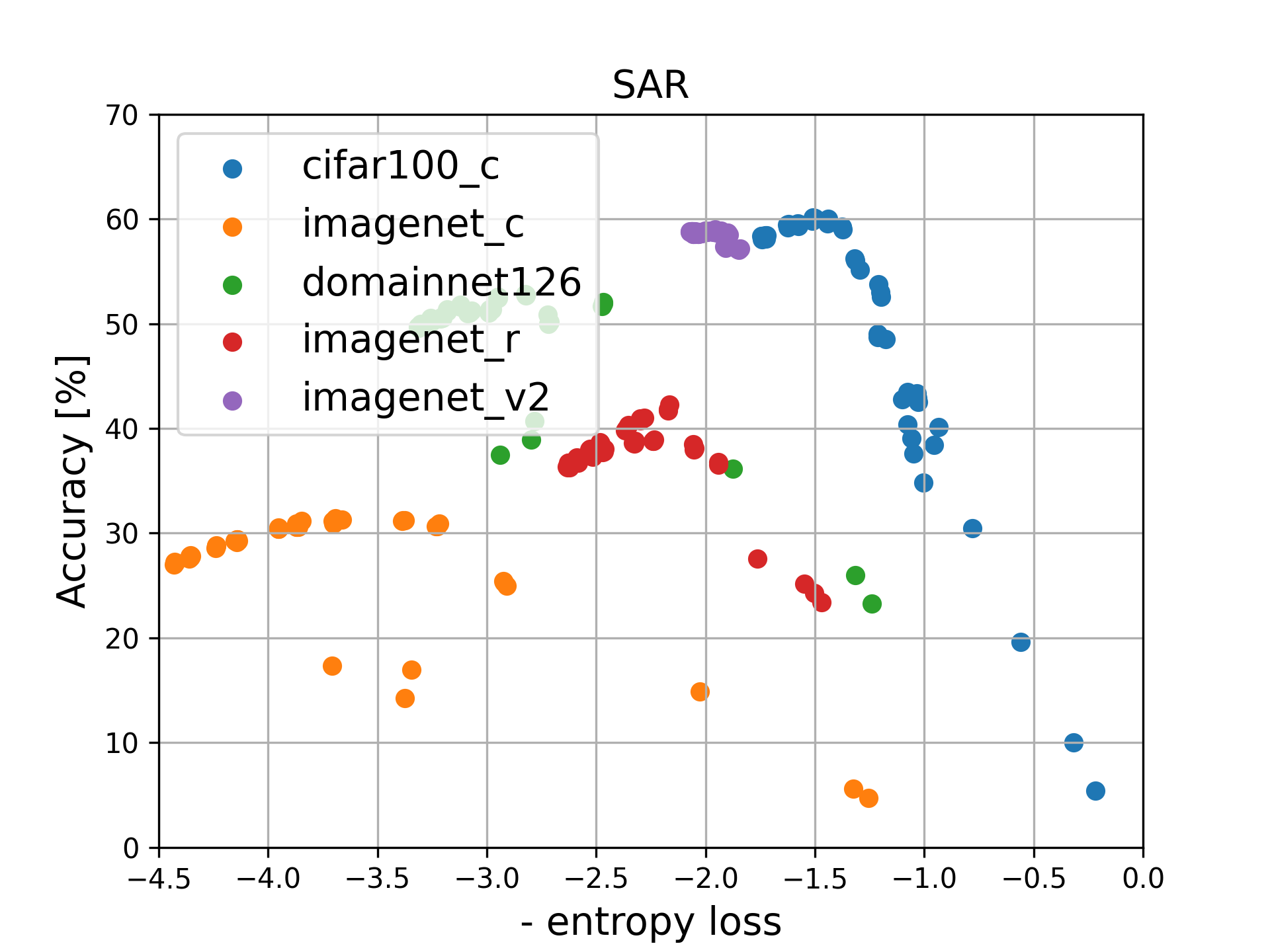}
    \end{subfigure}
     \begin{subfigure}{\corrwidth\textwidth}
        \centering
        \includegraphics[width=\textwidth]{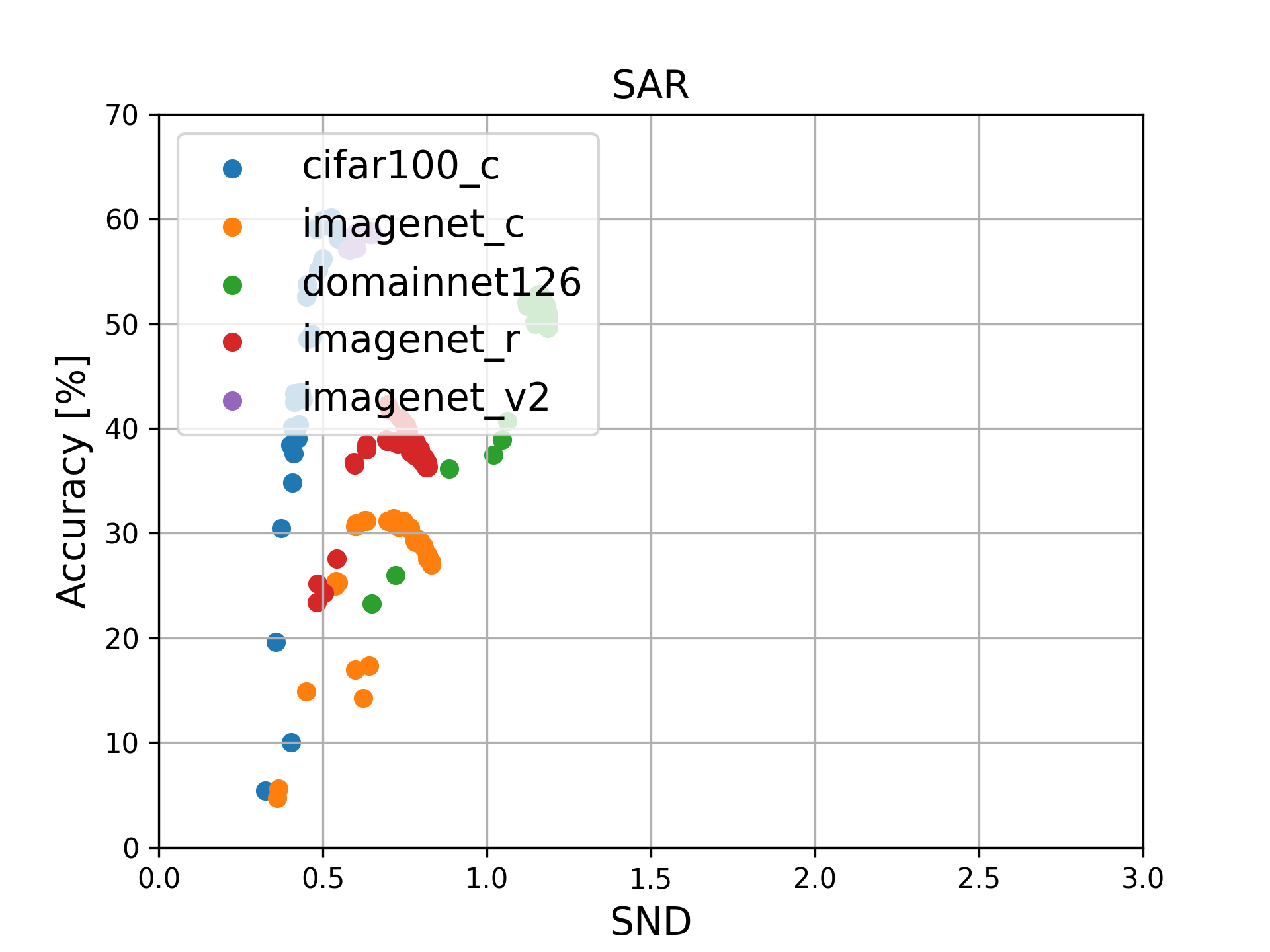}
    \end{subfigure}

    \begin{subfigure}{\corrwidth\textwidth}
        \centering
        \includegraphics[width=\textwidth]{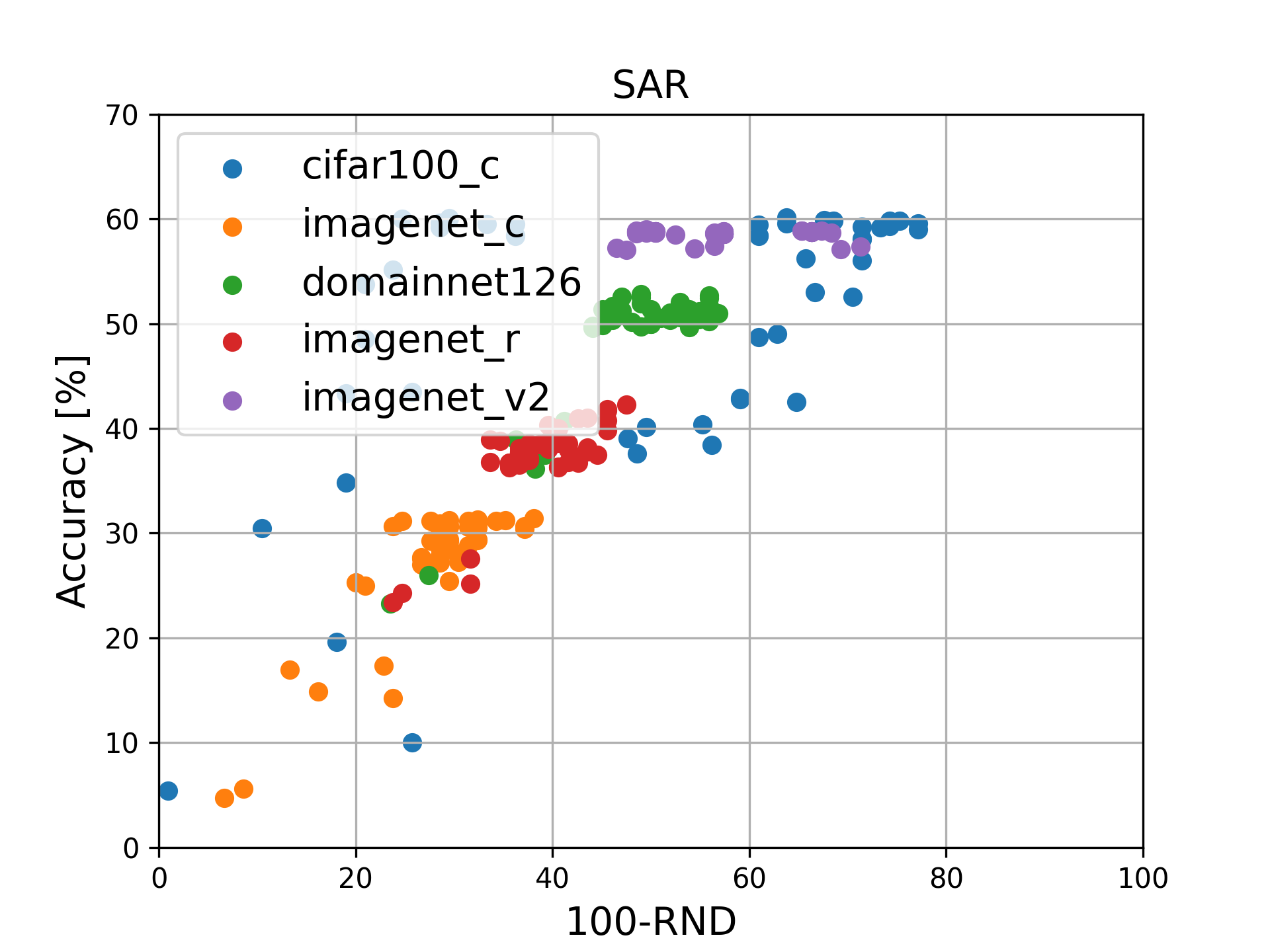}
    \end{subfigure}
     \begin{subfigure}{\corrwidth\textwidth}
        \centering
        \includegraphics[width=\textwidth]{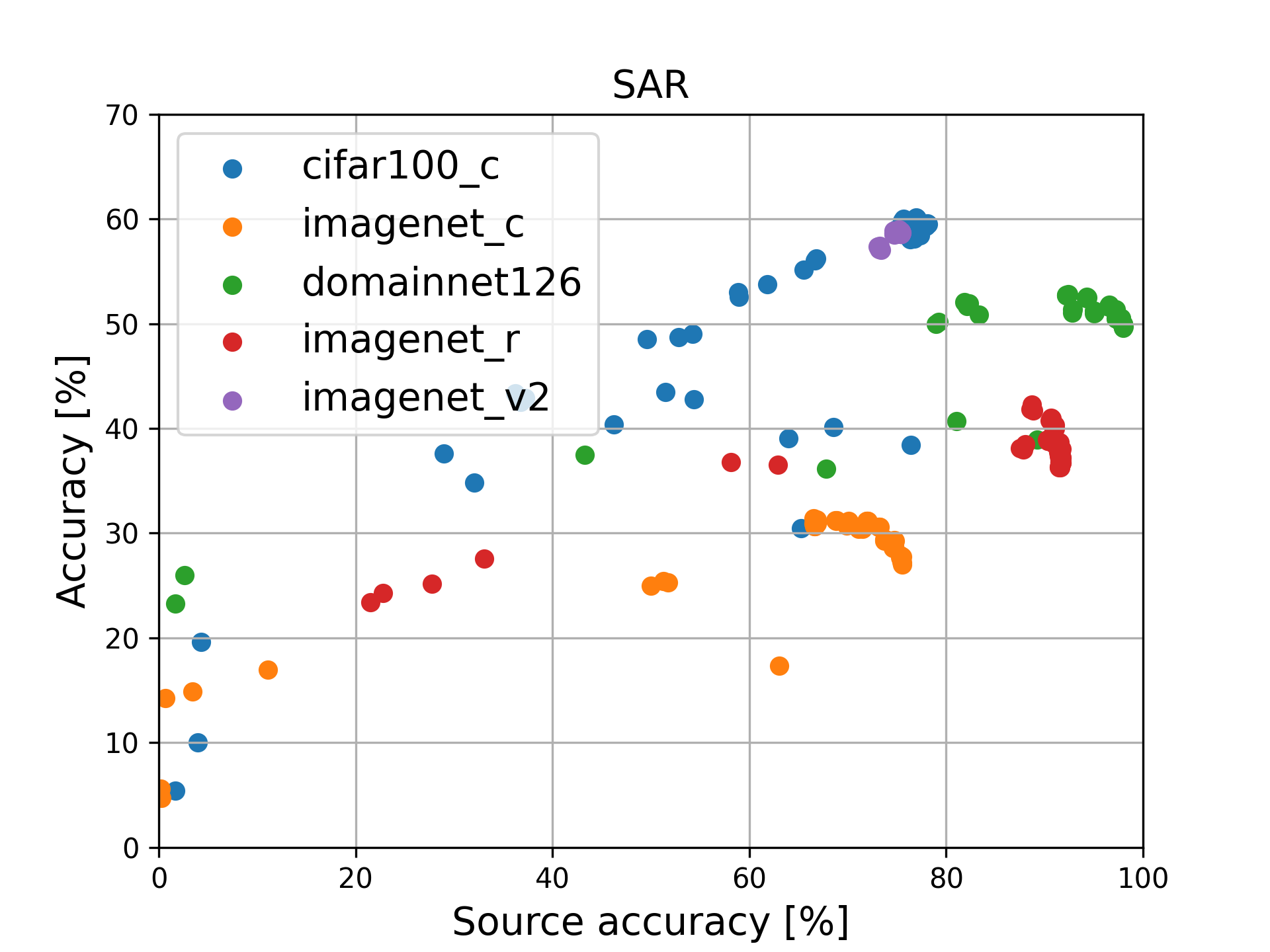}
    \end{subfigure}
         
    \caption{Correlation plots for the different methods between target accuracy (y-axis) and selected surrogate measure (x-axis).}
    \label{fig:corr_plots_appendix}
\end{figure*}

\newcommand{\mywidth}{0.41}

\begin{figure*}[ht]
    \centering
    \begin{subfigure}[t]{\mywidth\textwidth}
        \centering
        \includegraphics[width=\textwidth]{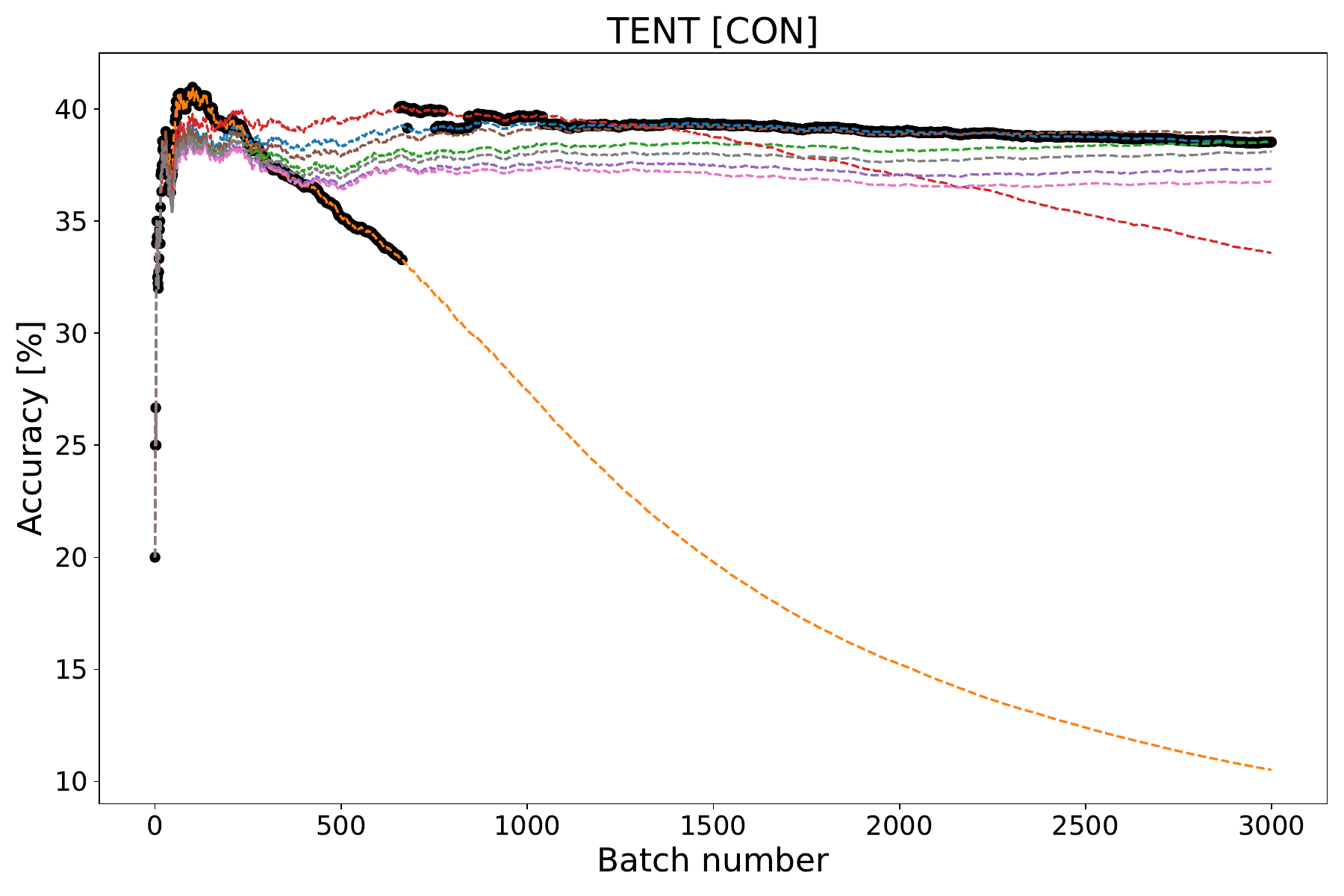}
    \end{subfigure}
    \begin{subfigure}[t]{\mywidth\textwidth}
        \centering
        \includegraphics[width=\textwidth]{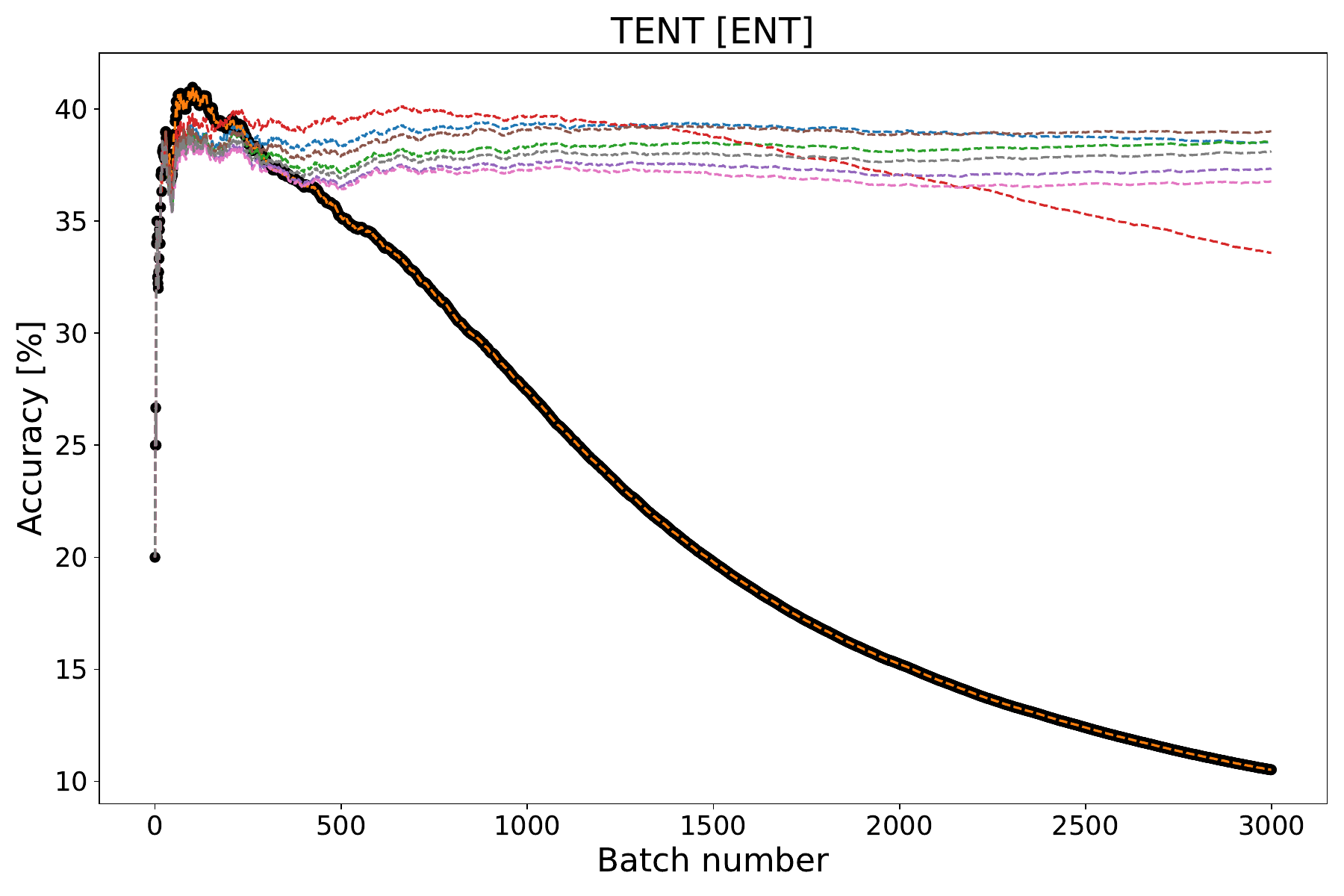}
    \end{subfigure}

    \begin{subfigure}[t]{\mywidth\textwidth}
        \centering
        \includegraphics[width=\textwidth]{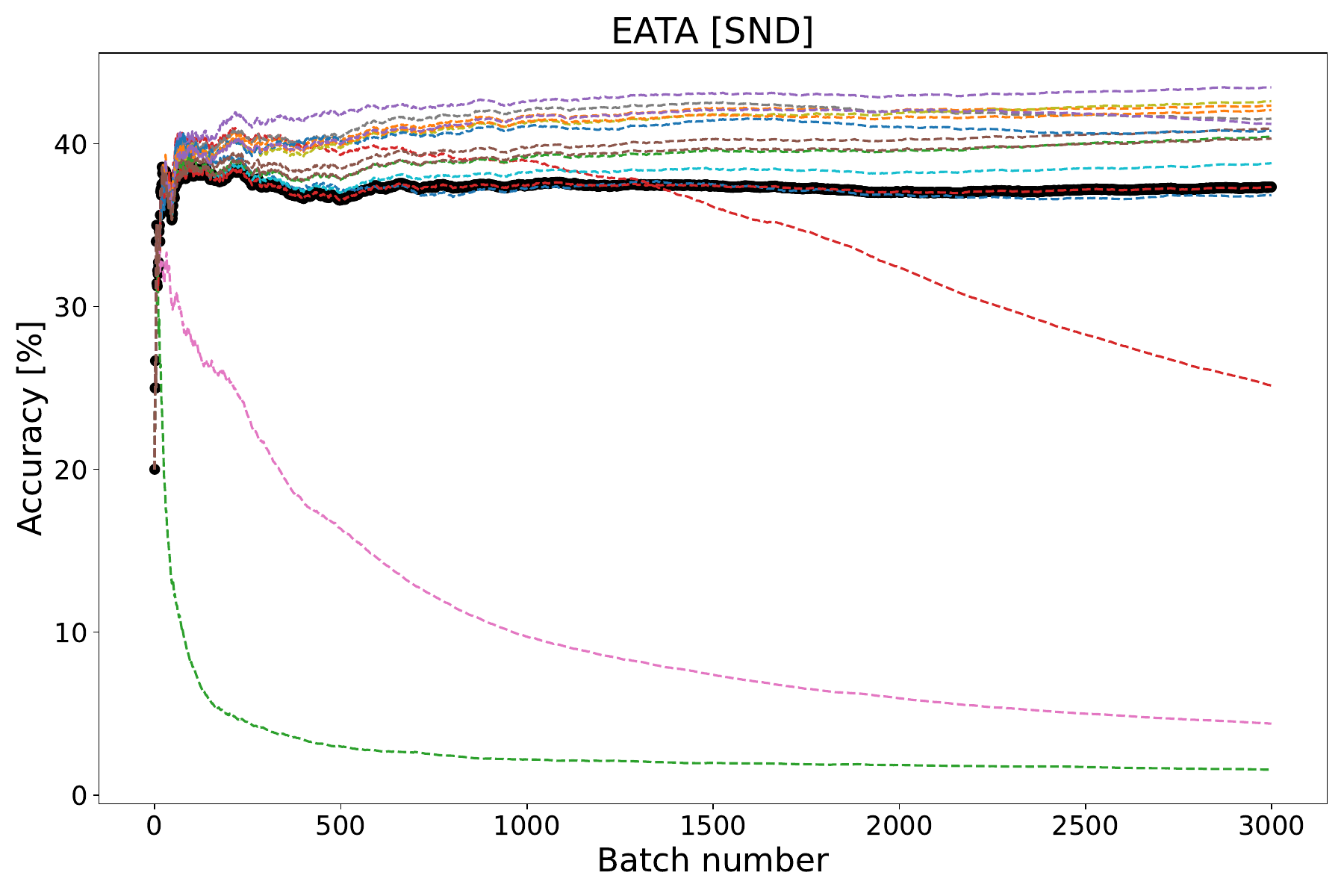}
    \end{subfigure}
    \begin{subfigure}[t]{\mywidth\textwidth}
        \centering
        \includegraphics[width=\textwidth]{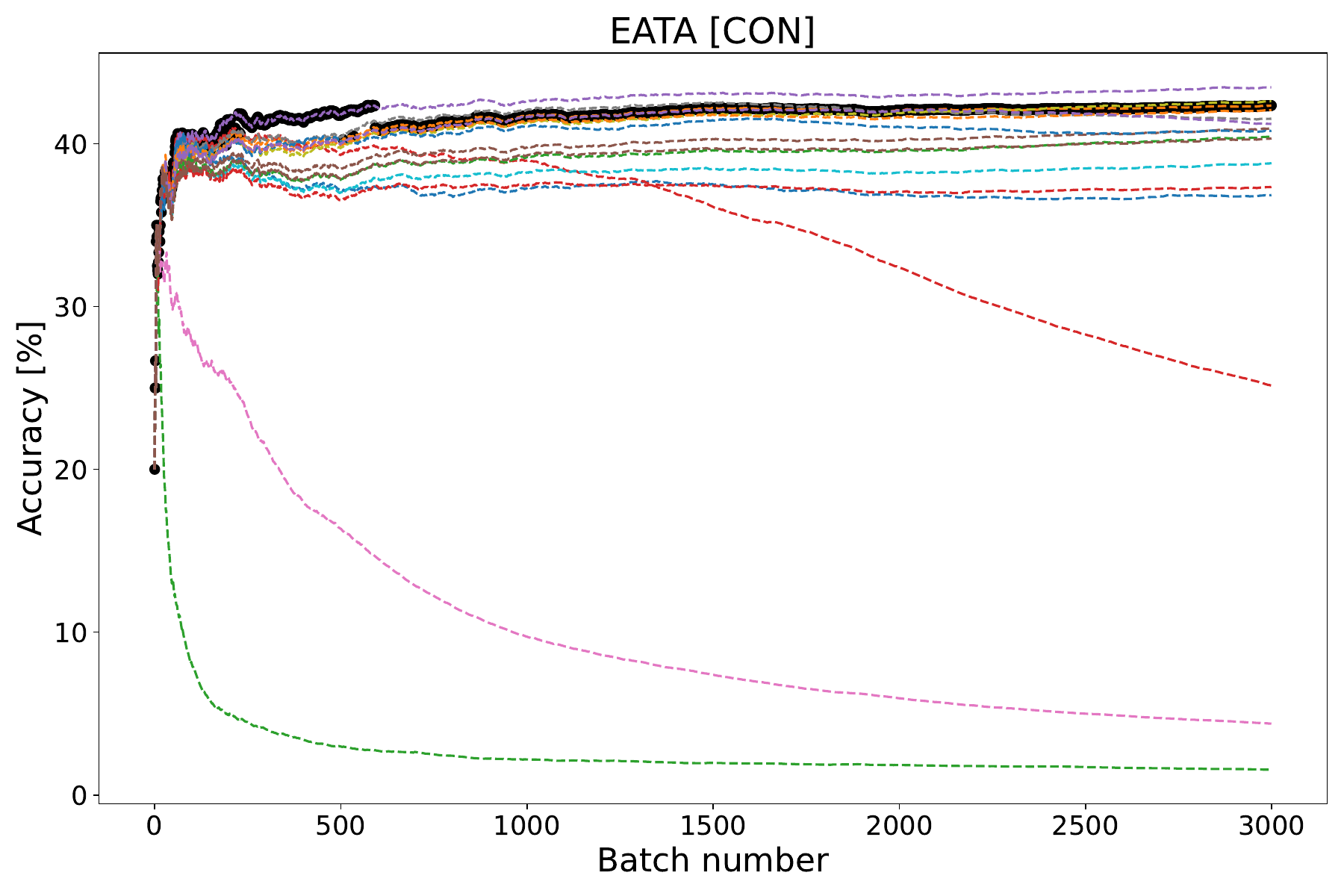}
    \end{subfigure}

    \begin{subfigure}[t]{\mywidth\textwidth}
        \centering
        \includegraphics[width=\textwidth]{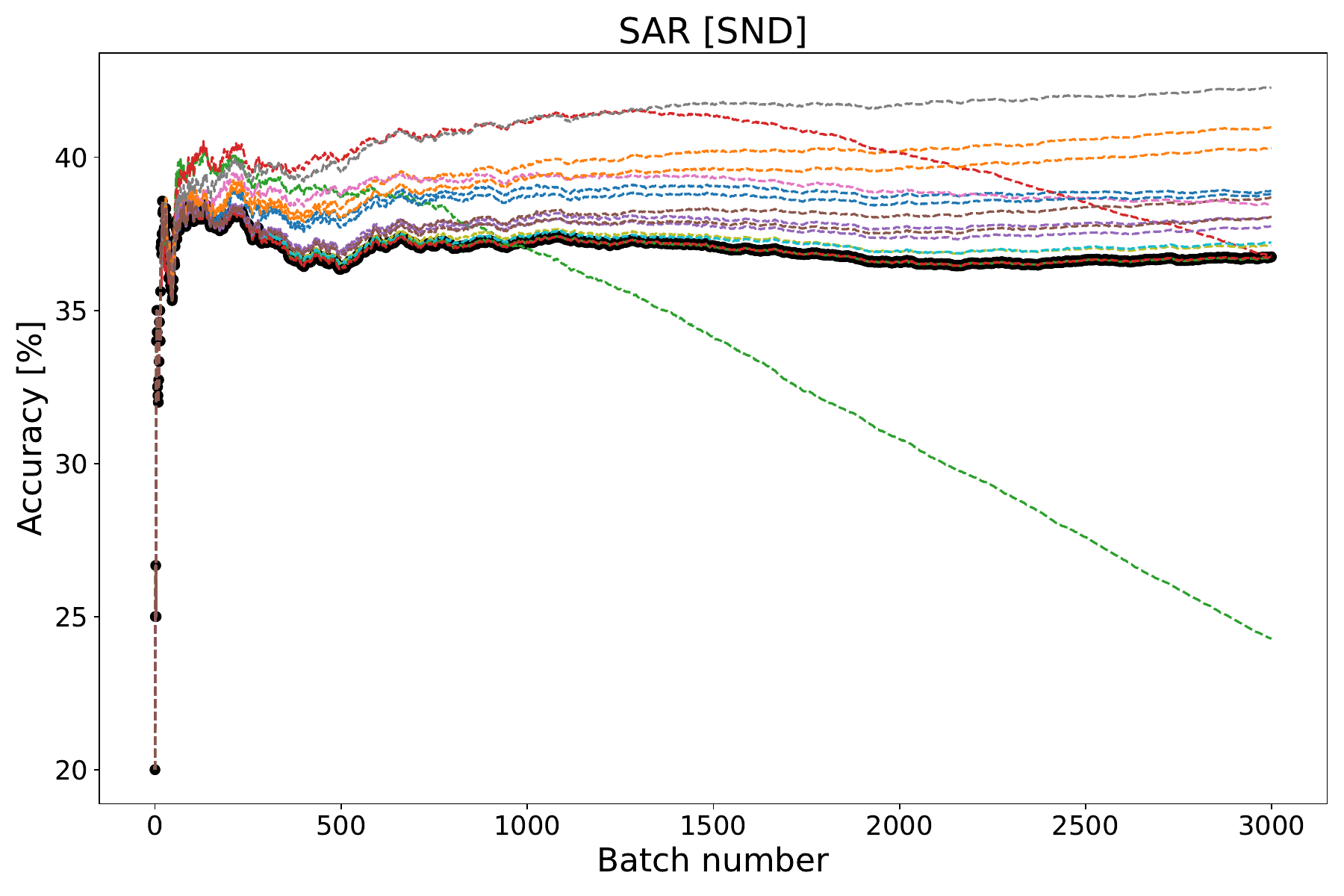}
    \end{subfigure}
    \begin{subfigure}[t]{\mywidth\textwidth}
        \centering
        \includegraphics[width=\textwidth]{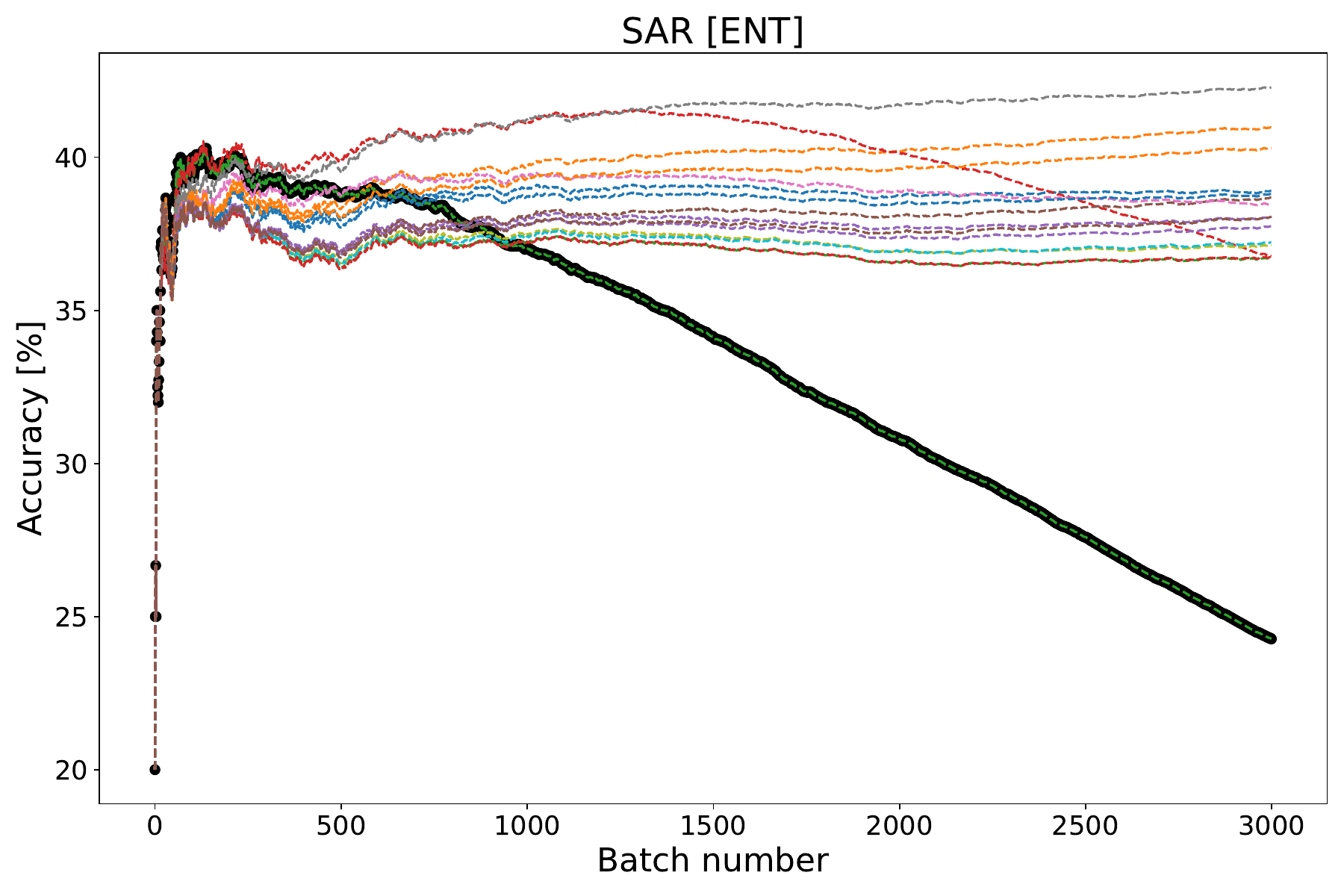}
    \end{subfigure}

    \begin{subfigure}[t]{\mywidth\textwidth}
        \centering
        \includegraphics[width=\textwidth]{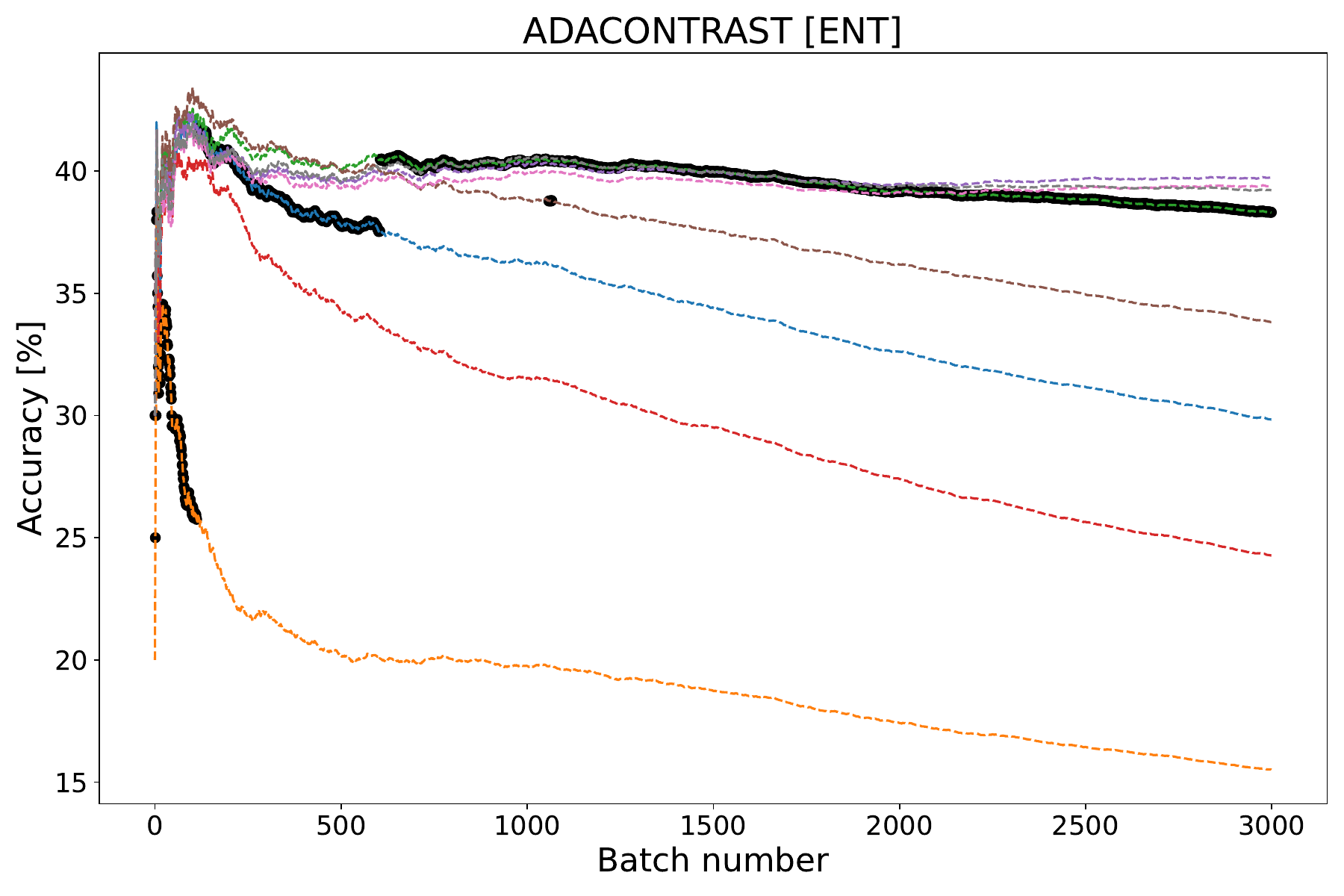}
    \end{subfigure}
    \begin{subfigure}[t]{\mywidth\textwidth}
        \centering
        \includegraphics[width=\textwidth]{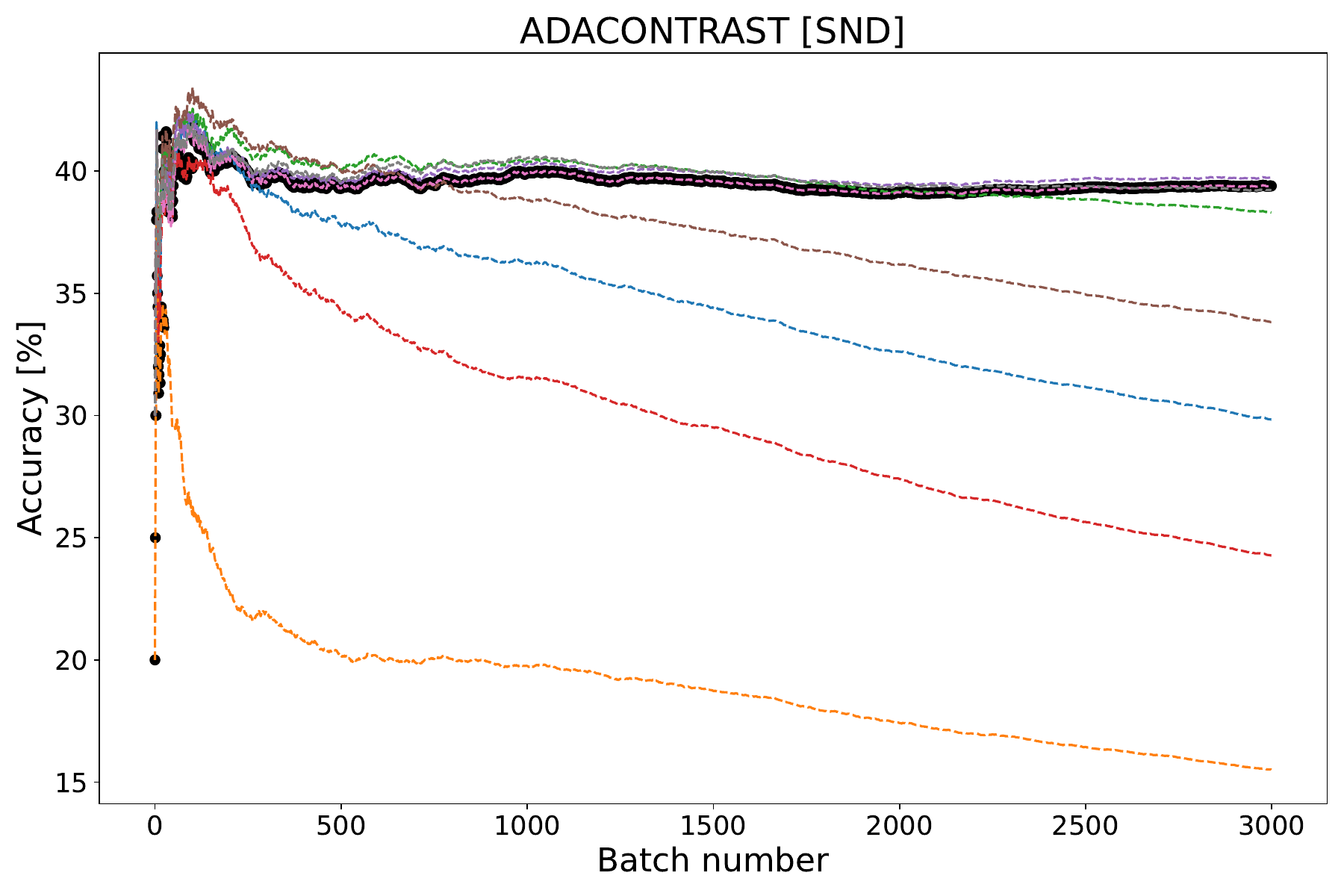}
    \end{subfigure}

    \begin{subfigure}[t]{\mywidth\textwidth}
        \centering
        \includegraphics[width=\textwidth]{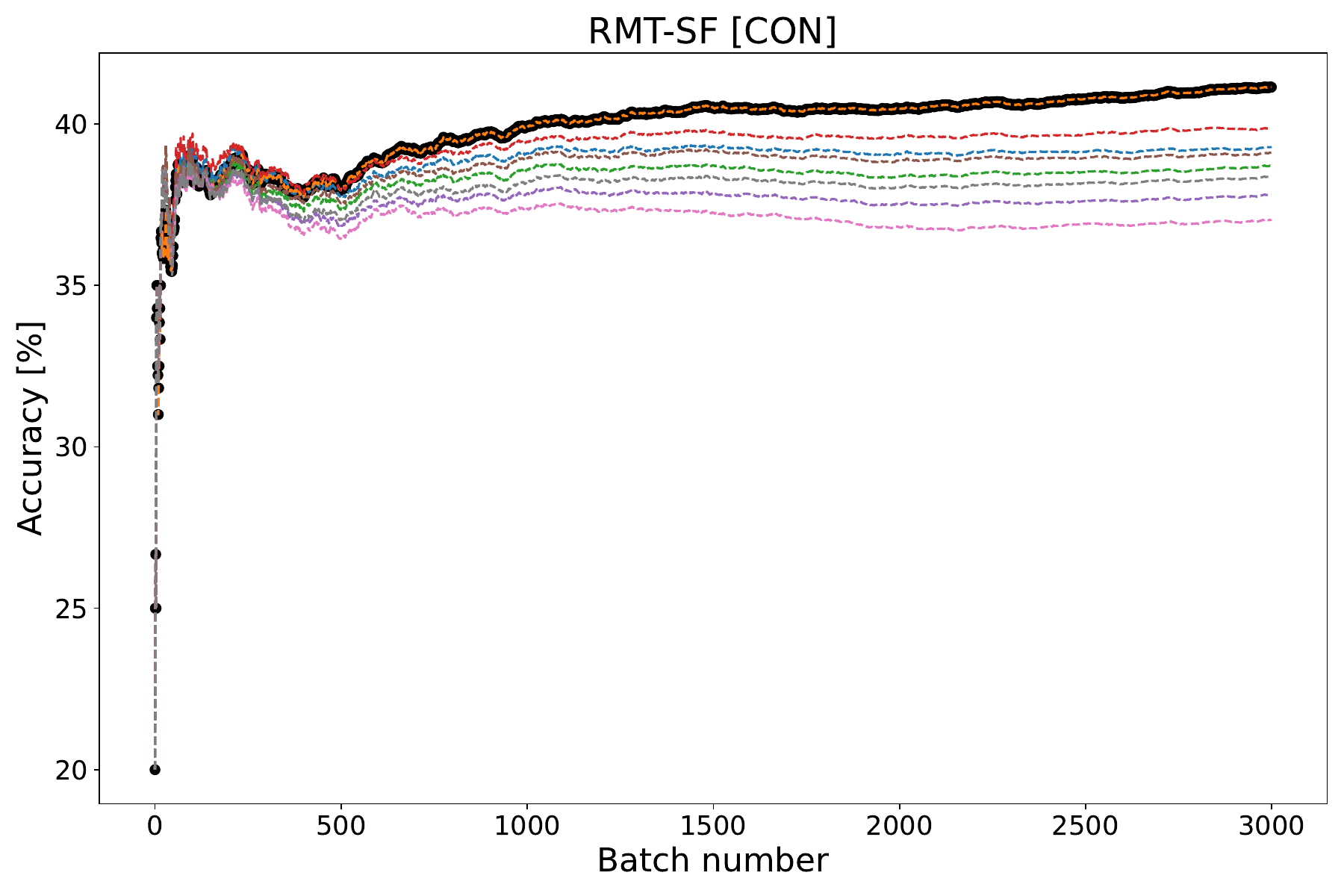}
    \end{subfigure}
    \begin{subfigure}[t]{\mywidth\textwidth}
        \centering
        \includegraphics[width=\textwidth]{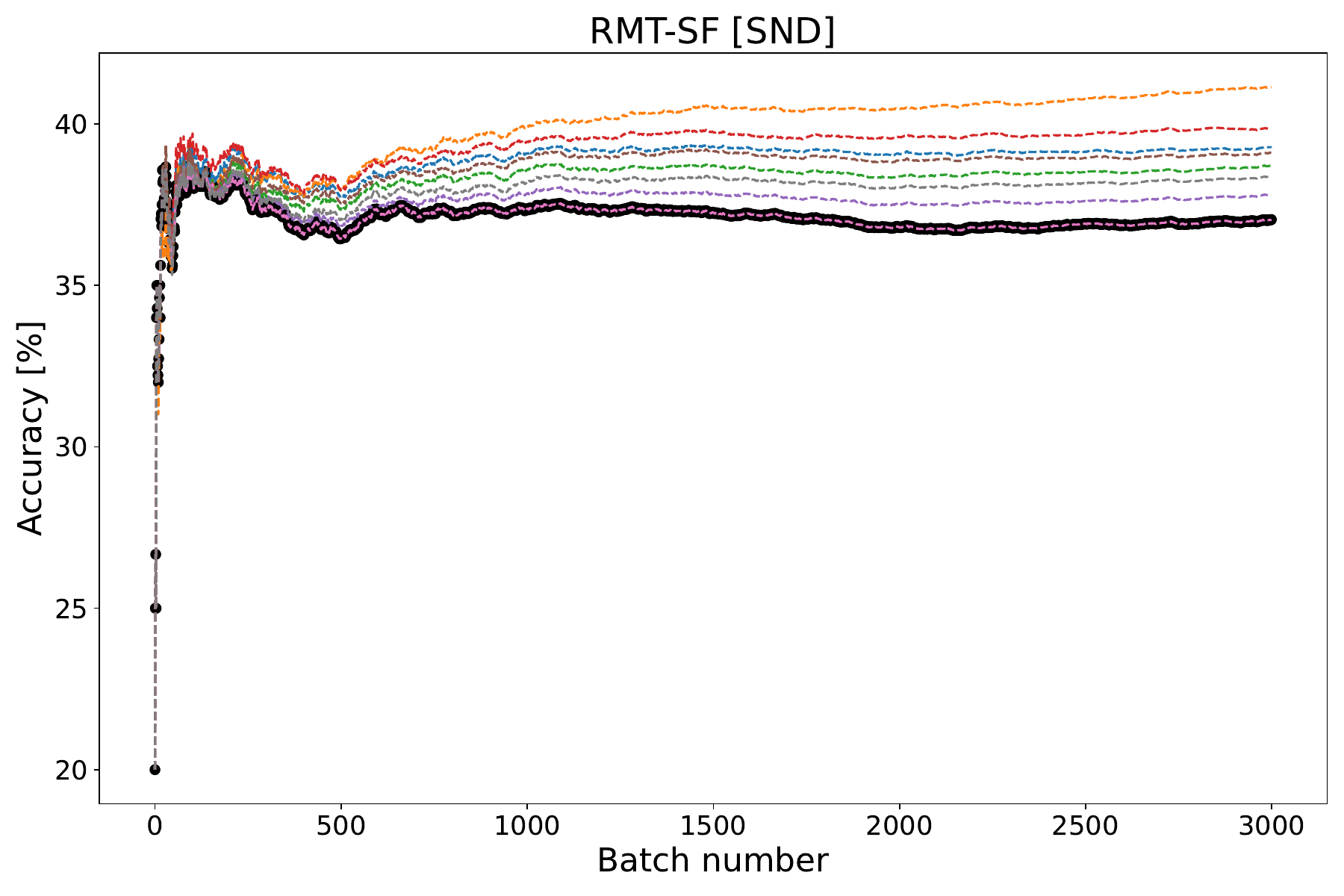}
    \end{subfigure}

    \caption{Accumulated accuracy on ImageNetR dataset, different lines relate to runs with different hyper-parameters. A black dot ($\bullet$) marks the optimal model according to the selected unsupervised strategy at each timestep. }
    \label{fig:online_imagenetr}
\end{figure*}

\begin{figure*}[ht]
    \centering
    \begin{subfigure}[t]{\mywidth\textwidth}
        \centering
        \includegraphics[width=\textwidth]{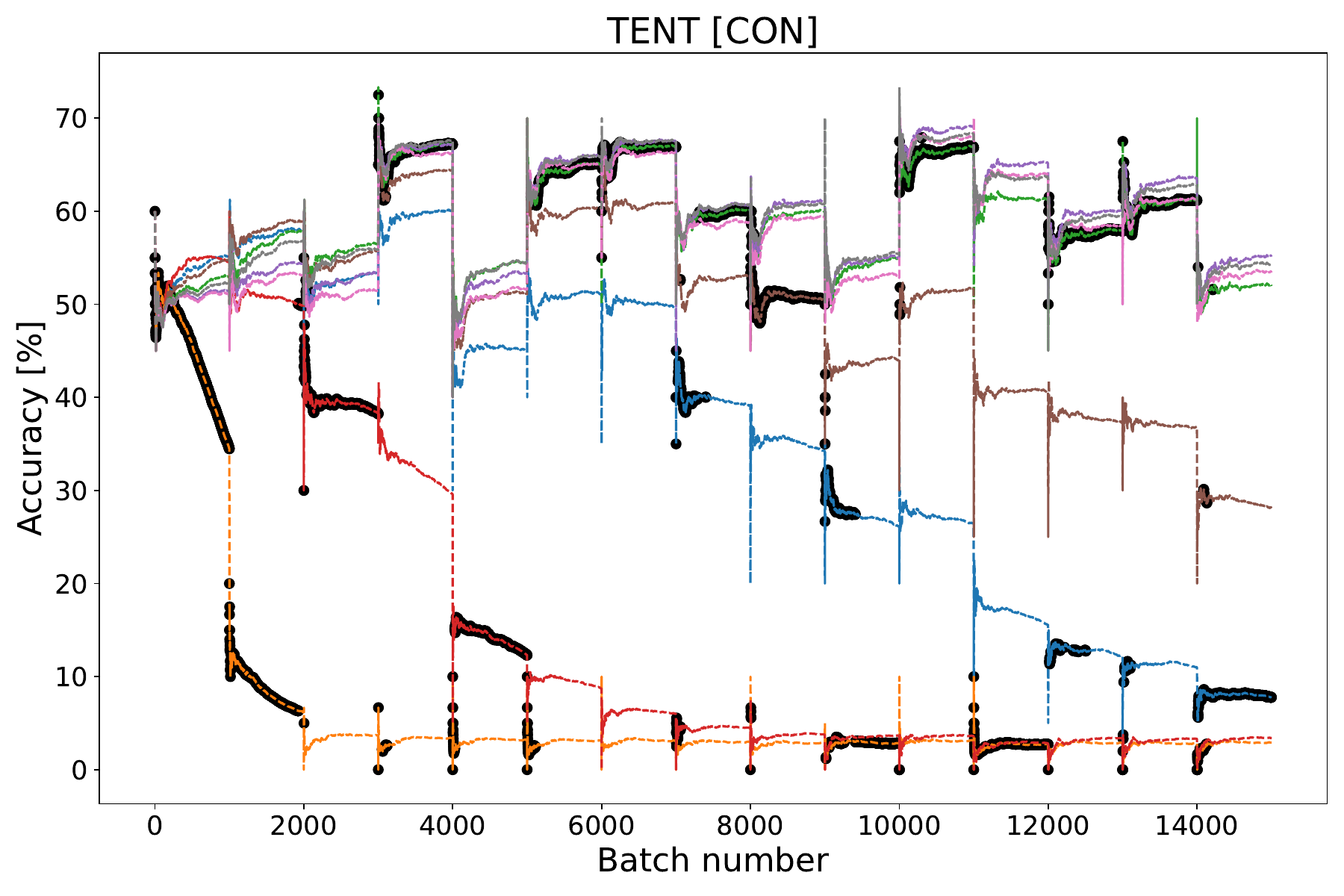}
    \end{subfigure}
    \begin{subfigure}[t]{\mywidth\textwidth}
        \centering
        \includegraphics[width=\textwidth]{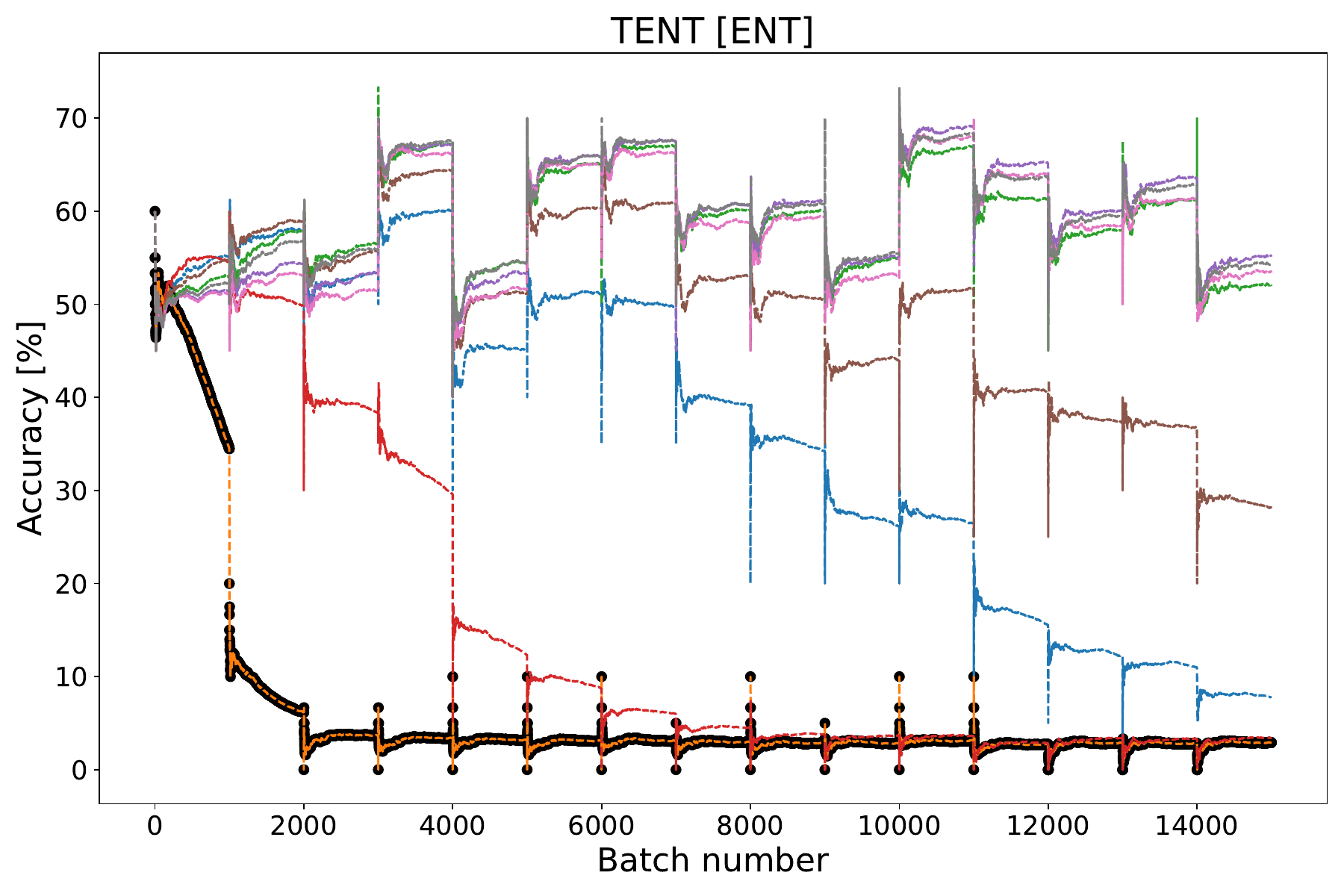}
    \end{subfigure}

    \begin{subfigure}[t]{\mywidth\textwidth}
        \centering
        \includegraphics[width=\textwidth]{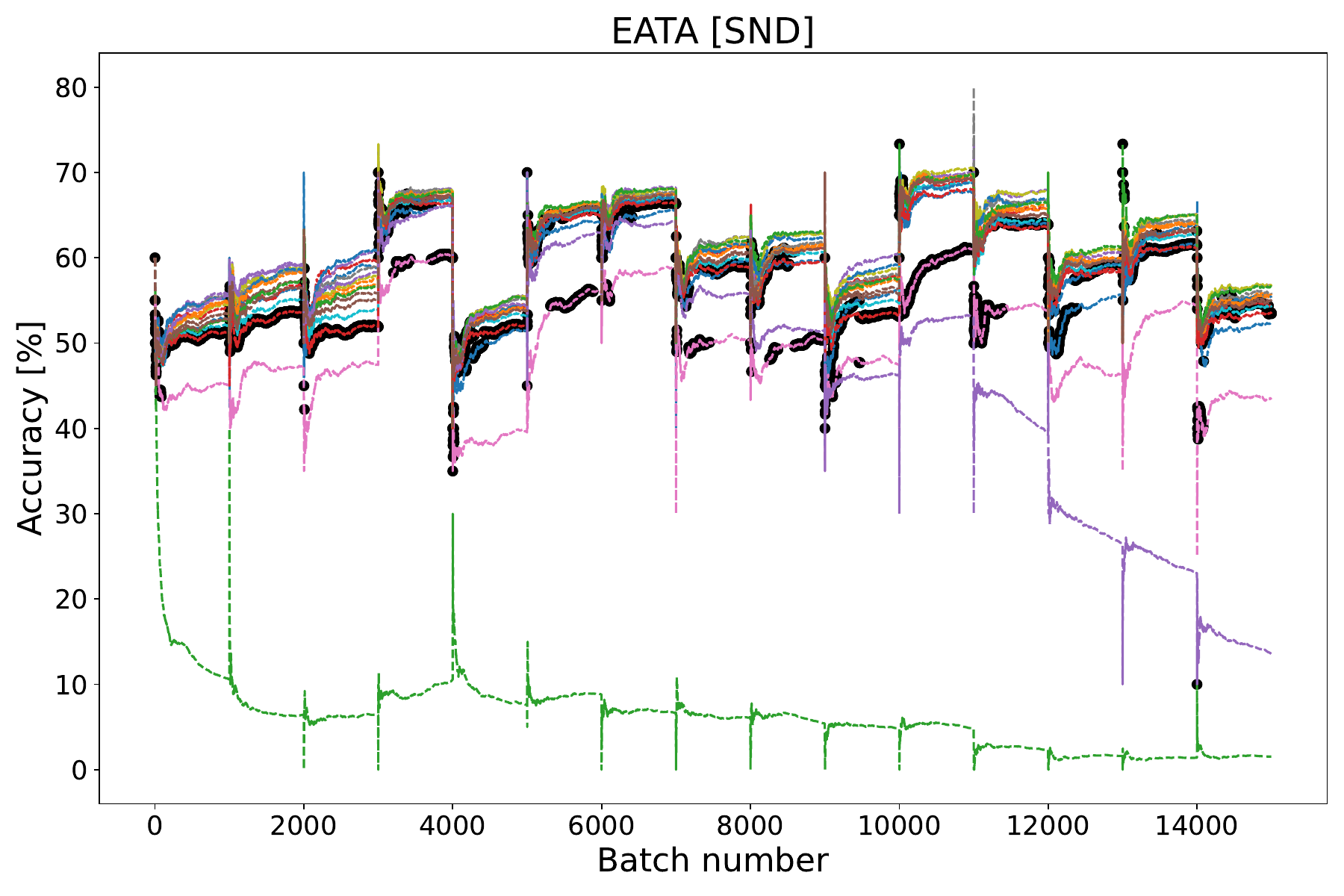}
    \end{subfigure}
    \begin{subfigure}[t]{\mywidth\textwidth}
        \centering
        \includegraphics[width=\textwidth]{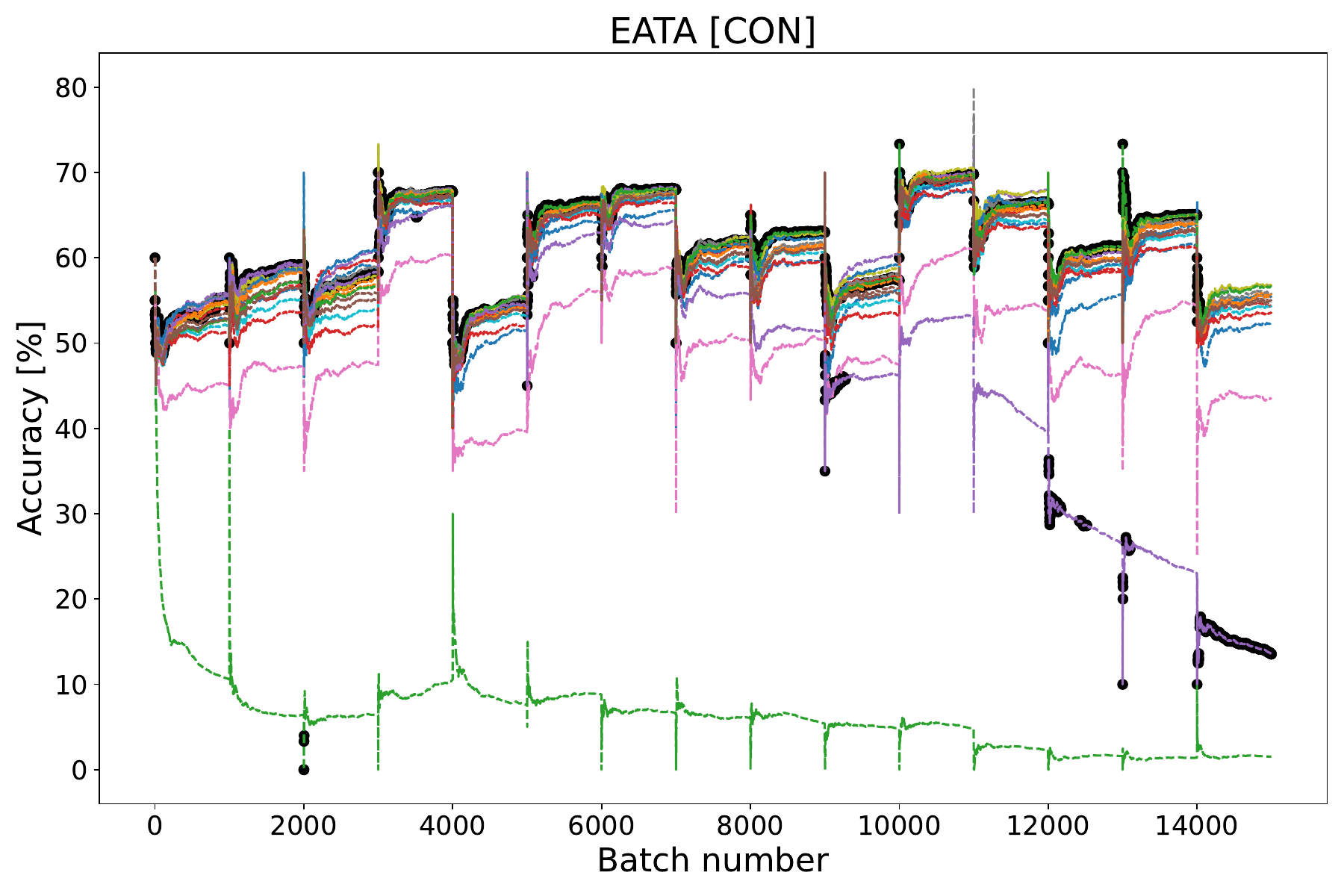}
    \end{subfigure}

    \begin{subfigure}[t]{\mywidth\textwidth}
        \centering
        \includegraphics[width=\textwidth]{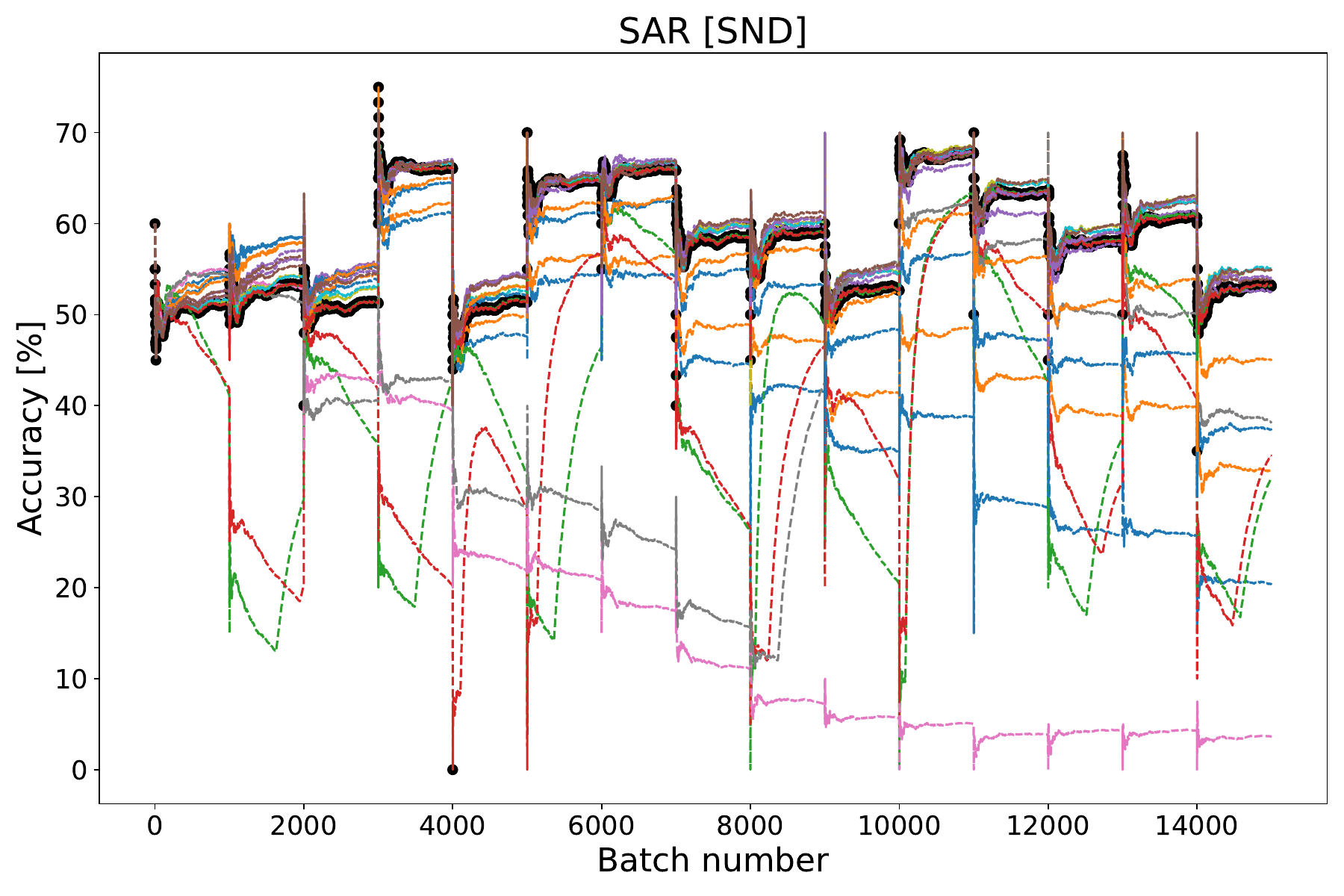}
    \end{subfigure}
    \begin{subfigure}[t]{\mywidth\textwidth}
        \centering
        \includegraphics[width=\textwidth]{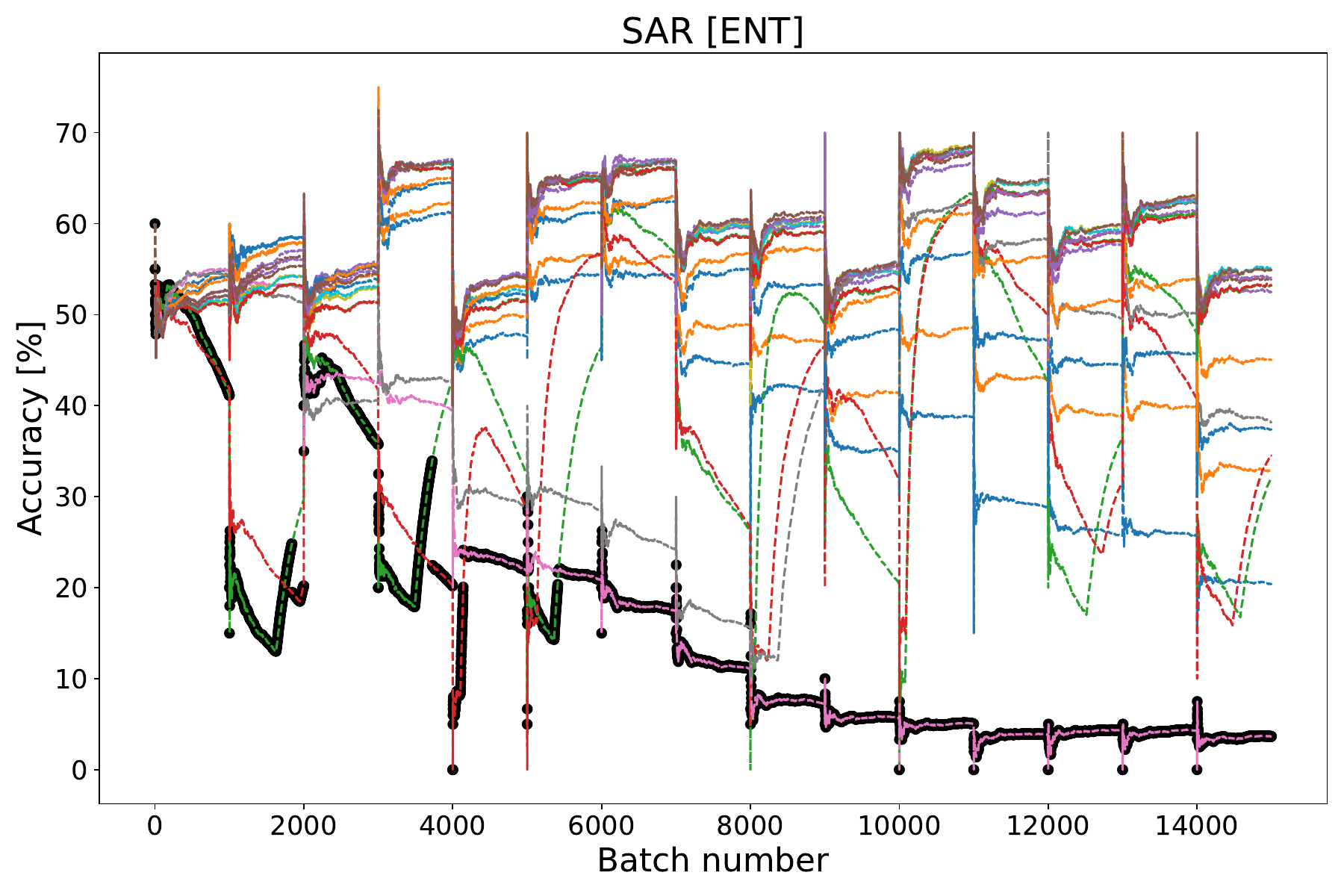}
    \end{subfigure}

    \begin{subfigure}[t]{\mywidth\textwidth}
        \centering
        \includegraphics[width=\textwidth]{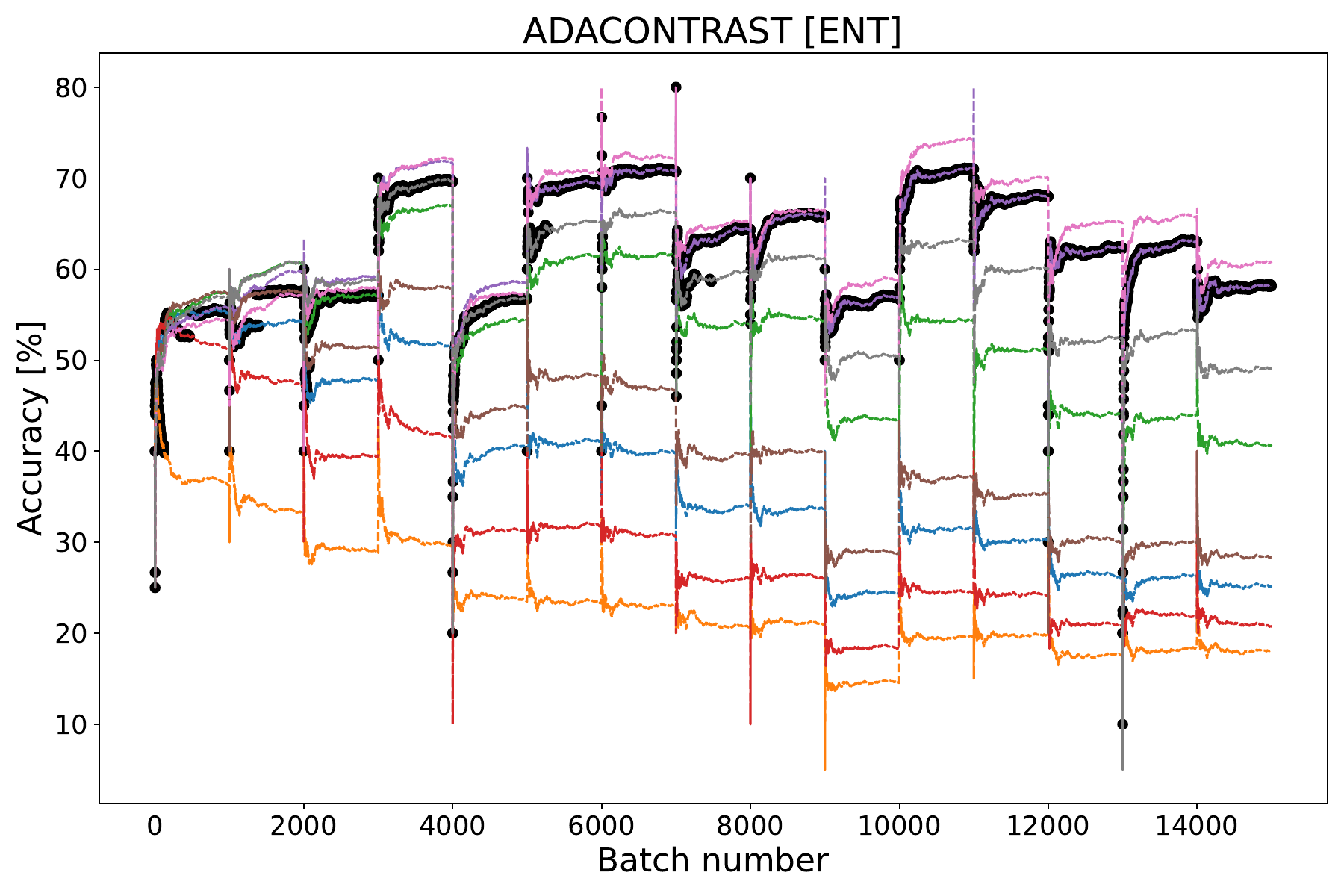}
    \end{subfigure}
    \begin{subfigure}[t]{\mywidth\textwidth}
        \centering
        \includegraphics[width=\textwidth]{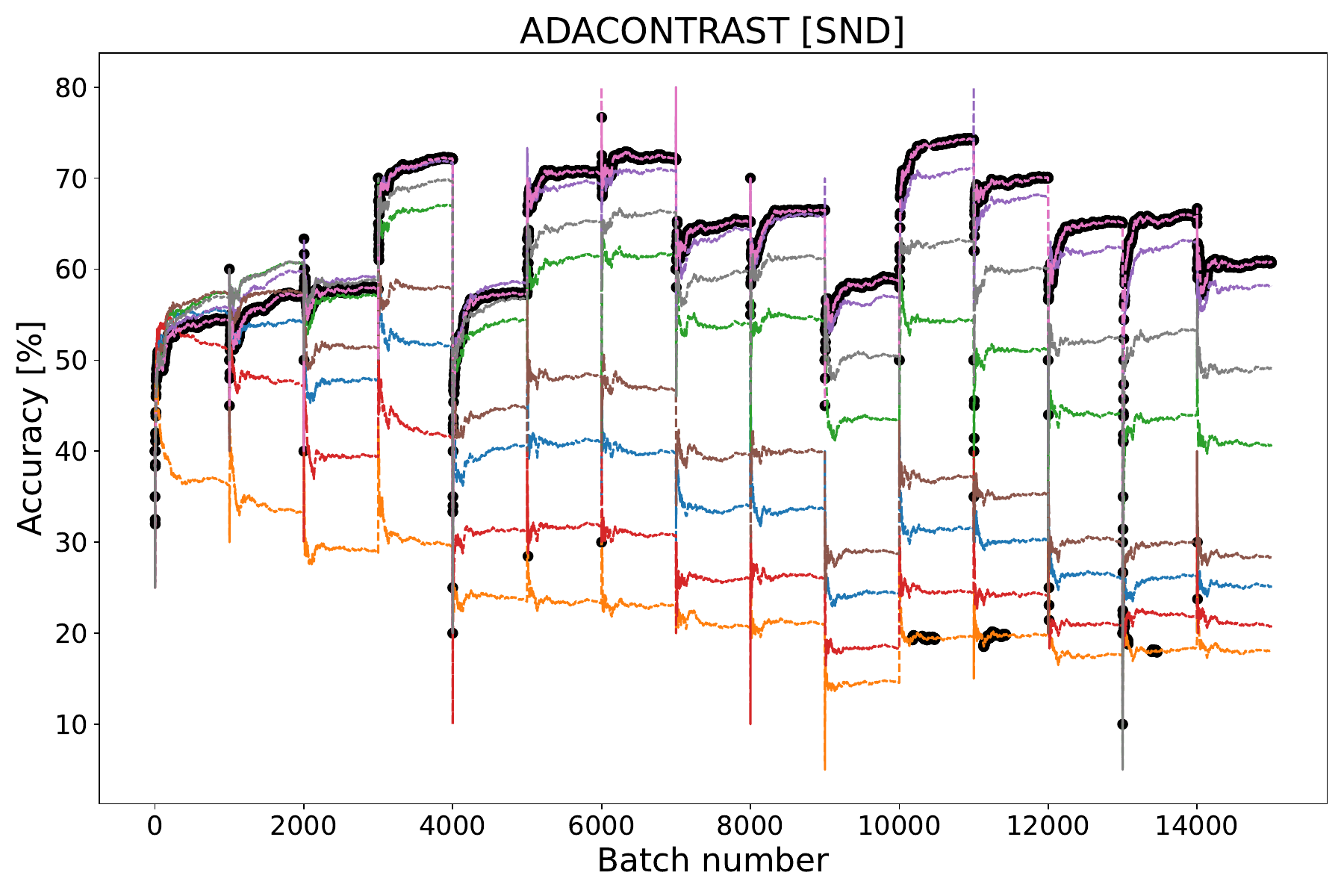}
    \end{subfigure}

    \begin{subfigure}[t]{\mywidth\textwidth}
        \centering
        \includegraphics[width=\textwidth]{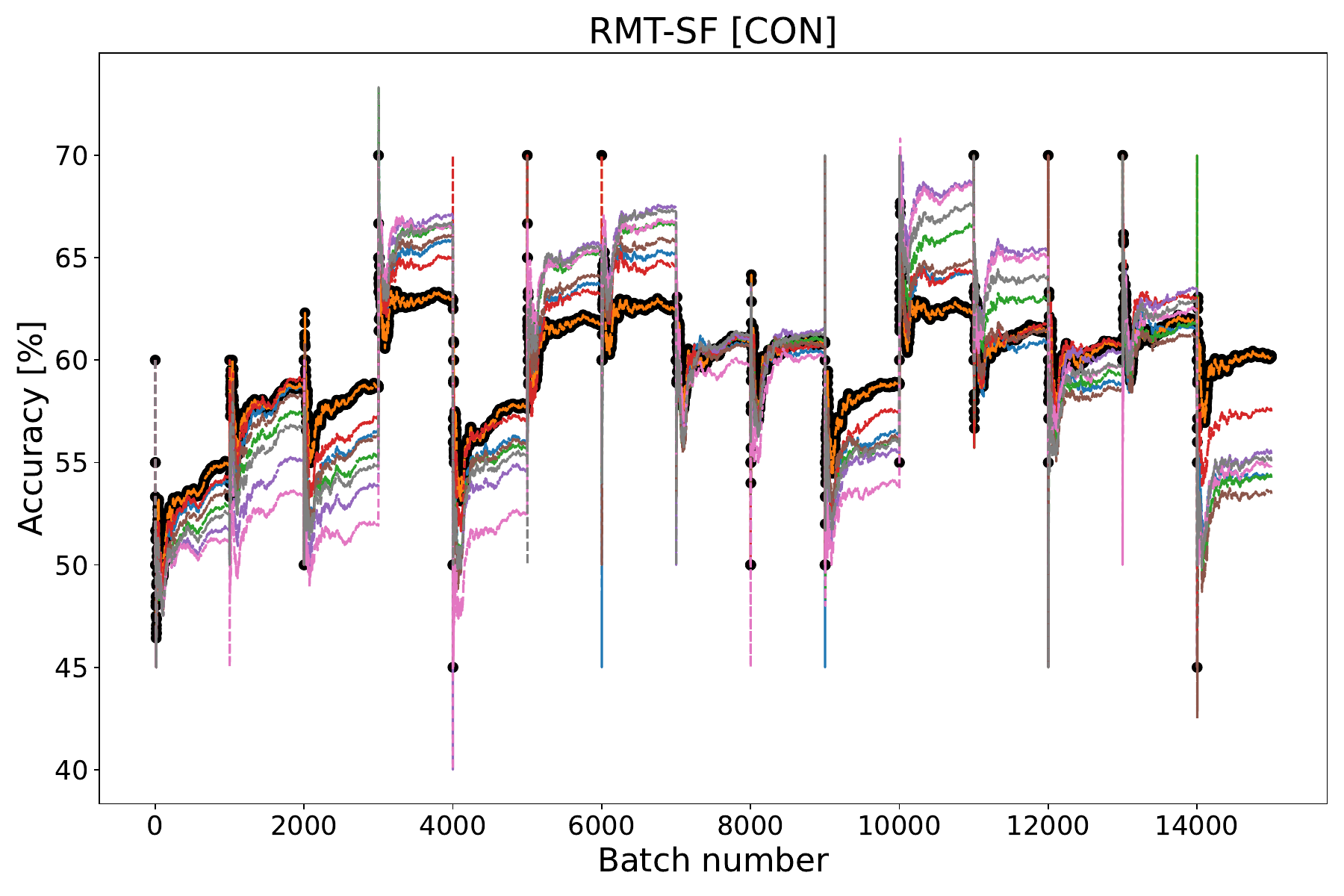}
    \end{subfigure}
    \begin{subfigure}[t]{\mywidth\textwidth}
        \centering
        \includegraphics[width=\textwidth]{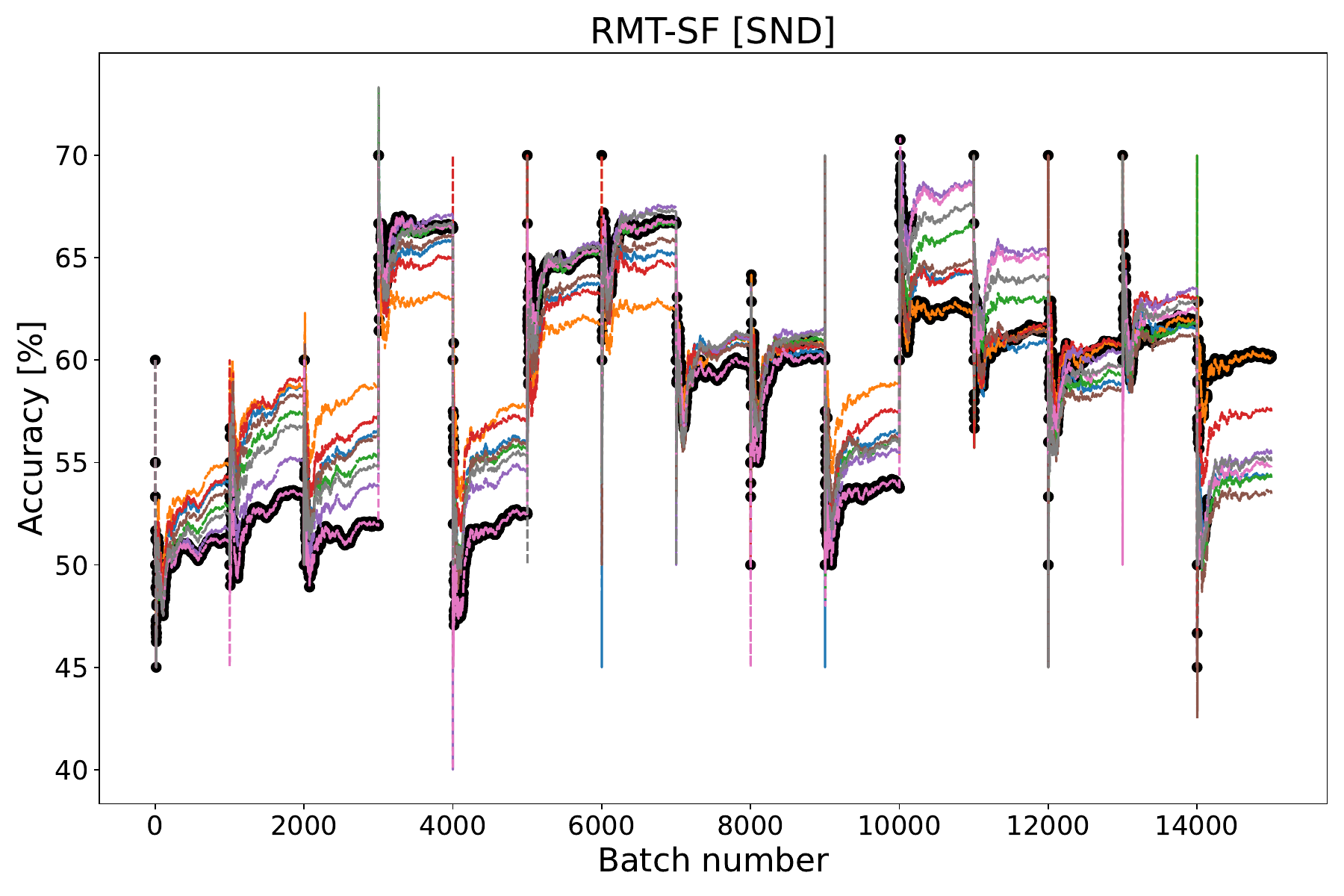}
    \end{subfigure}

    \caption{Accumulated accuracy on CIFAR-100-C dataset, different lines relate to runs with different hyper-parameters. A black dot ($\bullet$) marks the optimal model according to the selected unsupervised strategy at each timestep. Significant changes in accuracy at different time-steps mark the change in distortion type. }
    \label{fig:online_cifar}
\vspace{-0.5cm}
\end{figure*}

\begin{table*}[ht] 

\resizebox{\textwidth}{!}{

\begin{tabular}{l@{\hskip 3pt}l@{\hskip 3pt}c@{\hskip 3pt}c@{\hskip 3pt}c@{\hskip 3pt}c@{\hskip 3pt}c@{\hskip 3pt}c@{\hskip 3pt}c@{\hskip 3pt}c}
\textbf{Dataset} & \textbf{Method} & \textbf{\textsc{s-acc}} & \textbf{\textsc{c-acc}} & \textbf{\textsc{ent}} 
& \textbf{\textsc{con}} 
& \textbf{\textsc{snd}} & \textbf{\textsc{100-rnd}} & \textbf{\textsc{median}} & \textbf{\textsc{oracle}}\\
\midrule
\multirow{7}{*}{\textsc{CIFAR100-C}} & \textsc{source} & \multicolumn{8}{c}{53.55}\\
 & \textsc{tent} 
 &  \textcolor{dontkillmyeyesgreen}{59.93$\pm$0.15} 
 & 59.56$\pm$0.2 
 & \textcolor{debianred}{5.32$\pm$0.05} 
 & 59.56$\pm$0.20 
 & 40.78$\pm$25.04
 & 59.17$\pm$0.41 
 & 54.46$\pm$0.51 
 & 60.16$\pm$0.14\\ 

 & \textsc{eata}  
 & \textcolor{dontkillmyeyesgreen}{61.78$\pm$0.40} 
 & \textcolor{dontkillmyeyesgreen}{61.86$\pm$0.47} 
 & \textcolor{debianred}{21.01$\pm$20.12} 
 & 58.60$\pm$4.71 
 & 39.40$\pm$27.15
 & 61.95$\pm$0.57 
 & 60.51$\pm$0.36 
 & 62.44$\pm$0.35\\ 

 & \textsc{sar} 
 & \textcolor{dontkillmyeyesgreen}{59.47$\pm$0.11} 
 & 45.65$\pm$5.83 
 & \textcolor{debianred}{11.68$\pm$5.92} 
 & 48.75$\pm$0.21 
 & 58.26$\pm$0.13
 & 59.26$\pm$0.2 
 & 57.03$\pm$0.27 
 & 60.01$\pm$0.12\\ 
 
   & \textsc{adacontrast} 
   & \textcolor{dontkillmyeyesgreen}{\textbf{64.43$\pm$0.01}} 
 & 63.70$\pm$0.11 
 & 63.69$\pm$0.11 
 & \textcolor{debianred}{37.97$\pm$0.43}
 &  \textcolor{dontkillmyeyesgreen}{\textbf{64.43$\pm$0.01}}
 & 64.43$\pm$0.01 
 & 48.48$\pm$0.56 
 & 64.43$\pm$0.01\\ 

 & \textsc{memo} 
   & 60.66$\pm$0.02 
 & \textcolor{dontkillmyeyesgreen}{61.08$\pm$0.14} 
 & \textcolor{dontkillmyeyesgreen}{61.03$\pm$0.17} 
 & \textcolor{debianred}{59.32$\pm$0.01}
& 60.66$\pm$0.02
 & 60.87$\pm$0.20 
 & 60.82$\pm$0.08 
 & 61.28$\pm$0.02\\ 
 
 & \textsc{lame} & \multicolumn{8}{c}{53.01$\pm$0.03} \\ 
  
  & \textsc{rmt-sf} & \textcolor{debianred}{59.22$\pm$0.13} 
 & 60.04$\pm$0.1 
 & \textcolor{dontkillmyeyesgreen}{60.17$\pm$0.07} 
 & \textcolor{dontkillmyeyesgreen}{60.17$\pm$0.07} 
 & \textcolor{debianred}{59.22}$\pm$0.13
 & 59.93$\pm$0.13 
 & 60.06$\pm$0.12 
 & 60.33$\pm$0.11\\ 
 
\midrule
\multirow{8}{*}{\textsc{ImageNet-C}} & \textsc{source} & \multicolumn{7}{c}{17.97}\\
  & \textsc{tent} 
  & 27.98$\pm$0.07 
 & \textcolor{dontkillmyeyesgreen}{31.08$\pm$0.06} 
 & \textcolor{debianred}{1.52$\pm$0.1} 
 & 29.79$\pm$0.16 
  & 27.97$\pm$0.07
 & 30.29$\pm$0.76 
 & 28.83$\pm$0.03 
 & 31.37$\pm$0.03\\ 

 & \textsc{eata} 
 & \textcolor{dontkillmyeyesgreen}{28.87$\pm$0.11} 
 & 14.60$\pm$12.38 
 & \textcolor{debianred}{1.96$\pm$2.28} 
 & 23.39$\pm$16.54 
  & 19.34$\pm$13.57
 & 32.25$\pm$0.99 
 & 18.74$\pm$8.45 
 & 35.98$\pm$1.02\\ 

  & \textsc{sar} 
  & 27.14$\pm$0.09 
 & 28.72$\pm$0.11 
 & \textcolor{debianred}{8.41$\pm$4.6} 
 & \textcolor{dontkillmyeyesgreen}{31.19$\pm$0.22} 
  & 27.14$\pm$0.09
 & 30.93$\pm$0.35 
 & 29.30$\pm$0.45 
 & 31.30$\pm$0.08\\ 

  & \textsc{adacontrast} 
  & \textcolor{dontkillmyeyesgreen}{33.29$\pm$0.09} 
 & 31.81$\pm$0.14 
 & \textcolor{dontkillmyeyesgreen}{33.29$\pm$0.09} 
 & \textcolor{debianred}{3.90}$\pm$0.26 
  & 31.79$\pm$0.14
 & 31.81$\pm$2.01 
 & 18.94$\pm$0.13 
 & 33.29$\pm$0.09\\ 

   & \textsc{memo} 
  & 24.17$\pm$0.99 
 & 23.93$\pm$0.06 
 & \textcolor{debianred}{22.55$\pm$1.20} 
 & \textcolor{dontkillmyeyesgreen}{24.37$\pm$0.17} 
  & 23.37$\pm$2.13
 & 23.59$\pm$1.82 
 & 24.17$\pm$0.56 
 & 25.01$\pm$0.01 \\ 

 & \textsc{lame} & \multicolumn{8}{c}{17.72$\pm$0.02} \\ 

  & \textsc{rmt-sf} 
  & \textcolor{dontkillmyeyesgreen}{28.81$\pm$0.05} 
 & 27.54$\pm$2.6 
 & \textcolor{debianred}{23.76$\pm$0.1} 
 & \textcolor{debianred}{23.76$\pm$0.1} 
  & \textcolor{dontkillmyeyesgreen}{28.81$\pm$0.05}
 & 30.51$\pm$0.1 
 & 29.16$\pm$0.13 
 & 31.54$\pm$0.15
 \\ 

\midrule
\multirow{7}{*}{\textsc{DomainNet-126}} & \textsc{source} & \multicolumn{8}{c}{54.71}\\
  & \textsc{tent} 
 & 49.92$\pm$0.08 
 & \textcolor{dontkillmyeyesgreen}{52.13$\pm$0.06} 
 & \textcolor{debianred}{8.21$\pm$0.11} 
 & \textcolor{dontkillmyeyesgreen}{52.13$\pm$0.06} 
  & 49.91$\pm$0.085
 & 52.04$\pm$0.19 
 & 50.32$\pm$0.08 
 & 52.24$\pm$0.01\\ 

 & \textsc{eata} 
 & 50.17$\pm$0.06 
 & \textcolor{dontkillmyeyesgreen}{53.37$\pm$0.92} 
 & \textcolor{debianred}{19.76$\pm$23.79} 
 & \textcolor{dontkillmyeyesgreen}{53.59$\pm$0.95} 
  & 50.17$\pm$0.07
 & 53.30$\pm$0.54 
 & 51.88$\pm$0.19 
 & 54.44$\pm$0.27\\ 

  & \textsc{sar} 
  & 49.73$\pm$0.09 
 & 50.91$\pm$1.15 
 & \textcolor{debianred}{28.48$\pm$5.55} 
 & \textcolor{dontkillmyeyesgreen}{51.24$\pm$0.19} 
  & 49.82$\pm$0.15
 & 51.45$\pm$0.44 
 & 50.55$\pm$0.31 
 & 52.75$\pm$0.07\\ 

  & \textsc{adacontrast} 
  & \textcolor{dontkillmyeyesgreen}{\textbf{56.51$\pm$0.04}} 
 & 55.69$\pm$0.15 
 & 46.50$\pm$0.42 
 & \textcolor{debianred}{19.61$\pm$0.79} 
  & \textcolor{dontkillmyeyesgreen}{56.51$\pm$0.06}
 & 54.32$\pm$3.11 
 & 51.00$\pm$0.79 
 & 56.51$\pm$0.04 \\ 

   & \textsc{memo} 
  & \textcolor{dontkillmyeyesgreen}{\textbf{53.98$\pm$0.04}} 
 & 53.71$\pm$0.03 
 & \textcolor{debianred}{51.30$\pm$0.07} 
 & 53.64$\pm$0.01 
 & \textcolor{dontkillmyeyesgreen}{\textbf{53.98$\pm$0.04}}
 & 53.90$\pm$0.08 
 & 53.68$\pm$0.02 
 & 53.98$\pm$0.04 \\ 

 & \textsc{lame} & \multicolumn{8}{c}{54.34$\pm$0.01} \\ 
  
  & \textsc{rmt-sf} 
  & 51.02$\pm$0.05 
 & \textcolor{dontkillmyeyesgreen}{52.67$\pm$0.02} 
 & \textcolor{debianred}{46.36$\pm$0.64} 
 & \textcolor{debianred}{46.36$\pm$0.64} 
  & 51.01$\pm$0.05
 & 52.40$\pm$0.20 
 & 52.42$\pm$0.11 
 & 52.70$\pm$0.01\\ 

\midrule
\multirow{8}{*}{\textsc{ImageNet-R}} & \textsc{source} & \multicolumn{7}{c}{36.17}\\
   & \textsc{tent} 
   & 37.43$\pm$0.28 
  & \textcolor{dontkillmyeyesgreen}{\textbf{38.92$\pm$0.07}} 
  & \textcolor{debianred}{11.01$\pm$0.59} 
  & 38.30$\pm$0.17 
   & 36.51$\pm$0.18
  & 38.40$\pm$0.13 
  & 37.55$\pm$0.14 
  & 38.92$\pm$0.07\\ 

 & \textsc{eata} 
 & 37.17$\pm$0.13 
 & \textcolor{dontkillmyeyesgreen}{\textbf{43.09$\pm$0.75}} 
 & \textcolor{debianred}{2.35$\pm$1.46} 
 & 42.27$\pm$0.18 
  & 37.17$\pm$0.13
 & 42.03$\pm$0.43 
 & 39.70$\pm$0.83 
 & 43.09$\pm$0.75\\ 

  & \textsc{sar} 
  & 37.09$\pm$0.73 
 & \textcolor{dontkillmyeyesgreen}{\textbf{41.94$\pm$0.24}} 
 & \textcolor{debianred}{24.28$\pm$0.71} 
 & \textcolor{dontkillmyeyesgreen}{\textbf{41.94$\pm$0.24}} 
  & 36.45$\pm$0.20
 & 41.31$\pm$1.06 
 & 37.84$\pm$0.17 
 & 41.94$\pm$0.24\\ 

  & \textsc{adacontrast} 
  & 39.23$\pm$0.12 
 & \textcolor{dontkillmyeyesgreen}{\textbf{39.53$\pm$0.14}} 
 & 37.94$\pm$0.26 
 & \textcolor{debianred}{15.00$\pm$0.39} 
  & 39.23$\pm$0.12
 & 38.86$\pm$0.78 
 & 35.80$\pm$0.18 
 & 39.53$\pm$0.14\\ 

 & \textsc{memo} 
 & 40.40$\pm$0.01 
 & \textcolor{dontkillmyeyesgreen}{40.79$\pm$0.04} 
 & \textcolor{debianred}{39.35$\pm$1.19} 
 & \textcolor{debianred}{39.36$\pm$0.01} 
  &  40.40$\pm$0.01
 & 40.49$\pm$0.12 
 & 40.12$\pm$0.24 
 & 40.83$\pm$0.02\\ 

  & \textsc{lame} & \multicolumn{8}{c}{35.95$\pm$0.01} \\ 

  & \textsc{rmt-sf} 
  & \textcolor{debianred}{36.81$\pm$0.16} 
 & \textcolor{dontkillmyeyesgreen}{\textbf{40.97$\pm$0.32}} 
 & \textcolor{dontkillmyeyesgreen}{\textbf{40.97$\pm$0.32}} 
 & \textcolor{dontkillmyeyesgreen}{\textbf{40.97$\pm$0.32}} 
  & \textcolor{debianred}{36.81$\pm$0.16}
 & 40.64$\pm$0.78 
 & 38.77$\pm$0.10 
 & 40.97$\pm$0.32\\ 

\midrule

\multirow{7}{*}{\textsc{ImageNet-V2}} & \textsc{source} & \multicolumn{8}{c}{58.72}\\
  & \textsc{tent} 
 & \textcolor{dontkillmyeyesgreen}{58.69$\pm$0.10} 
 & \textcolor{dontkillmyeyesgreen}{58.73$\pm$0.07} 
 & \textcolor{debianred}{55.82 $\pm$0.24} 
 & 55.99$\pm$0.06 
  & 57.41$\pm$0.11 
 & 57.69$\pm$1.56 
 & 58.70$\pm$0.11 
 & 58.89$\pm$0.13\\ 

 & \textsc{eata} 
 & \textcolor{dontkillmyeyesgreen}{58.72$\pm$0.06} 
 & \textcolor{dontkillmyeyesgreen}{58.72$\pm$0.06} 
 & \textcolor{debianred}{21.52$\pm$25.56} 
 & 56.83$\pm$0.27 
  & \textcolor{dontkillmyeyesgreen}{58.67$\pm$0.12} 
 & 58.18$\pm$0.69 
 & 58.62$\pm$0.12 
 & 58.93$\pm$0.09\\ 

  & \textsc{sar} 
  & \textcolor{dontkillmyeyesgreen}{58.70$\pm$0.10} 
 & \textcolor{dontkillmyeyesgreen}{58.65$\pm$0.09} 
 & \textcolor{debianred}{57.11$\pm$0.04} 
 & 57.26$\pm$0.09 
  & \textcolor{dontkillmyeyesgreen}{58.67$\pm$0.11}
 & 58.33$\pm$0.68 
 & 58.71$\pm$0.09 
 & 58.90$\pm$0.07\\ 

  & \textsc{adacontrast} 
  & \textcolor{dontkillmyeyesgreen}{\textbf{62.28$\pm$0.19}} 
 & 62.07$\pm$0.11 
 & \textcolor{dontkillmyeyesgreen}{\textbf{62.28$\pm$0.19}} 
 & \textcolor{debianred}{26.68$\pm$1.68} 
  & \textcolor{dontkillmyeyesgreen}{\textbf{62.28$\pm$0.19}}
 & 62.17$\pm$0.26 
 & 58.05$\pm$0.33 
 & 62.28$\pm$0.19 \\ 

   & \textsc{memo} 
  & 62.91$\pm$0.16 
 & \textcolor{dontkillmyeyesgreen}{63.12$\pm$0.13} 
 & 62.19$\pm$0.09 
 & \textcolor{debianred}{62.02$\pm$0.02} 
 &   62.91$\pm$0.16
 & 62.32$\pm$0.22 
 & 62.66$\pm$0.03 
 & 63.20$\pm$0.06 \\ 

 & \textsc{lame} & \multicolumn{8}{c}{1.07$\pm$0.01} \\ 
  
  & \textsc{rmt-sf} 
  & \textcolor{dontkillmyeyesgreen}{\textbf{58.70$\pm$0.15}}
 & 58.66$\pm$0.02 
 & 58.14$\pm$0.05 
 & \textcolor{debianred}{42.28$\pm$22.41} 
  & \textcolor{dontkillmyeyesgreen}{\textbf{58.70$\pm$0.15}}
 &  58.4$\pm$0.26 
 & 48.36$\pm$14.48 
 & 58.75$\pm$0.15\\ 

\midrule
 
\end{tabular}
}

\caption{Final accuracy of evaluated test-time adaptation methods under different model selection strategies. \textcolor{dontkillmyeyesgreen}{Green} color marks the best surrogate strategy (five first columns) for each of the adaptation methods (\textbf{row-wise}), and the \textcolor{debianred}{red} color marks the worst one.  \textcolor{dontkillmyeyesgreen}{\textbf{Green bolded text}} marks unsupervised results that match oracle selection strategy.
}
\label{table:main_results}
\vspace{-0.5cm}
\end{table*}




\end{document}